\documentclass{article}
\newcommand{\paper}[2]{\cite{#1}} 

\usepackage{longtable}

\usepackage{arydshln}
\usepackage{amsmath}
\usepackage{graphicx}
\usepackage{enumitem}
\usepackage[numbers]{natbib}

\usepackage[greek,english]{babel}
\usepackage{xspace}
\usepackage{lmodern}
\usepackage{listings}
\usepackage{pgfplots}
\pgfplotsset{compat=1.11}
\usepackage{hyperref}
\makeatletter
\newcommand\footnoteref[1]{\protected@xdef\@thefnmark{\ref{#1}}\@footnotemark}
\makeatother

\usepackage{caption}
\usepackage{algorithm}
\usepackage{algorithmicx}
\usepackage{algpseudocode}
\newcommand{\mycomment}[1]{}
\newcommand{\todo}[1]{{\color{red}#1}}
\newcommand{\new}[1]{{#1}}  

\newcommand{\neo}[1]{{\color{blue}#1}}  
\newcommand{\revision}[1]{{\color{purple}#1}}  
\newcommand{\revisionsecond}[1]{{\textbf{\color{blue}#1}}}  

\usepackage{csquotes}
\usepackage{longtable}

\mycomment{

}

\newcommand{\myfootnote}[1]{{\ \small{[#1]}}} 

\usepackage{arxiv}

\usepackage[utf8]{inputenc} 
\usepackage[T1]{fontenc}    
\usepackage{hyperref}       
\usepackage{url}            
\usepackage{booktabs}       
\usepackage{amsfonts}       
\usepackage{nicefrac}       
\usepackage{microtype}      
\usepackage{lipsum}
\usepackage{graphicx}
\graphicspath{ {./images/} }

\usepackage{graphicx}%
\usepackage{multirow}%
\usepackage{amsmath,amssymb,amsfonts}%
\usepackage{amsthm}%
\usepackage{mathrsfs}%
\usepackage[title]{appendix}%
\usepackage{xcolor}%
\usepackage{textcomp}%
\usepackage{textgreek}
\usepackage{manyfoot}%
\usepackage{booktabs}%
\usepackage{algorithm}%
\usepackage{algorithmicx}%
\usepackage{algpseudocode}%
\usepackage{listings}%

\title{NLP for the Greek language: a longer survey}

\author{
Katerina Papantoniou \\
  FORTH-ICS, \\
 Crete, Greece \\
  \texttt{papanton@ics.forth.gr} \\
   \And
Yannis Tzitzikas \\
  FORTH-ICS \\  \& \\
 Computer Science Department, University of Crete,\\ 
 Crete, Greece \\
  \texttt{tzitzik@ics.forth.gr} 
}

\begin{document}
\maketitle




\begin{abstract}
	English language is in the spotlight of the Natural Language Processing (NLP) community with other languages, like Greek, lagging behind in terms of offered methods, tools and resources. Due to the increasing interest in NLP, in this paper we try to condense  research efforts for the automatic processing of Greek language covering the last three decades. In particular, we list and briefly discuss related works, resources and  tools, categorized according to various processing layers and contexts. We are not restricted to the modern form of Greek language but 
	also cover Ancient Greek and various Greek dialects. This survey can  be useful for researchers and students interested in  NLP tasks, Information Retrieval and Knowledge Management for the Greek language.  
\end{abstract}


\section{Introduction}


There is a wide variety of methods, tools and resources for processing text in the English language. 
However this is not the case for the Greek language
even though it
has a long documented history 
spanning at least 3,400 years of written records (including texts in syllabic script),
and
28 centuries (Archaic period - new) of written text with alphabet \citep{Horrocks2010,Britanica2021GreekLanguage}. The over 2500 years literary tradition of Greek is also notable.
\mycomment{
    With alphabet: 
     Archaic and Classical periods (beginning with the adoption of the alphabet, 
     from the 8th to the 4th century BCE)
   Ara me alfavito exoume 8 aiwnes BC kai 20 meta, synolo 28.
    From Wikipedia:
    The Greek language holds an important place in the history of the Western world and Christianity; the canon of ancient Greek literature includes works in the Western canon such as the epic poems Iliad and Odyssey. Greek is also the language in which many of the foundational texts in science, especially astronomy, mathematics and logic and Western philosophy, such as the Platonic dialogues and the works of Aristotle, are composed; the New Testament of the Christian Bible was written in Koiné Greek. Together with the Latin texts and traditions of the Roman world, the study of the Greek texts and society of antiquity constitutes the discipline of Classics.
    During antiquity, Greek was a widely spoken lingua franca in the Mediterranean world, West Asia and many places beyond. It would eventually become the official parlance of the Byzantine Empire and develop into Medieval Greek.[5] In its modern form, Greek is the official language in two countries, Greece and Cyprus, a recognized minority language in seven other countries, and is one of the 24 official languages of the European Union. The language is spoken by at least 13.4 million people today in Greece, Cyprus, Italy, Albania, and Turkey and by the Greek diaspora.
    Greek roots are often used to coin new words for other languages; Greek and Latin are the predominant sources of international scientific vocabulary.
}
To aid those that are interested in 
using, developing or advancing the
techniques for Greek processing, in this paper we survey related works
and resources
organized in categories.
We hope this collection and categorization of works to be useful for students and researchers 
interested in  
NLP tasks,
Information Retrieval and 
Knowledge Management 
for the Greek language. 
Indeed, there is an everlasting and vivid discussion on how computational tools 
can help in the study and research of Greek texts starting from  
the rich production of Ancient Greek literary work \citep{Packard1973,Galiotou2014DigitalHumanities,zaki2015problems,Barker2016DigitalHumanities,Monachini2018DigitalCA}
ending with Modern Greek and the various Greek dialects.



Modern Greek is an independent branch of the Indo-European family of languages, the sole descendant of Ancient Greek.
It is the official language in Greece and  one of the two official languages in Cyprus (along with Turkish).  There are Greek-speaking enclaves in Albania, Bulgaria, Italy, North Macedonia, Romania, Georgia, Ukraine, Lebanon, Egypt, Israel, Jordan, and Turkey.  The Greek language  is also spoken by a significant number of  people belonging to Greek diaspora i.e., communities of people with Greek origin living outside of Greece and Cyprus. Such Grecophone communities exist in USA, e.g., Tarpon Springs, Astoria,  Australia e.g., Melbourne,  Canada, Germany, the United Kingdom and in other countries too.



A few  characteristics or  particularities of the Greek language
that make some NLP tasks challenging, follow. 
Greek words are typically polysyllabic with function words to be usually shorter than content words~\citep{Chitiri1994}. There is a highly consistent alphabetic orthography with minor complexities~\citep{Verhoeven2021}.
Morphology is a developed component in the grammar of Greek since it  exhibits productive word-formation patterns for both derivation e.g., suffixation, prefixation 
and compounding \citep{Ralli2016}. 
It has an extensive set of productive derivational affixes, whereas the rich productivity of compounds stems from a relatively limited system of compounding. Greek  is  a language highly inflected based predominantly on suffixes but there are also few infixes and one inflectional prefix.
%
There are many different suffixes in the nominal system due to the number (singular and plural),
    the four 
    cases (nominative, genitive, accusative, and 
    vocative), the three genders i.e.,  masculine, feminine and neuter, let alone irregular formations. Case, number and gender are marked on the noun as well as on articles and adjectives modifying it. Verb in turn have two different stems (imperfective and perfective), 
    while there are several cases of irregular verbs.

  In terms of syntax,  Greek language has a fairly free word order and it is a null-subject language (e.g., the phrase ``Jason has started" 
  can be  expressed in Greek both as ``\selectlanguage{greek}Ο Ιάσονας ξεκίνησε\selectlanguage{english}",
     and ``\selectlanguage{greek}Ξεκίνησε.\selectlanguage{english}" depending on the context) \citep{Warburton1987Pro}. Other characteristic features of the language are:
     the clitic doubling (a tendency to use pre-verbal clitic object pronouns redundantly e.g., ``I gave Maria the sweets" /``\selectlanguage{greek}(Της) (τα) έδωσα της Μαρίας τα γλυκά \selectlanguage{english}", ``I show John"
     / ``\selectlanguage{greek}(Τον) είδα (τον Γιάννη)\selectlanguage{english}) and the lack the  lack of a typical infinitive. In the course of the years, we observe the merging of the dative and the genitive case. More specifically,  indirect objects are expressed partly through genitive forms of nouns or pronouns, and partly through a periphrasis consisting of the preposition \selectlanguage{greek}σε / \selectlanguage{english}to while some  dative fossils still remain\footnote{More examples can be found at \url{https://www.eleto.gr/download/Orogramma/Dotikes_KValeontis.pdf}.} for example  \selectlanguage{greek}ελλείψει / \selectlanguage{english}in shortage, \selectlanguage{greek}επ’ αυτοφώρω / \selectlanguage{english} red-handed. Lastly, the nominative and accusative case syncretism creates interesting ambiguities in sentences like  \selectlanguage{greek}Ανατροπές έφεραν τα εκλογικά αποτελέσματα / \selectlanguage{english}The election results brought upheavals.

   \mycomment{  SOS: Isws na paei 3.5
  An example of the productivity of derivational morphology of the \revisionsecond{Modern} Greek language is given in the  following three words
  that  stem from the same base:
    \selectlanguage{greek}
        πράττω, πράξη, πρακτικός,
    \selectlanguage{english} a verb (/act/), a noun (/act/) and an adjective (/practical/) respectively.
    \mycomment{\url{http://www.asteri.ws/en/verb/stem-en.html}
    stems instead of themes and imperfective and perfective instead of present and ...    
    }
}
    
{On the evolution of the language, over the centuries, 
  we could mention that Modern Greek language's rich vocabulary has retained many Ancient Greek words either as whole words or as stems. These words have morphologically adapted to the rules of Modern Greek and also many ancient words  have also evolved grammatically or semantically. In terms of grammar, some of the changes include  the loss of the dative case,  the loss of the inﬁnitive and the loss of the dual number~\citep{Margoni2023ModernGreek}. Ancient Greek had verbs while the modern language has verb phrases (analytical forms of verbs). Greek seems to be affected by both language internal factors and by borrowing
from languages it has been in contact with~\citep{ManolessouRalli+2015+2041+2061}. Most of the older ``loans'' has been fully adapted  to the system of the language while some newer loans mostly from English and French have remained inact causing difficulties in the computational processing tools. An analytical description of Modern Greek Grammar can be found in  ~\citep{Holton2012GreekGrammar}.
  




\mycomment{YT: 
  Apo kapoies diafaneies poy eixa grapsei  prin 15 xronia, tote me ton Greek stemmer, exw kapoia sxolia pou ta evala sto google doc.
}
  \mycomment{KP: 
Highly consistent alphabetic orthography with minor complexities. Some phonemes map to more than one graph.
}

The rest of this paper is organized as follows.
Section \ref{sec:RW} describes 
previous surveys on Greek NLP, 
and the methodology that we followed.
Section \ref{sec:survey}  surveys the works on Greek NLP. Section \ref{sec:EvaluationCollections} 
describes some noteworthy  resources and evaluation collections that are available,
and finally,
Section \ref{sec:CR} concludes the paper.
\section{Other surveys and methodology}
\label{sec:RW}
	
There are not many papers that attempt to survey this area.
We have found the 2010 paper 
\citep{tsalidis2010nlp},
the white paper \citep{gavrilidou2012greek} and its recent update~\citep{gavriilidou2022greek}.
There is also some surveys about particular topics like \citep{tsakalidis2018sentiment} that focuses on tools for sentiment analysis in Greek, 
 \citep{Nikiforos2021greeksrurveydataset}
that surveys 
social web datasets for Greek and the applicable mining techniques,  
 \citep{Krasadakis2022Survey} that surveys  NLP for the Greek legal domain and  \citep{Vagelatos2022Survey} that is focused in  the  healthcare NLP infrastructure for the Greek Language. 

For the Ancient Greek, we found an entry in ``Encyclopedia of Ancient Greek Language and Linguistics'' \citep{2013RelatedWork} that briefly discusses available digital resources and methods for the processing of Ancient Greek.

\new{
    The current paper
    is  an expanded and more analytical version of the paper
    \citep{PapantoniouTzitzikas2020setn}.
    In comparison to that paper, 
    the current survey:
    is more complete
    in the sense that \citep{PapantoniouTzitzikas2020setn} contained 99 references,
    while the current paper contains 221
    references,
    it provides a more rich categorization,
    and
    it contains 
    more   tables that summarize the available information
    for aiding the reader to find out what it is available,
    As regards the latter, 
    the current paper
    contains 
    tables
    that describe
    the available
    (i)
    domain specific annotated datasets 
    (Table \ref{datasets}),
    (ii)  embeddings for the Modern Greek  (Table \ref{embedding}) and  Ancient Greek  (Table \ref{ancientembedding}) language,
    (iii) toolkits and their functionality  
    (Tables \ref{toolkits} and \ref{ancienttoolkits}),
    (iv) tools and related references (Table \ref{tools}), and 
    (v) tools and resources offered via online interfaces (Table \ref{interfaces}). 
}

\subsection{Methodology for the current survey}
\label{sec:Methodology}

For finding the related works we used Google Scholar, ResearchGate, DBLP, and the ACL Anthology in the period: April 2020 - December 2022,
without any restriction on the publication date. Although we did not used any filter, 
the majority of works that we retrieved concerns the last three decades. 
We  started by collecting all papers that contain the word ``Greek'' in the title 
and we used various extra keywords and keyphrases for finding the related papers. 

In particular we used the following queries : 
``Greek" plus one of the following  (we list them on alphabetical order and in lowercase):
\begin{displayquote}
  {\em 
    ``abusive  language, anaphora resolution, ancient nlp, ancient, argument  extraction, argument  mining, authorship attribution, authorship, chatbot, computational  journalism, computational  linguistics, computational  pragmatics, computational  semantics, computational lexicons, controlled natural languages, coreference, corpora  list, corpora, corpus, cypriot, dataset, deception  detection, dialects, dialogue systems, education nlp, embeddings, entailment, fake news, greeklish, hate speech, hyphenation, law, legal, lemmatization, machine translation, modern, morphology, named  entities  extraction, named  entities  recognition, named  entity  linking, natural  language  processing, natural  language  understanding, natural language generation, natural language processing literature review, natural language processing survey, nlp in  health, ocr, offensive language, opinion analysis, phonetics, phonology, abusive language, pos, question answering, readability, sentiment analysis, sentiment lexicons, slang, spelling checkers, stemmers, stemming, summarization, syllabification, syntactic parsers, syntax, text processing, transformers, transliteration".
    }
\end{displayquote} 

Whenever we considered necessary, we expanded our searches in order to find follow-up, more mature works or related research projects, growing in this way the thread of our searches. A similar process
 was  followed  for finding resources and evaluation collections. The queries were modified accordingly, and apart from the aforementioned sources we also used  general-purpose web search engines.

We  did not focus on  papers that deal with  multilingualism  and could potentially contain Greek as well. Consequently, the survey is by no means complete,  however we tried to find the more relevant and fundamental works and resources for giving a concise overview. 
However 
the reader must be aware and keep an eye also on approaches like \textit{transfer learning},  \textit{distant supervision} and \textit{data augmentation} since they can be of great benefit on NLP tasks in the absence of resources \citep{Alyafeai2020arXivtransfer, hedderich-etal-2021-survey}.
In the sections that follow when we refer to Ancient Greek,  we mean the historical language {as determined in the  ISO 639-3 code set classification with  language index \textit{grc}}\footnote{\url{https://iso639-3.sil.org/sites/iso639-3/files/downloads/iso-639-3_Name_Index_Latin1.tab}}. The usual cutoff for this distinction is the date of 1453 (the Fall of Constantinople).

    Later,
    in   Section \ref{sec:Quantitative},
    we  provide  the time distribution of the examined papers.



\section{Survey of works on Greek NLP}
\label{sec:survey}

At first, in Section \ref{sec:Categorization},
we provide a categorization of the various NLP tasks (shown in Figure \ref{fig:OverviewWithYears}).
Then, 
in Section  \ref{sec:Quantitative},
we provide a quantitative overview of
the surveyed works.
The rest subsections,
from 
Section \ref{sec:OCR} 
to 
Section \ref{sec:Misc},
are based on the categorization introduced in Section  \ref{sec:Quantitative}. Tables~\ref{tbl:GreekPOSwork} and~\ref{tbl:GreeNERtools} that are embedded in Section~\ref{sec:survey} summarize works in POS tagging and Named Entity Recognition (NER) respectively.

\subsection{Categorization}
\label{sec:Categorization}

In this survey we group
the  available works
according to various processing layers and contexts,
namely:
object character recognition (\S \ref{sec:OCR}),
phonetics (\S \ref{sec:Phonetics}),
morphology (\S \ref{sec:Morphology}),
syntax (\S \ref{sec:Syntax}),
embeddings (\S \ref{sec:Embeddings}),
semantics (\S \ref{sec:Semantics}),
pragmatics (\S  \ref{sec:Pragmatics}),
sentiment analysis (\S \ref{sec:SA}),
question answering (\S \ref{sec:QA}),
natural language generation (\S \ref{sec:NLG}),
machine translation (\S \ref{sec:MT}).
An overview of the topics that are addressed
(up to some degree) is  given in Figure 
\ref{fig:OverviewWithYears}.
%

\begin{figure*}
	\centering
	\fbox{\includegraphics[width=0.99\linewidth]{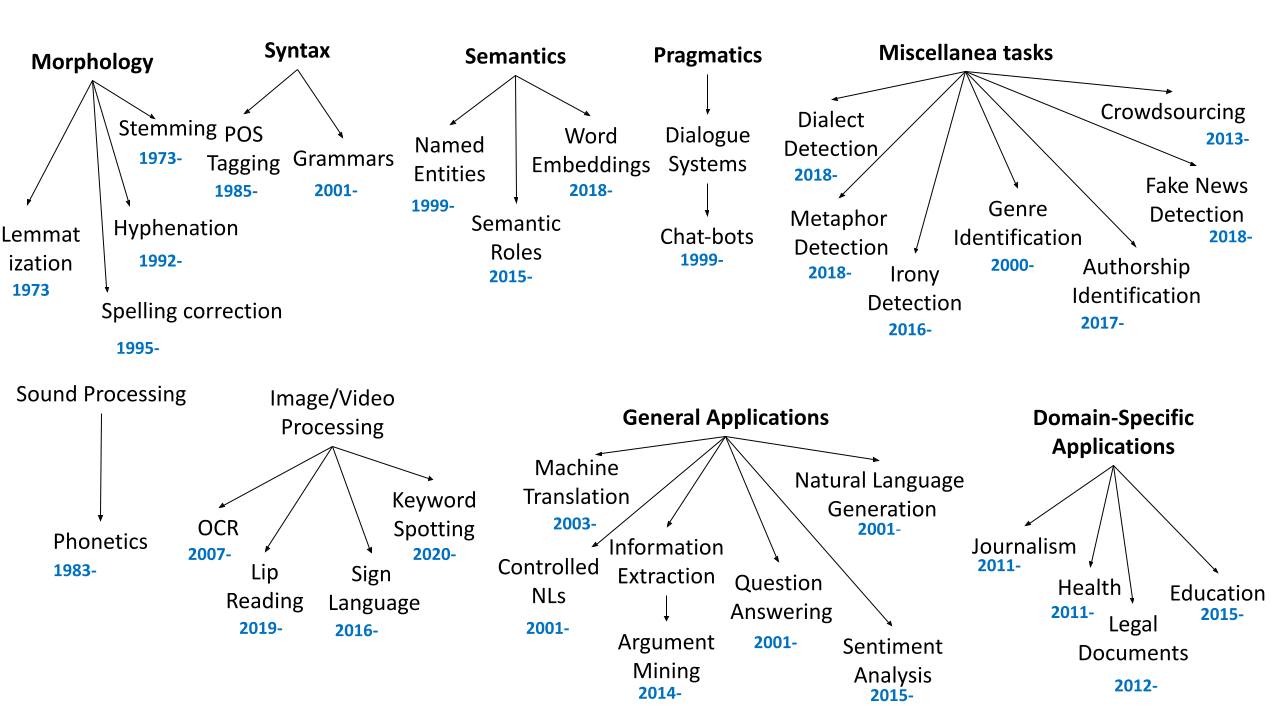}}
	
	\caption{Categorization of works in the survey enriched with the publication year of the oldest work per category that we have included in this survey.}
	\label{fig:OverviewWithYears}
\end{figure*}

\mycomment{===============AUTO XWRIS TA YEARS=============
    \begin{figure}
    	\centering
    	\includegraphics[width=0.99\linewidth]{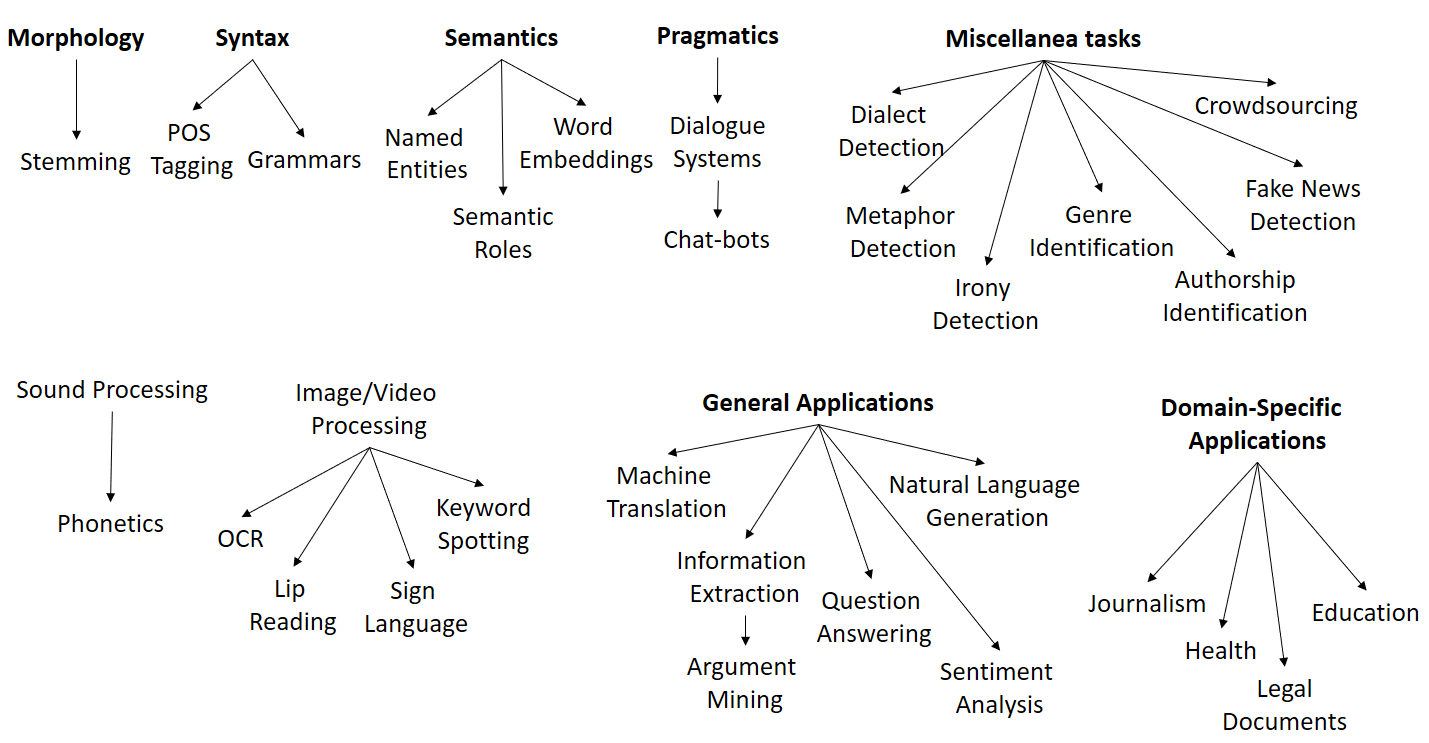}
    	
    	\caption{The categorization of NLP tasks that we use in this survey}
    	\label{fig:Overview}
    \end{figure} 
==}

However we also include a section 
that refers to works that capture several layers,
as well as sections on related topics (\S \ref{sec:Misc}),
including 
dialect detection,
lip reading,
Greek sign language,
keyword spotting,
argument mining,
computational journalism,
health domain,
crowd-sourcing,
metaphor detection,
irony detection,
genre identification,
fake news detection,
legal domain,
authorship,
education - analytics,
controlled natural languages.
Resources and evaluation collections are described afterwards,
in Section \ref{sec:EvaluationCollections}.

We are aware that a work may fit 
to more than one category,
however we have decided  to avoid  duplicates.
We have also decided to avoid 
the inclusion of quantitative results
because such  numbers would  not be comparable
since in most cases they are based 
on experiments over different collections
(the interested user can refer to the mentioned papers
for quantitative results).
The works in each section
are described in chronological order.
Whenever it was more sensible we also grouped some papers thematically. Note also that some works or     projects  
have been published in more than one paper.
In such cases we include one of these works; 
the interested reader could perform more searches to find subsequent or related works 
through the usual channels (e.g., google scholar, dblp, authors' pages or pages of the organizations they are affiliated to) since it is impossible this survey to be exhaustive.
%
As regards distribution in time,
  Figure \ref{fig:OverviewWithYears}
  also shows the 
  year intervals
  of the found works by category.

    To aid the understanding
    of the categorization and the 
    output of the basic NLP tasks,
    
    Figure 	\ref{fig:Example}
    shows a small and typical NLP example, i.e., a small Greek phrase,
    specifically ``\selectlanguage{greek}{Ο Αριστοτέλης είδε την κοπέλα με το τηλεσκόπιο}\selectlanguage{english}''
    (Aristotle saw the girl with the telescope),
    and 
    how the results
    of some basic NLP tasks
    might be.
   
    Notice the various ambiguities that exist:
    ``Aristotle'' could refer 
    either 
    to the ancient philosopher, 
    or to Aristotle Onassis (Greek shipping magnate),
    or even to a modern person,
    while
     the phrase ``with the telescope'' may refer to ``saw'' 
    (i.e., Aristotle saw the girl {\em through his} telescope)
    or to ``girl''
    (e.g., Aristotle saw the girl {\em carrying her} telescope),
    exhibiting a kind of syntactic ambiguity i.e., prepositional phrase (PP) attachment.

Of course, there are many other types of ambiguities and challenges that computational processing tools must grasp. 
The {\em free word order}  can pose some difficulties since even this simple utterance can be formed in various alternative ways giving emphasis on different parts
(e.g.  the above phrase could be formed as
''\selectlanguage{greek}{την κοπέλα με το τηλεσκόπιο είδε ο Αριστοτέλης}\selectlanguage{english}'').
In addition, 
the {\em clitic doubling}
(e.g. expressing the above utterance  as 
    ''\selectlanguage{greek}{την κοπέλα με το τηλεσκόπιο \underline{την} είδε ο Αριστοτέλης}\selectlanguage{english}''),
and
the {\em pro-drop} feature
must be also taken into account in tasks like syntactic parsing, machine translation, natural language generation, coreference resolution etc.

Such ambiguities make
    the general objective
    of  NLU (Natural Language Understanding) 
    very challenging.
    We can understand the intended meaning(s)
    only by considering 
    (and properly analyzing)
    the context of the text.
    

    \mycomment{== previous verion with more 
        {\bf OLD:}
        For instance, 
        the slightly 
        modified  phrase  
        of Figure 	\ref{fig:Example},
        ``\selectlanguage{greek}Ο Αριστοτέλης είδε την κοπέλα με το κινητό\selectlanguage{english}''
             (Aristotle saw the girl with the mobile),
        could have several meanings,
        e.g., ``Aristotle'' could refer 
        either 
        to the ancient philosopher, 
        or to Aristotle Onassis (Greek shipping magnate),
        or even to a modern person,
        while
        the  ``mobile'' could refer to a mobile object or to a cell phone,
        finally the phrase ``with the mobile'' may refer to ``saw'' 
        (e.g., Aristotle saw the girl {\em through his} mobile)
        or to ``girl''
        (e.g., Aristotle saw the girl {\em carrying her} mobile).
    ===}
    \mycomment{=== FOR REASONS OF SPACE===
        just a few follow:
        \begin{itemize}
            \item M1:
            \selectlanguage{greek}
            Ο Αριστοτέλης (ο Αρχαίος φιλόσοφος)  είδε την κοπέλα με το κινητό (τρίποδο),
            \selectlanguage{english}
            \new{
            (Aristotle (the ancient philosopher) saw the girl with the mobile (tripod),
            }
            \item M2: 
            \selectlanguage{greek}
            Ο Αριστοτέλης (Ωνάσης) είδε την κοπέλα με το κινητό (τραπέζι)
            \selectlanguage{english}
            \new{
            (Aristotle (Onassis) saw the girl with the mobile (table),
            }
            \item M3: 
            \selectlanguage{greek}
            Ο Αριστοτέλης (σύγχρονο πρόσωπο) είδε την κοπέλα με το κινητό (του)
            \selectlanguage{english}
            \new{
            (Aristotle (a modern person) saw the girl with the mobile (his cell phone),
            }
            \item M4: 
            \selectlanguage{greek}
            Ο Αριστοτέλης (σύγχρονο πρόσωπο) είδε την κοπέλα με το κινητό (της)
            \selectlanguage{english}
            \new{
            (Aristotle (a modern person) saw the girl with the mobile (her cell phone)
            }
        \end{itemize}
        \selectlanguage{english}
    ===}

\begin{figure*}
	\centering
	\includegraphics[width=0.99\linewidth]{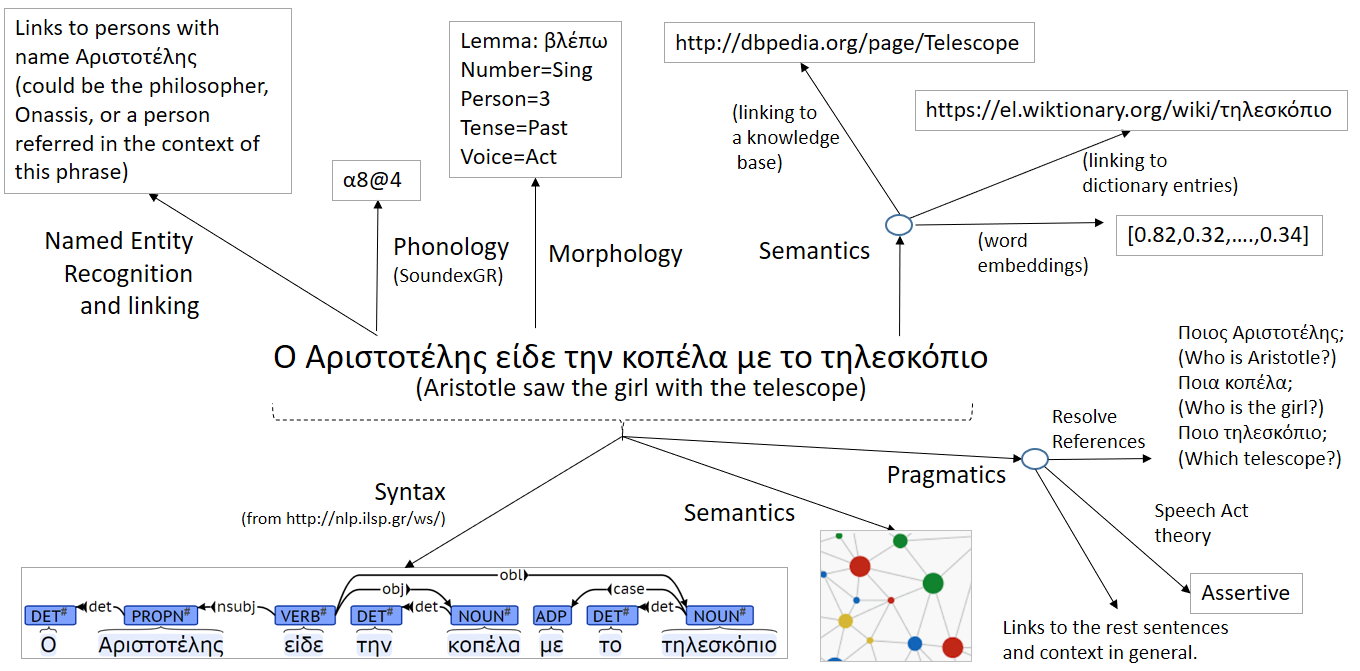}
	\caption{Example of some basic NLP tasks over a Greek phrase}
	\label{fig:Example}
\end{figure*}


\subsection{Quantitative Overview}
\label{sec:Quantitative}

A few  questions that we have had at the beginning were:
(a) what is the distribution of papers per year,
(b) how the number of papers about Ancient Greek 
    compares with those about  Modern Greek?, 
    and (c) does  the zest for the advance of Greek NLP  is  only a matter of researchers and institutions residing in Greece?
%
%
%
\mycomment{===initial version
    \begin{itemize}
        \item Are there more paper about Ancient Greek than Modern Greek?
        \item What is the distribution of papers per year? 
        \item many works on sentiment analysis (with specialized solutions for the Greek language (why?)
            \begin{itemize}
                \item need, many works are in the context of projects, so sentiment analysis one of the most useful applications on NLP with clear linkage to the society
                \item sentiment is affected with social and cultural aspects
             This may result to the need of specialized solutions lexicons or temperamental characteristics for the greek language.
            \end{itemize}
    \end{itemize}

===}
Table \ref{tbl:DistPerYear} shows the {\em distribution in decades} of  the papers mentioned in this survey.  
We have split the last decade in two rows (corresponding to 5 and 7 years respectively) 
to make clear the trend.
%
We  have observed:
(i) a significant overall increase over the last years,
(ii) an increase in works based on neural networks, 
(iii) a relatively high  number of papers on sentiment analysis with specialized solutions for the Greek language, and 
(iv) an emerging interest for dialogue systems.
Another interesting remark 
is that we found  more works about Ancient Greek than we expected,
            19\% of the works that we found concerns Ancient Greek.

As regards to the {\em origin of the authors} of these
works  we noticed a clear distinction between studies about Ancient Greek and Modern Greek.  Most of the studies referring to the Modern Greek language originated from Greek universities and institutions  with some exceptions i.e., the work in \cite{Nikolaenkova2019machinetranslation}, plus three master theses~\citep{ntais2006development}, \citep{Chondrogianni2011Pragmatics} and \citep{pitenis2020offensive}, that are conducted abroad. In contrast, the authors of works related to Ancient Greek  are affiliated with organizations outside Greece. It is indicative that the first NLP work reported in this survey, mostly for historical reasons, concerns a POS tagger created to assist students in the learning of Ancient Greek~\citep{Packard1973}.


\mycomment{===
        \begin{itemize}
            \item We observe a significant overall increase the last years
            \item There are more works about Ancient Greek than we expected,
                    19\% of the works that we found concern and Ancient Greek.
          
            \item We observe an increase in works based on neural networks 
                but also some subfields of NLP such OCR steadily are based on such approaches.

            \item We have observed a relatively high 
                number of 
                papers on sentiment analysis with specialized solutions for the Greek language.
                %
        \end{itemize}
==}
\mycomment{========== GENERAL THOUGHTS
 \item possible direction: transfer learning between rich and low resource languages
 
 ===}

\mycomment{====
    Why no native speakers select Greek language 
    (except from the  self-explained interest on the Ancient Greek) as a case study 
    (some claim relatively rich morphology (to put citation) and others that is a
    less studied language with low coverage of linguistic resources. 
===}

\begin{table}[htbp]
    \caption{Distribution of papers  and resources per year.}
    \label{tbl:DistPerYear}
	\small
	\begin{tabular}{|l||p{1.5cm}|p{1.5cm}|p{1.2cm}|p{1.6cm}||p{0.9cm}|}\hline
	Period  &  Modern Greek  & Ancient Greek &Dialects & Greek Sign Language & Total \\\hline\hline	
	[1990-2000)    {\em  (10 years)}
	    &  10
	    \mycomment{
(\cite{epitropakis1993high},
\cite{goutsos1994towards},
\cite{sgarbas1995pc},
\revisionsecond{\cite{Vagelatosetal95Spelling}},
\cite{karkaletsis1999named},
\cite{petasis1999resolving},
\cite{fourakis1999acoustic},
\cite{orphanos1999decision},
\new{\cite{Charalambous1992Hyphenation}},
\new{\cite{Noussia1997Hyphen}}
	     )   }
	    &  2

	\mycomment{
(\cite{Crane1991Morpheus},
\new{\cite{Crane1995perseus}}

)
}

	    & -
	    
	    & - 
	    
	    &  12 \\\hline
	[2000-2010)   {\em (10 years)}
	    &   29
 \mycomment{(\cite{hatzigeorgiu2000design},
\cite{demiros2000named},
\revisionsecond{\cite{Stamatatos2000TextCategorization}},
\revision{\cite{Papageorgiou2004Multimodal}},
\revision{\cite{Kontos2000arista}},
\revisionsecond{\cite{mikros2007topic}},
\revisionsecond{\cite{Kontos2000stock}},
\revision{\cite{Karberis2002}},
\revison{\cite{anastasiou2009greek}},
\new{\cite{tambouratzis2000automatic}},
\cite{Papageorgiou2000POS},
\cite{petasis2001greek},
\revisionsecond{\cite{vassiliouetal2003evaluating}},
\cite{dimitromanolaki2001large},
\new{\cite{sfakianaki2002acoustic}},
\cite{2005Mamakissummary},  
\cite{prokopidis2005theoretical},
\cite{ntais2006development},
\cite{GreekMitosStemmer2006},
\cite{lucarelli2007named},
\revision{\cite{Chalamandaris2006Greeklish}},
\revision{\cite{Tsourakis2007Greeklish}},
\revision{\cite{Stamou2008Greeklish}},
\cite{vamvakas2007greek},
\cite{papadakos2008anatomy},

\new{\cite{Marzelou2008EntailmentCorpus}},
\new{\cite{Maragoudakis2001NaturalLI}},
\cite{Avramidis2008machinetranslation},
\new{\citep{Maragoudakis2003POS}},
\new{\citep{Lyras2007Phonetics}},
\new{\citep{Kermanidis2009Paraphrasing}},
\new{\citep{Giannakopoulos2009summary}}
\cite{adam2010efficient},
\new{\cite{Kordoni2004HPSG}},
\cite{saroukos2009enhancing})
}
	    &  4
	        \mycomment{
	        \cite{ntzios2007old},
	      \new{\cite{Lee2008nearest}},
	      \new{\cite{haug2008Proiel}},
	            \cite{bamman2009ownership})
	        }
	 & 1
	  \mycomment{
	 \revision{\cite{cyslang2019}}
	 }
	 & 2
	 \mycomment{\cite{Efthimiou2007GSC},
	 \new{\cite{Filippou2004Tex}}
	 
	 }
	 
	 &  36 \\\hline
[2010-2015)   {\em (5 years) }
	    &  25
	        \mycomment{(
 \cite{Goutsos2010corpora},
 \revision{\cite{Dimitropoulou2010subtitles}},
 \revision{\cite{Komianos2012BigFive}},
 \revisionsecond{\cite{Markopoulos2012Law}},
 \revision{\cite{Lyras2010Greeklish}},
\cite{vamvakas2010handwritten},
\cite{tsalidis2010nlp},
\revisionsecond{\cite{Giannakopoulosetal2011multiling}},
\revision{\cite{Panteli2011greeklish}},
\cite{Vagelatos2011biomedical},
\cite{tsoumari2011coreference},
\cite{gavrilidou2012greek},
\cite{Androutsopoulos2013NLG},
\new{\cite{Chondrogianni2011Pragmatics}}
\new{\cite{Vonitsanou2011keyword}}
\new{\cite{IPLR2012}},
\cite{Tsimpouris2014LegalAcronym},
\new{\cite{gf2011}},
\new{\cite{2011Petasisphdthesis}},
\new{\cite{Giouvanakis2013game}},
\new{\cite{Simaki2012}},
\new{\cite{Prokopidis2011}}
\new{\cite{samaridi2014parsing}},
\new{\cite{GAD2014}},
\new{\cite{Pavlidou2012CSG}},
\cite{Goudas2014argument},
) }
	    &   3 \mycomment{(\cite{white2012training},
	    \cite{rydberg2011social},
	    \new{\cite{sorra2014anicentpos}},
	    )}
	   &  1 \mycomment{\new{\cite{Karanikolas2013amigre}}}
	   
	   & -
	    &   29\\\hline
	[2015-2022)   {\em (7 years) }  &   99
	    \mycomment{
	   (\cite{kuzmenko2015disambiguation},
	   \revision{\cite{stamou2020vmwe1}},
	   \revision{\cite{GCC2015}},
      \revisionsecond{\cite{mccarthy-etal-2020-unimorph}},\revisionsecond{\cite{papantoniouetal2022nernel}},
    \revisionsecond{\cite{Mikros2021Tool}},
	   \revision{\cite{mountantonakis2022QA}},
	   \revision{\cite{Alexandridis2021SentimentBERT}},
	     \revision{\cite{stamou2020vmwe}},
	     \revision{\cite{Antoniadis2021PassBot}},
	   \revision{\cite{Stamou2022GOL}},
	   \revision{\cite{Dritsa2022Parliament}},
	   \revision{\cite{fitsilis_fotios_2021_4747452}},
	    \revision{\cite{Sichani2019OCR}},
\cite{Mikros2015tweetsauthorship},
\cite{Gatos2015GRPOLYDB},
\cite{simistira2015recognition},
\cite{kalamatianos2015sentiment},
\cite{markopoulos2015sentiment},
\cite{sardianos2015argument},
\revision{\cite{Pontiki2018Xenophobia}},
\revision{\cite{pontiki-etal-2020-verbal}},
\cite{charalampakis2016comparison},
\cite{Varlokosta2016aphasia},
\cite{takoulidou2016social},
\cite{athanasopoulou2016schizophrenia}
\cite{Ntais2015stemmercomparisoon},
\cite{papanikolaou2016journalism},
\cite{gakis2017design},
\cite{ferra2017tale},
\cite{athanasiou2017novel},
\cite{Kyparissiadis2017Lexicon},
\cite{marakakis2017apantisis},
\cite{prokopidis2017universal},
\cite{spatiotis2017examining},
\cite{pavlopoulos2017abusiveingreek},
\cite{boukala2018absurdity},
\cite{mavridis2018fake},
\cite{outsios2018word},
\cite{tsakalidis2018sentiment},
\cite{florou2018neural},
\cite{garofalakis2018project},
\cite{tsakalidis2018nowcasting},
\cite{angelidis2018named},
\cite{karamitsos2019chatbots},
\cite{kefalidou2019apo},
\cite{outsios2019evaluation},
\cite{lioudakis2019ensemble},
\cite{Nikolaenkova2019machinetranslation},
\cite{Mikros2018Comparison},
\cite{karanikolas2019machine},
\cite{partalidou2019design},
\cite{toulakis2019language},
\cite{kastaniotis2019lip},
\cite{spatiotis2019examining},
\cite{kesidis2020providing},
\cite{pitenis2020offensive},
\cite{soundexGR2021},
\cite{Koutsikakis2020bert},
\cite{gianitsos2019Stylometric},
\cite{tsekouras2020Observatory},
\new{\cite{tambouratzis2016applying}},
\new{\cite{markantonatou-etal-2019-idion}},
\new{\cite{Eloeva2019Corpora}}
\new{\cite{lekea2018hate}},
\revisionsecond{\cite{park2019css10}},
\new{\cite{lampridis2020lyrics}},
\new{\cite{tataridis2019chatbot}},
\new{\cite{Fotopoulou2017FromET}},
\new{\cite{Bartziokas2020}},
\new{\cite{Prokopidis2020}},
\new{\cite{Isard2016}},
\new{\cite{Palogiannidi2016affective}},
\new{\cite{Beleveslis2019}},
\new{\cite{karidi2019}},
\new{\cite{lopes-etal-2016-spedial}},
\new{\cite{papaloukas2021multigranular}},
\new{\cite{Rigas2020Stroke}},
\new{\cite{themistocleous2011computational}},
\new{\cite{nikiforos-kermanidis-2020-supervised}},
\new{\cite{Papantoniou2021nel}},
\new{\cite{Christou2021Katharevousa}},
\new{\cite{Chalkidis2017LegislationSW}},
\new{\cite{korre2021elerrant}},
\new{\cite{Perifanos2021Hate}},
\new{\cite{Tzimokasetal2015Readability}},
\new{\cite{Neofytou2018diakeimenou}},
\new{\cite{Vagelatos2021Logopaignio}},
\new{\cite{Chatzipanagiotidis2021Readability}},
\new{\cite{Papadopoulou2021Nooj}},
\new{\cite{Papantoniou2021ElApril}},
\new{\cite{Wenzek2020CCNet}},
\new{\cite{Dikonimaki2021aueb1}},
\new{\cite{Smyrnioudis2021aueb2}},
\new{\cite{Kyriakakis2018ueb}},
\new{\cite{Alexandridis2021PaloBERT}},
\new{\cite{Kasselimis2020Aphasia}},
\new{\cite{Tantos2016GLC}},
\new{\cite{Ventoura2021TheanoAG}}

)
	    }
	    &  33
\mycomment{
(\cite{assael2019restoring},
\revision{\cite{yousef2022Alignment}},
\cite{bary2017memory},
\revision{\cite{Yamshchikov-etal-2022-plutarch}},
\revision{\cite{McGillivray2021AGVaLex}},
\revision{\cite{Robertson2021Pogretra}},
\revision{\cite{Assael2022}},
\revision{\cite{Palladino2020}},
\revisonsecond{\cite{greCy2022}},
\cite{celano2015semantic},
\cite{celano2016part},
	    \new{\cite{liossis2019ancientpos}},
\cite{gorman2019ancientdependencytree},
\cite{gorman2020dependency},
\cite{keersmaekers2019ancienttreebank},
\cite{keersmaekers2019historical},
\cite{perrone2019gasc},
\cite{robertson2017large},
\cite{stamatatos2017Rhesus},
\cite{rodda2019vector},
\new{\cite{Kontges2020genre}},
\new{\cite{vatri_mcgillivray_2018}},
\new{\cite{Eulexis2019dictionaries}},
\new{\cite{keersmaekers-2020-automatic}}
\new{\cite{Palladino2020}},
\revision{\citep{degraaf2022LREC}}
\new{\citep{CLTK20019}},
\new{\cite{gorman2019}},
\new{\cite{Singh2021ancient-greek-bert}},
\new{\cite{Che2018elmo}},
\new{\cite{keersmaekers2021glaux}},
\new{\cite{McGillivray2019Polysemy}},
\new{\cite{Erdmann2019Herodotus}}

)
}    
	    & 6 \mycomment{
	    \cite{Themistocleous2019dialectclassification},
	   
	   \revison{\cite{hadjidas2015multicast}},
	   
	   	    \cite{sababa2018classifier},
	    \cite{boito2018grikocorpus},
	    \revision{\cite{rijhwani2020ocr}},
	    \cite{anastasopoulos2018POSgriko}
	    }
	    
	    &
	    5 \mycomment{
	    \cite{gkigkelos2017greek},
	    \cite{kouremenos2018novel},
	    \cite{simos2016greek},
	    \cite{Adaloglou2020comprehensive},
	    \cite{Efthimiou2018Polytroponcorpus}
	    }

	    &  143\\\hline

	TOTALS      
	    &   {\bf 164} 
	    &  {\bf 42} 
	       &  {\bf 8} 
	         &  {\bf 7} 
	    &  {\bf 221}  \\\hline
	\end{tabular}
\end{table}

\mycomment{=====
        Now Table \ref{tbl:Affiliations} 
        provides some information about about affiliations.
        They were counted as follows:  \todo{....}.
        
        \begin{table}[htbp]
        	\small
        	\begin{tabular}{|l|l|l|l|}\hline
        	Affiliations   & Country  &  Frequency \\\hline
        	X               & UK       &  3 \\\hline
        	Y               & GR       &  3 \\\hline
        	Z               & RU       &  2 \\\hline
        	W               & FR       &  1 \\\hline
        	A               & BR       &  1 \\\hline
        	\end{tabular}
        	\caption{Most Frequently occurring affiliations }
        	\label{tbl:Affiliations}
        \end{table}
==}

%


With respect to the time period when a topic is first introduced in the literature, 
Figure \ref{fig:OverviewWithYears} presents the topic categorization enriched with year intervals.

  Another interesting question,
  is how many papers
  concern each particular topic.
  Such counts
  can be construed
  as indicators
  of how much/well
  a topic has been elaborated in the literature.
  In addition, 
  they enable the reader
  to identify gaps
  and directions that have not been explored so far.
  To this end,
  Figure  \ref{fig:plotByTopicSorted}
  shows 
  how many papers
  occur in the corresponding subsection
  sorted in descending order.
 
  We can observe that 
  the most frequent topics
  are:  syntax/POS tagging, morphology and sentiment analysis. The same information is presented as a tree map in Figure \ref{fig:treeMap}.
Finally, Figure \ref{fig:plotByTopicAncientSorted}
shows the number of papers per topic, but for Ancient Greek only.

\mycomment{===
\begin{figure}[htbp]
	\centering
	\includegraphics[width=0.95\linewidth]{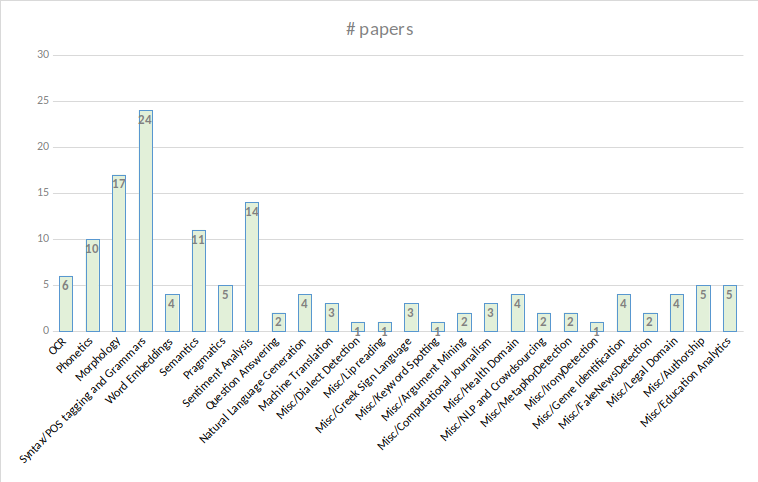}
	\caption{\neo{Number of papers per topic}}
	\label{fig:plotByTopic}
\end{figure} 
====}

\begin{figure}[htbp]
	\centering
	\includegraphics[width=0.95\linewidth]{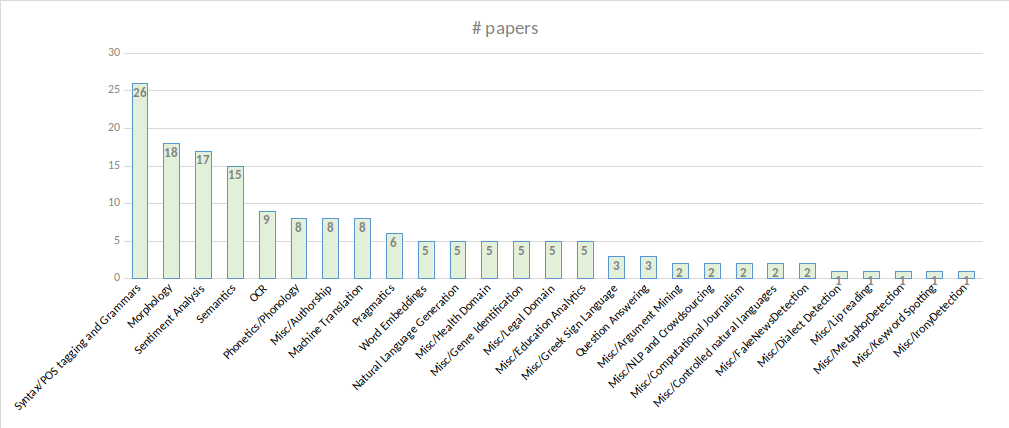}
	\caption{Number of papers per topic (sorted)}
	\label{fig:plotByTopicSorted}
\end{figure}

\begin{figure}[htbp]
	\centering

	\includegraphics[width=0.65\linewidth]{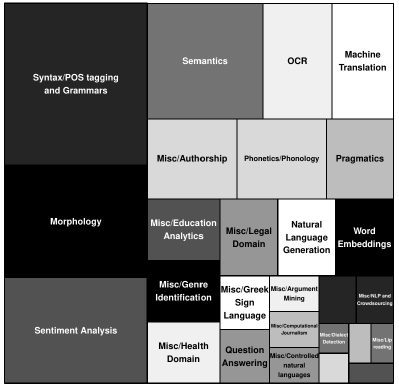}
	
	\caption{Number of papers per topic as a tree map}
	\label{fig:treeMap}
\end{figure}

\begin{figure}[htbp]
	\centering
	\includegraphics[width=0.45\linewidth]{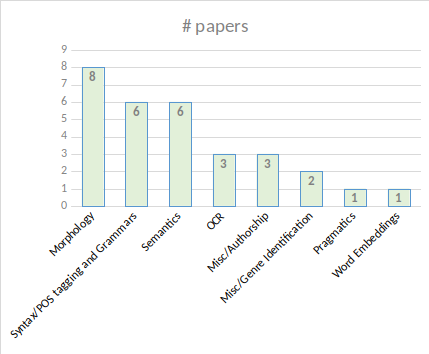}
	\caption{Number of papers about Ancient Greek per topic (sorted)}
	\label{fig:plotByTopicAncientSorted}
\end{figure} 


\subsection{Optical Character Recognition (OCR)}
\label{sec:OCR}
Optical Character Recognition (OCR) 
refers to the process of 
converting images of typed, handwritten or printed text 
into machine-readable text.
\subsubsection{Ancient Greek}
\label{sec:OCRancient}
We have identified three works about OCR over Ancient Greek.

\paper{ntzios2007old}{An old greek handwritten OCR system based on an efficient segmentation-free approach}
focus on the problem of recognizing old Greek manuscripts 
and propose a segmentation-free recognition technique tested in 
historical manuscript collections which are written in lowercase letters and originate from St. Catherine’s Mount Sinai Monastery.
\mycomment{
    Based on
    an open and closed cavity character representation, we
    propose a novel, segmentation-free, fast and efficient
    technique for the detection and recognition of characters and character ligatures. First, we detect open and
    closed cavities that exist in the skeletonized character
    body. Then, the classification of a specific character o
}

\paper{white2012training}{Training Tesseract for Ancient Greek OCR}   
describes a process for training the open-source Tesseract OCR engine\myfootnote{\url{https://github.com/tesseract-ocr/tesseract}} to support Ancient Greek. One of the challenges in this work is the recognition of the two types of diacritical marks of Ancient Greek, breathing marks and accents.

\mycomment{
    It covers the general procedures involved in training a new
    language for Tesseract, both training the script with common printed fonts and adding hints about how the language works to improve recognition. It discusses the particular challenges that arose with Ancient Greek, in the main due to Tesseract’s English language heritage. It goes on to describe the various strategies and small programs which were written to overcome these. It concludes with recommendations for changes to Tesseract to make OCR training easier and further improve recognition accuracy
}

\paper{robertson2017large}{Large-Scale Optical Character Recognition of Ancient Greek}
focus on large-scale optical character recognition of ancient (or polytonic) Greek;
the approach is based on the Gamera framework \myfootnote{\url{https://gamera.informatik.hsnr.de/index.html}},
and they report results from  processing  1,200 volumes comprising about 330 million Greek words.
\mycomment{
    This paper documents our campaign to undertake the large-scale optical character 
    recognition of ancient, or polytonic, Greek. 
    Building upon the Gamera OCR engine and developing a suite of post-processing tools, 
    including automatic spellcheck, we processed 1,200 volumes comprising 329,002,271 Greek words. 
    A sample of 10 pages is studied in detail; they demonstrate the degree to which each step of
    post-processing improved the results, and with which source documents. 
    These pages attain an average character accuracy of about 96
==}


\subsubsection{Modern Greek}
\label{sec:OCRmodern}

\paper{vamvakas2007greek}{Greek Handwritten character recognition}
and 
\paper{vamvakas2010handwritten}{Handwritten character recognition through two-stage foreground sub-sampling}
describe methods for Greek handwritten character recognition
evaluated over handwritten character and digit databases.
In the first one a hybrid feature extraction scheme was employed,
while in the second one
the extraction is based on recursive subdivisions of the character image so that the resulting sub-images 
at each iteration have balanced (approximately equal) numbers of foreground pixels.

  \mycomment{
    2010 paper:
    In this paper, we present a methodology for off-line handwritten character recognition. The proposed methodology relies on a new feature extraction technique based on recursive subdivisions of the character image so that the resulting sub-images at each iteration have balanced (approximately equal) numbers of foreground pixels, as far as this is possible. Feature extraction is followed by a two-stage classification scheme based on the level of granularity of the feature extraction method. Classes with high values in the confusion matrix are merged at a certain level and for each group of merged classes, granularity features from the level that best distinguishes them are employed. Two handwritten character databases (CEDAR and CIL) as well as two handwritten digit databases (MNIST and CEDAR) were used in order to demonstrate the effectiveness of the proposed technique. The recognition result achieved, in comparison to the ones reported in the literature, is the highest for the well-known CEDAR Character Database (94.73\%) and among the best for the MNIST Database (99.03\%)
    
    2007 paper:
    In this paper, we present a database and methods for off-line isolated Greek handwritten character recognition. The Computational Intelligence Laboratory (CIL) Database consists of 35,000 isolated and labelled Greek handwritten characters. This database was tested with an existing structural approach for Greek handwritten characters as well as with a novel approach based on a hybrid feature extraction scheme. According to this approach, two types of features are combined in a hybrid fashion. The first one divides the character image into a set of zones and calculates the density of the character pixels in each zone. In the second type of features, the area that is formed from the projections of the upper and lower as well as of the left and right character profiles is calculated. For the classification step, Support Vectors Machines (SVM) and Euclidean Minimum Distance Classifier (EMDC) are used. 
 }
 
\paper{simistira2015recognition}{Recognition of historical Greek polytonic scripts using LSTM networks} 
present an approach for the recognition of Greek polytonic scripts by training and testing a Long Short-Term Memory (LSTM) \citep{lstm1997} based recognizer 
using the OCRopus framework\myfootnote{\url{https://github.com/tmbarchive/ocropy}}. 

In this context, 
the authors created the publicly available  database Polyton-DB that extends the previous GRPOLY-DB~\citep{Gatos2015GRPOLYDB}  consisting of scanned pages only. The Polyton-DB  contains  15,689 textlines of synthetic and authentic printed  Greek  polytonic  script.

\mycomment{
    This paper reports on high-performance Optical Character Recognition (OCR) experiments using Long Short-Term Memory (LSTM) Networks for Greek polytonic script. Even though there are many Greek polytonic manuscripts, the digitization of such documents has not been widely applied, and very limited work has been done on the recognition of such scripts. We have collected a large number of diverse document pages of Greek polytonic scripts in a novel database, called Polyton-DB, containing 15; 689 textlines of synthetic and authentic printed scripts and performed baseline experiments using LSTM Networks. Evaluation results show that the character error rate obtained with LSTM varies from 5.51\% to 14.68\% (depending on the document) and is better than two well-known OCR engines, namely, Tesseract and ABBYY FineReader.
}

\paper{Sichani2019OCR}{OCR for Greek polytonic (multi accent) historical printed
documents: development, optimization and quality control} describe the design and  development of an OCR tool
for the recognition of Greek printed polytonic scripts. The proposed  framework  includes   the whole process including data gathering and structuring, OCR
tool development, user interface development, experiments on
the training procedure of the tool, evaluation, post-correction
and quality control of the results.

\subsubsection{Endangered dialects}

\paper{rijhwani2020ocr}{OCR Post Correction for Endangered Language Texts} develop an OCR post-correction method tailored for endangered dialects by adopting  a  sequence-to-sequence architecture. 
This work also introduces   a benchmark dataset  with transcribed images in three endangered languages,
namely Ainu, Griko, and Yakkh. Griko~\citep{Griko} is a Greek dialect spoken in the area of Southern Italy called Grecìa Salentina.
All the resources are available at \myfootnote{\url{https://shrutirij.github.io/ocr-el/}}.

\subsection{Phonetics/ Phonology}
\label{sec:Phonetics}

Phonetics is a branch of linguistics that studies the sounds of human speech. Phonology is the study of the organization and function of speech sounds in a given
system (language). Although the differences are well-established sometimes these terms are equated \citep{arvaniti2007greek}.

	The 1972 book~\citep{newton1972generative}
	studies 
	Greek phonology in general, while~\cite{trudgill2009greek} focuses on the Greek dialect vowel systems.~\cite{arvaniti2007greek} describes the 2007 state of the art in Greek phonetics.

\paper{epitropakis1993high}{Intonation for Greek TTS (Text To Speech)}
present an  algorithm for the generation  
of intonation (F0 contours) for the Greek Text-To-Speech system developed at Wire Communications Laboratory.
%
\paper{fourakis1999acoustic}{Acoustic characteristics of Greek vowels}  
analyze the acoustic characteristics of the Greek vowels
(duration, fundamental frequency, amplitude, and others).
Along the same line,
\cite{sfakianaki2002acoustic}
analyzes the acoustic characteristics of Greek vowels produced by adults and children.
\paper{Lyras2007Phonetics}{Learning Greek Phonetic Rules using Decision-Tree Based Models} 
apply a decision-tree based approach for learning Greek   phonetic rules. IPAGreek (\cite{themistocleous2011computational})
	is an  implementation (available at \cite{Themistocleous2017})
	of Standard Modern Greek and Cypriot Greek ``phonological grammar''. 
	The application enables users to transcribe text written in Greek orthography 
	into the International Phonetics Alphabet (IPA).

\paper{karanikolas2019machine}{Machine learning of phonetic transcription rules for Greek} 
proposes an automatic machine learning (ML)
approach that learns rules 
of how to transcribe Greek words into the International Phonetics Association’s (IPA’s) phonetic alphabet.
\mycomment{
    \paper{karanikolas2019machine}{Machine learning of phonetic transcription rules for Greek} 
    describes a machine learning (ML) approach that learns rules 
    of how to transcribe Greek words into the International Phonetics Association’s (IPA’s) phonetic alphabet. It mines rules of phonologic transcription from training data 
    (phonologically transcribed Greek words) that are available on the internet and in other resources.
==}
\paper{Themistocleous2019dialectclassification}{Dialect Classification From a Single Sonorant Sound Using Deep Neural Networks}
describes classification approaches based on deep neural networks for distinguishing two Greek dialects, 
namely Athenian Greek, the prototypical form of Standard Modern Greek and Cypriot Greek.
\mycomment{
    the first classification approach aims to distinguish two Greek dialects, namely Athenian Greek, the prototypical form of Standard Modern Greek and Cypriot Greek using measures of temporal and spectral information (i.e., spectral moments) from four sonorant consonants /m n l r/. The second classification study aims to distinguish the dialects using coarticulatory information (e.g., formants frequencies F1 − F5, F0, etc.) from the adjacent vowel in addition to spectral and temporal information from sonorants. In both classification approaches, they have employed Deep Neural Networks, which we compared with SVN, Random Forests, and Decision Trees.
}
Finally,
\paper{soundexGR2021}{SoundexGR: An Algorithm for Phonetic Matching for the Greek Language} propose a family of algorithms for phonetic matching for Greek text \myfootnote{\url{https://github.com/YannisTzitzikas/SoundexGR}}, for example the 4 characters  length phonemic transcription  
of
``\selectlanguage{greek}Αριστοτέλης\selectlanguage{english}''
is ``a8@4'',
while the full phonemic transcription is ``aristotelis''.


\subsection{Morphology}
\label{sec:Morphology}
Morphological parsing is the analysis of word structure and it is usually a prerequisite for  many text processing tasks. Indeed, some of the first efforts in processing Greek text were towards the creation of tools for this level of linguistic analysis~\citep{Packard1973,Crane1991Morpheus}.
Due to the rich morphology of Greek language, related tasks 
such the automated lemmatization
(i.e., the removal of inflectional suffixes for getting the base form of a word), 
stemming 
(i.e., the reduction of all the inflected forms of word into a common canonical form without the need to be meaningful), hyphenation (i.e., splitting a token over two lines with a hyphen symbol)
and spell checking, are challenging.


\subsubsection{Ancient Greek}

\mycomment{ to be excluded
\new{
\paper{Packard1973}{Computer-Assisted morphological analysis of ancient Greek} presents maybe the earliest automated morphological tagger for Ancient Greek. It was developed to aid American university students to learn Greek and specifically to overcome obstacles related to the complexity of Greek morphology. The tagger is written in Assembly language and provides lexical and grammatical analysis of 40,000 words of ancient Greek. For each text statistical summaries of the morphology are provided, as well as complete concordances organized both according to dictionary lemma and morphological category.
}
}

~\cite{Charalambous1992Hyphenation} presents a set of rules for the hyphenation of Ancient Greek 
taking into account characteristics such as the diversity of compound terms, declension, diacritics 
as well as special cases in which identical words have different meanings and different hyphenation. 
The resulting resource is incorporated in the ScholarTex package~\citep{Haralambous1991ScholarTeX}.

\paper{Lee2008nearest}{A nearest-neighbor approach to the automatic analysis of Ancient Greek Morphology} proposes a morphological analyzer that
is based on the \textit{nearest-neighbor machine learning framework}~\citep{Gutin2002} and does not require hand-crafted rules contrary to previous existing approaches such as in the classical works~\citep{Packard1973} and~\citep{Crane1991Morpheus}. It is also able to predict unseen words and to rerank its predictions by exploiting an unlabelled corpus of Ancient Greek.

\mycomment{
tabouratzis2007ancientpos
prior work of Tambouratzis
}

\paper{tambouratzis2016applying}{Applying particle swarm optimisation to the morphological segmentation of words from Ancient Greek texts}
investigates the effectiveness of \textit{evolutionary computation algorithms} 
in the morphological segmentation of words into subword
segments, focusing on the definition of stems and endings.

\mycomment{
The present article investigates the effectiveness of evolutionary computation algorithms in a
specific optimisation task, namely morphological segmentation of words into subword
segments, focusing on the definition of stems and endings. More precisely, particle swarm
optimisation (PSO) is compared to an earlier study on the same task using ant colony
optimisation (ACO), using a number of different optimisation criteria, for each of which
independent experiments are run. In the present article, the system architecture has been …
}

\paper{bary2017memory}{Lemmatizer for Ancient Greek} 
describe GLEM, 
a publicly available lemmatizer\myfootnote{\label{glem}\url{https://github.com/GreekPerspective/glem}} for Ancient Greek that uses POS information 
for disambiguation. GLEM is able to assign output to unseen words.

\paper{assael2019restoring}{Restoring ancient text using deep learning: a case study on Greek epigraphy}\myfootnote{\url{https://github.com/sommerschield/ancient-text-restoration}} 
focus on the recovery of missing characters from a damaged text input 
using \textit{deep neural networks}.

\paper{Assael2022}{Restoring and attributing ancient texts using deep neural networks}\myfootnote{\url{https://github.com/deepmind/ithaca}} 
is a close-related and follow-up work of~\citep{assael2019restoring} that utilizes  a deep neural network trained simultaneously to perform the tasks of textual restoration, geographical
attribution and chronological attribution for ancient Greek  inscriptions. 



\subsubsection{Modern Greek}

\paper{sgarbas1995pc}{A PC-KIMMO-based morphological description of Modern Greek} 
present the implementation of a morphological processor for the Modern Greek language 
that is based on the two-level morphology model  introduced by  \cite{koskenniemi1983two}. 
\mycomment{
    This paper presents the implementation of a morphological processor for the Modern Greek language. The processor is based on the two-level morphology model introduced by Koskenniemi (1983). The presented description is complete for all regular Greek nouns, adjectives, and verbs which constitute the basis of the entire Greek declension system. Modern Greek is a highly inflectional language. The thirty-six two-level rules and the structure of sixty-three declension categories presented in the Appendices have been developed using the PC-KIMMO environment and are compatible with PC-KIMMO version 2.0 software. Irregular forms do exist in Greek morphology but at present they are faced separately and they are not included in this implementation. Sample results and the base lines of the two-level model are also presented.
}

A rule-based hyphenator has been presented in \paper{Noussia1997Hyphen}{A Rule-Based Hyphenator for Modern Greek} that derives rules from grammar and phonology, while a few years later 
\paper{Filippou2004Tex}{Hyphenation Patterns for Ancient and Modern Greek} 
presents hyphenation patterns created for Ancient and Modern Greek to correct mistakes
or to complete patterns previously  
found on related  CTAN (Comprehensive TEX Archive Network)\myfootnote{https://www.ctan.org/} packages.

\paper{Vagelatosetal95Spelling}{Utilization of a lexicon for spelling correction in modern Greek} presents an interactive spelling correction system for Modern Greek. The system is based on a  morphological lexicon and emphasis  was given  
not only on  dictionary's coverage, but also on describing 
the rich inflectional morphology of Greek as economically as possible.

\paper{ntais2006development}{A Stemmer for the Greek Language}\myfootnote{\url{https://github.com/skroutz/greek_stemmer}}
describes a stemmer for the Greek language
that follows the known Porter algorithm for the English language 
and it is developed according to the grammatical rules of the Greek language.
Along the same line, \paper{papadakos2008anatomy}{The Anatomy of Mitos Web Search Engine}  
present 
a stemmer for the Greek language 
that follows again an affix removal approach (as the  Porter’s Algorithm).
\mycomment{
    By studying the Greek language grammar rules, a collection of common suffixes for nouns 
    and adjectives was gathered, including comparative, participle, singular and plural forms 
    in all genders and also verbs in different voices and tenses. 
    This collection was expanded using 270  ”productive suffixes” for words produced by other
    words with the same semantics, totaling 782 suffixes.
}
\paper{saroukos2009enhancing}{Greek Stemmer enchancements} 
and \cite{Ntais2015stemmercomparisoon} introduce another  stemmer\myfootnote{\label{saroukos}\url{http://saroukos.com/stemmer}} for the Greek language 
and compare it with previous implementations,
while
\paper{adam2010efficient}{Stemming and Tagging  for the Greek Language} 
provide a stemming and tagging procedure for the Greek language that can be easily adapted to existing systems.

\subsubsection{Endangered dialects}
\paper{Karanikolas2013amigre}{Structuring a Multimedia Tri-Dialectal Dictionary} 
present the design and the implementation process of a tri-dialectal dictionary for three Greek dialects in Asia Minor, 
namely Pontic, Cappadocian and Aivaliot.

\subsection{POS tagging and syntactic parsing}
\label{sec:Syntax}


Part-of-speech (POS) tagging (also called grammatical tagging) is the process of marking up a word in a text as corresponding to a particular part of speech, based on both its definition and its context. Syntax concerns the principles governing the arrangement of words in sentences, clauses and phrases as well as the relations among these parts \citep{Britanica2016Syntax}. The creation of computational grammars that adapt frameworks from theoretical linguistics 
and the part-of-speech tagging,
e.g., see an example of such analysis for the word 
``\selectlanguage{greek}είδε\selectlanguage{english}'' in the phrase of Figure \ref{fig:Example},
are the backbone tasks, as the study of the literature showed. 
Again the high degree of inflection of the Greek language  can add difficulty in computational approaches. 
It is indicative that a tagset for the Greek language, e.g., \myfootnote{\url{http://nlp.ilsp.gr/nlp/tagset_examples/tagset_en/index.html}}, 
can be many times     larger than that for English, e.g., Penn Treebank \citep{Taylor2003Penn}.

\subsubsection{Ancient Greek}
\label{sec:syntaxAncient}

 \paper{bamman2009ownership}{Dependency treebank for the Ancient Greek}
 describe the  first release of the Ancient Greek Dependency Treebank\myfootnote{\url{https://perseusdl.github.io/treebank_data/}} (AGDT), 
 a 190,903-word syntactically annotated corpus of literary texts including the works 
 of Hesiod, Homer and Aeschylus. 
\mycomment{
    While the far larger
    works of Hesiod and Homer (142,705 words) have been annotated under     a standard treebank production method of soliciting annotations from two   independent reviewers and then reconciling their differences, we also put
    forth with Aeschylus (48,198 words) a new model of treebank production   that draws on the methods of classical philology to take into account the personal responsibility of the annotator in the publication and ownership of a
    “scholarly” treebank.
=}

The thesis of  \paper{sorra2014anicentpos}{Design and development of a standard system of morphological analysis of names of the Ancient Greek language} describes the implementation of a rule-based POS tagger for Ancient Greek. 

\paper{celano2016part}{POS Tagging for Ancient Greek}
compare five  statistical POS taggers, i.e., the Mate tagger~\citep{Bohnet2010Mate}, the Hunpos tagger \citep{Hunpos2007}, RFTagger \citep{RFTagger2008}, the OpenNLP POS tagger, and the NLTK Unigram tagger. The data from Ancient Greek Dependency Treebank \citep{bamman2009ownership}
were used to train and compare the POS taggers. The results, with respect to the overall accuracy, showed that the Mate tagger performed better than any other tagger.

Another thesis \citep{liossis2019ancientpos} 
describes the implementation of a  POS-tagger and a lemmatizer  by employing traditional machine learning algorithms i.e., \textit{support vector machines (SVM)} \citep{cortes1995support}  and \textit{Conditional Ranfom Fields (CRF)}~\citep{CRF}.

\paper{gorman2020dependency}{Dependency Treebanks of Ancient Greek Prose}
describes a dataset comprising a collection of dependency syntax trees of 
representative texts from ancient Greek prose authors 
(Aeschines, Antiphon, Appian, Athenaeus, Demosthenes, Dionysius of Halicarnassus, Herodotus, Josephus, Lysias, Plutarch, Polybius, Thucydides, and Xenophon), 
totalling to date more than 550,000 tokens. 
\mycomment{
    It is hand-annotated by one person, using the
    Arethusa program on the Perseids website. Original texts were obtained from the Perseus
    Digital Library, and some (as indicated) were computer pre-parsed at the Pedalion Projec
    Original texts were obtained from the Perseus Digital Library, and some (as indicated) were computer pre-parsed at the Pedalion Project. The database is stored in a stable form (2019-12-31) on Zenodo (DOI: 10.5281/zenodo.3596076) and in a continuously updated form on GitHub in .xml format (https://vgorman1.github.io/). The repository can be used for pedagogical purposes and for research in linguistics analysis and corpus linguistics, stylistics, natural language processing, classification, and literary and historical analysis.
}

\paper{keersmaekers2019ancienttreebank}{Creating, Enriching and Valorizing Treebanks of Ancient Greek} 
present a ``roadmap'' on how different already available treebanks in Ancient Greek, 
namely Perseus Ancient Greek Dependency Treebanks, Aphthonius, Pedalion, Gorman, Sematia and PROIEL, 
can be combined to enhance the  progress in the automated parsing of 
classical and post-classical Greek texts.



\subsubsection{Modern Greek}
\label{sec:syntaxModern}

To facilitate the reader, 
below we first discuss the works related to POS tagging (which are numerous),
and then the rest works about syntax.
An overview of the Greek POS works is given
in Table~\ref{tbl:GreekPOSwork}.


\begin{itemize}[label={-},leftmargin=0.2cm, nosep]
\setlength\itemsep{0.0em}
\item {\textit{POS tagging}}
\end{itemize}
\paper{kotsanis1985lexifanis}{Lexifanis: A Lexical Analyzer of {M}odern {G}reek} 
describe a rule-based POS tagger by employing dictionaries of non-inflected words, affixation morphology and limited surface syntax rules. 

\paper{orphanos1999decision}{POS Tagging for the Greek language}
describe a machine learning approach for POS disambiguation and unknown word guessing.
Both problems are cast as classification tasks carried out by decision trees. 
\mycomment{
    The data model acquired is capable of capturing the idiosyncratic behavior of underlying linguistic phenomena. Decision trees are induced with three algorithms; the first two produce generalized trees, while the third produces binary trees. To meet the requirements of the linguistic datasets, all three algorithms are able to handle set-valued attributes. Evaluation results reveal a subtle differentiation
    in the performance of the three algorithms, which achieve an accuracy range of 93-95\% in POS
    disambiguation and 82-88\% in guessing the POS of unknown words
} 

In \paper{petasis1999resolving}{Resolving part-of-speech ambiguity in the Greek language using learning techniques} and 
in the PhD thesis \paper{2011Petasisphdthesis}{Machine learning in natural language processing} the use of \textit{transformation based error-driven learning} \citep{Brill1995algorithm} for resolving part-of-speech
ambiguity in the Greek language is investigated. 
\mycomment{
     This article investigates the use of \textit{transformation-based error-driven learning} \cite{Brill1995algorithm} for resolving part-of-speech
ambiguity in the Greek language. The aim is not only to study the performance, but also to examine its
dependence on different thematic domains. Results are presented here for two different test cases: a corpus
on ``management succession events'' and a general-theme corpus. The two experiments show that the
performance of this method does not depend on the thematic domain of the corpus, and its accuracy for the
Greek language is around 95%
}

\paper{Papageorgiou2000POS}{A Unified POS Tagging Architecture and its Application to Greek}
propose a unified POS tagging architecture that uses a feature-based multi-tiered approach to cover the requirements of tagging highly inflective languages such as Greek.
\mycomment{
    This paper proposes a flexible and unified tagging architecture that could be incorporated into a number of applications like information extraction, cross-language information retrieval, term extraction, or summarization, while providing an essential component for subsequent syntactic processing or lexicographical work. A feature-based multi-tiered approach (FBT tagger) is introduced to part-of-speech tagging. FBT is a variant of the well-known transformation based learning paradigm aiming at improving the quality of tagging highly inflective languages such as Greek. Additionally, a large experiment concerning the Greek language is conducted and results are presented for a variety of text genres, including financial reports, newswires, press releases and technical manuals. Finally, the adopted evaluation methodology is discussed.
}

\paper{Kontos2000arista}{ARISTA Generative Lexicon for Compound Greek Medical Terms} propose an approach based on rules written in \textit{Prolog} language  for the generation of a  lexicon of Greek compound medical terminology. 
A similar approach is followed in \paper{Kontos2000stock}{Greek Verb Semantic Processing for Stock Market Text Mining} for the construction of a computational lexicon of Greek verbs for the stock market domain.


In \paper{Maragoudakis2003POS}{A Bayesian Part-Of-Speech and Case Tagger for Modern Greek} a Bayesian  network  probabilistic  model is introduced for POS tagging in Greek texts.

\paper{kuzmenko2015disambiguation}{Automatic Disambiguation in the Corpora of Modern Greek and Yiddish}
present a method for morphological disambiguation in the absence of any gold standard corpora using several statistical models, such as the Brill algorithm \cite{Brill1995algorithm}. The methods were tested both on Greek and Yiddish.

\paper{partalidou2019design}{Open source Greek POS Tagger and Entity Recognizer}
describe a machine learning approach to POS tagging and named entity recognition for Greek, focusing on the extraction of morphological features and classification of tokens
into a small set of classes, i.e., location, organization, person and facility for named entities. 

\paper{nikiforos-kermanidis-2020-supervised}{Part-Of-Speech Tagger for the Greek Language of the Social Web}
describe a part-of-speech tagged dataset of social text in Greek that contains 31,697 tokens originated from 2,405 tweets. In addition, a supervised part-of-speech tagger based on \textit{naive Bayes} algorithm and specialized for such datasets, is presented. 

\ \\

\begin{itemize}[label={-},leftmargin=0.2cm, nosep]
\setlength\itemsep{0.0em}
\item {\textit{Grammar}}
\end{itemize}

\paper{Kordoni2004HPSG}{Deep Analysis of Modern Greek} 
present a 
computational Modern Greek grammar written in Head-Driven Phrase Structure Grammar (HPSG) \citep{PollardSag94HPSG}.

\paper{prokopidis2005theoretical} {Theoretical and practical issues in the construction of a Greek dependency treebank}
describe the construction of the Greek Dependency Treebank
while \paper{prokopidis2017universal}{} describe the work towards the harmonization of the Greek Dependency Treebank\myfootnote{\label{gdt} \url{http://gdt.ilsp.gr} and \url{https://universaldependencies.org/treebanks/el\_gdt/index.html}}
with the Universal Dependencies v2 standard, and the extension of the treebank with enhanced dependencies.\\ 
\paper{Kermanidis2009Paraphrasing}{Empirical Paraphrasing of Modern Greek Text in Two Phases: An Application to Steganography} 
describe the application of paraphrasing to steganography. For the paraphrasing step a set of shallow empirical rules are applied followed by a filtering process through a supervised machine learning approach. The syntactic transformations are shallow and require minimal linguistic resources (e.g., [I went] [to the excursion] [yesterday].$->$ [I went]  [yesterday] [to the excursion].)  

\paper{papadopoulou-2013-gf}{Proceedings of the Student Research Workshop associated with RANLP 2013, pages 126–133,Hissar, Bulgaria, 9-11 September 2013. GF Modern Greek Resource Grammar} describes
an implementation of a Modern  Greek  grammar in     Grammatical Framework  as  part  of  the  Grammatical Framework     Resource     Grammar     Library \citep{gf2011}. 

\paper{samaridi2014parsing}{Parsing Modern Greek verb MWEs with LFG/XLE grammars}
 present an approach to integrate verb MWEs
 (Verb Multiword Expressions) 
 in an LFG grammar of
 Modern Greek 
 and
 IDION \cite{markantonatou-etal-2019-idion}
 \myfootnote{
    \url{idion.ilsp.gr/data}}
that
 contains about 2,000 verb MWEs (VMWEs) of which about 850 are fully documented 
 as regards their syntactic flexibility, their semantics and the semantic relations with other VMWEs.


\mycomment{== From the abstract==
     on-going effort to
    integrate verb MWEs in an LFG grammar of
    Modern Greek (MG). Text is lemmatized and
    tagged with the ILSP FBT Tagger and is fed
    to a MWE filter that marks
    Words_With_Spaces in MWEs. The output is
    then formatted to feed an LFG/XLE grammar
    that has been developed independently. So far
    we have identified and classified about 2500
    MWEs, and have processed 40
    manipulating only the lexicon and not the
    rules of the grammar. 
==}

\paper{gakis2017design}{Design and construction of the Greek grammar checker}
present a tool for  analyzing morphologically and syntactically sentences, phrases, and words in order to correct syntactic, grammatical, and stylistic errors. The proposed approach takes into account some of the particularities of Modern Greek such as its highly inflectional nature, the ``free word order'' form and the lexical ambiguity.


\subsubsection{Endangered dialects}
\paper{anastasopoulos2018POSgriko}{Part-of-Speech Tagging on an Endangered Language: a Parallel Griko-Italian Resource} 
evaluate POS tagging techniques on an actual endangered language, Griko. 
    The combination of a semi-supervised method with cross-lingual transfer seems to be the more appropriate for this task  given the small size of the used corpus.


\begin{table}[htb!]
\caption{An overview of Greek POS work. In the second column ``ell" stands for Modern Greek,
            while ``grc" for Ancient Greek.}
\label{tbl:GreekPOSwork}
{\begin{minipage}{25pc}
\begin{tabular}{@{\extracolsep{\fill}}p{3.5cm}p{1cm}p{2cm}p{3cm}p{1cm}}
\hline


Work &Lang.                 &Approach & Evaluation data & Section\\ \hline
\citep{kotsanis1985lexifanis} & ell & rules & - & [~\ref{sec:Syntax} ]  \\  
\citep{orphanos1999decision} & ell & supervised &  137,765
tokens (7,624 sentences) from various sources i.e., student writings, literature, newspapers,
and technical, financial and sports magazine & [~\ref{sec:Syntax} ]  \\ 

\citep{Papageorgiou2000POS} & ell & supervised & 210 files from different genres of texts (447K tokens)  & [~\ref{sec:Syntax} ]  \\ 
\citep{petasis2001greek} & ell & supervised & 15,990 tokens & [~\ref{sec:Syntax} ] \\ 
\citep{Maragoudakis2003POS} & ell & supervised &  137,765
tokens (7,624 sentences) from various sources i.e., student writings, literature, newspapers,
and technical, financial and sports magazine & [~\ref{sec:Syntax} ]  \\ 
\citep{sorra2014anicentpos} & grc & rules & Perseus & [~\ref{sec:Syntax} ]  \\ 
\citep{celano2016part} & grc & supervised & Perseus  &[~\ref{sec:Syntax} ] \\ 
\citep{anastasopoulos2018POSgriko} & grico & a. semi-supervised method with cross-lingual transfer and b. active learning& 114 narratives&[~\ref{sec:Syntax} ] \\ 

\citep{partalidou2019design}& ell & machine learning & news articles & [~\ref{sec:Syntax} ] \\ 
\citep{liossis2019ancientpos}& grc & supervised & Perseus (except from Doric dialect) & [~\ref{sec:Syntax} ] \\ 
\citep{nikiforos-kermanidis-2020-supervised}& ell & supervised & 2,578 Greek tweets (31,697 tokens) & [~\ref{sec:Syntax} ] \\ 
\citep{Koutsikakis2020bert}& ell & supervised  & 2,521 Greek sentences from the Universal Dependencies Treebank & [~\ref{sec:Embeddings}]\\ 
\citep{Singh2021ancient-greek-bert}& grc & supervised  & The (Ancient Greek part of the) Perseus
Digital Library and The First1KGreek Project
Database &[~\ref{sec:Embeddings} ]\\

\hline
    \end{tabular}
  \end{minipage}}
\label{GreekPOSwork}
\end{table}

\subsection{Word embeddings}
\label{sec:Embeddings}
A word embedding method
generally tries 
to map words, phrases or documents,
to vectors of real numbers so that semantic, syntactic and world knowledge 
can be captured and reflected in similar vectors based on \textit{distributional hypothesis} \citep{harris54DH}. 
To grasp the idea,
the word   \selectlanguage{greek}τηλεσκόπιο\selectlanguage{english} of our running example,
according to the FastText algorithm~\citep{bojanowski2016enriching} trained on a Greek corpus,
is represented by 
a vector of 300 float numbers
of the form $\langle 0.74044, -0.070209, \ldots, 0.23586, 0.48267\rangle$.


\mycomment{nomizo oti prepei na mi milisoume mono gia simasiologiki pliroforia na valoume oti word embeddings mporei na deixoun kai se alla epipeda opos tis sintaxis. Epomenos poli sosta einai kai se autonomi katigoria.

old:
A Word Embedding method
generally tries 
to map words, phrases or event sentences to vectors of real numbers so that 
semantically similar  words (or phrases)
are mapped to 
similar vectors.
}

\paper{outsios2018word}{Word Embeddings from Large-Scale Greek Web Content}
present word embeddings\myfootnote{\label{gwe}\url{http://nlp.polytechnique.fr/resources-greek}} and other linguistic resources trained on a 50 GB corpus. The corpus was created by crawling the Greek internet web.

\paper{lioudakis2019ensemble}{An Ensemble Method for Producing Word Representations for the Greek Language }
present an ensemble method, named Continuous Bag-of-Skip-grams (CBOS),
for producing  word representations for the Greek language. 

\paper{outsios2019evaluation}{Evaluation of Greek Word Embeddings}
describe a Greek version of WordSim353 test collection \citep{Finkelstein2002WordSim} for a basic evaluation 
of word similarities, and the testing 
of seven word vector models.

\paper{Koutsikakis2020bert}{GREEK-BERT: The Greeks Visiting Sesame Street}\myfootnote{\label{greekBERT}
    \url{https://github.com/nlpaueb/greek-bert}
} present a Greek edition of the BERT \citep{devlin2018bert} pre-trained on large Greek corpora, i.e., the Greek part of Wikipedia, the Greek part of European Parliament Proceedings Parallel Corpus\myfootnote{\url{https://www.statmt.org/europarl/}}, the Greek part of OSCAR\myfootnote{\url{https://oscar-corpus.com/}}, a cleansed version of Common Crawl.

One year later, \paper{Singh2021ancient-greek-bert}{A Pilot Study for BERT Language Modelling and Morphological Analysis for Ancient and Medieval Greek}
released an Ancient Greek version of BERT. The model was initialised from \cite{Koutsikakis2020bert} model and subsequently trained on monolingual Ancient Greek data.





Since embeddings 
are important in various tasks and several algorithms have been proposed in the literature \citep{Embeddings2020Background},
Tables~\ref{embedding} and~\ref{ancientembedding}
in Section \ref{sec:CorporaDatasets}
list and provide more details
about the available (monolingual) embeddings in Modern and Ancient Greek respectively.

\subsection{Semantics}
\label{sec:Semantics}

The semantic processing
usually 
considers
a representation of sentence structure (syntactic tree),
a representation of the possible meaning(s) of each word,
and aims at producing 
a representation of the meaning of the sentence.
This process includes steps like
word sense disambiguation,
semantic role tagging, 
named entity recognition and linking,
and other tasks. Semantic role labeling aims at mapping spans in sentences to a predicate-argument structure. The structure describes  eventualities and participants in the sentence and is often described as answering ``Who did what to whom'' question.
Named entity recognition aims at identifying named entities,
e.g. ``\selectlanguage{greek}Αριστοτέλης\selectlanguage{english}''
in the phrase of Figure 	\ref{fig:Example}. Named entity linking (NEL), also referred as entity linking  is the task of semantically annotating entities mentioned in text with entities described in knowledge
bases (KBs).



\subsubsection{Ancient Greek}
\label{sec:SemanticsAncient}

\paper{celano2015semantic}{Semantic role annotation in the ancient greek dependency treebank}
 present an annotation scheme for {\em semantic role} annotation for the Ancient Greek Dependency Treebank.

\paper{perrone2019gasc}{Genre-aware semantic change (GASC) for Ancient Greek} 
describe a \textit{Bayesian model} to study dynamically the {\em semantic change of word senses} in ancient text that leverages categorical metadata about the texts' genre to boost inference and uncover the evolution of meanings in Ancient Greek corpora.

\paper{rodda2019vector}{Vector space models of Ancient Greek word meaning,  and a case study on Homer} 
describe a  vector space model where every word is represented by a vector which encodes information about its linguistic context(s).

\paper{McGillivray2019Polysemy}{A computational approach to lexical polysemy in Ancient Greek} aim at computationally modelling Ancient Greek {\em diachronic semantics} based on Bayesian learning.

\mycomment{
    antics, we observe that some words have several meanings, thus displaying lexical polysemy. In this article, we present the first phase of a project that aims at computationally modelling Ancient Greek semantics over time. Our system is based on Bayesian learning and on the Diorisis Ancient Greek corpus, which we have built for this purpose. We illustrate preliminary results in
}

\paper{keersmaekers-2020-automatic}{Automatic semantic role labeling in Ancient Greek using distributional semantic modeling} 
provides an approach for automatic {\em role labeling} in Ancient Greek based on a \textit{random forest} classifier \citep{Breiman2001}. A small semantically annotated corpus of Ancient Greek was used, annotated with a large amount of linguistic features, including form of the construction, morphology, part-of-speech, lemmas, animacy, syntax and distributional vectors of Greek words. Overall, the distributional semantic vectors features proved more important in the model.

\paper{Palladino2020}{NER on Ancient Greek with minimal annotation} 
present a NER model for the recognition of  place-names and ethnonyms trained on a small manually annotated list. Their approach use \textit{Conditional Ranfom Fields} and
relies on the recurring elements in the context.



\subsubsection{Modern Greek}
\label{sec:SemanticsModern}

Table~\ref{tbl:GreeNERtools} provides an overview of works on Named Entity Recognition.

\paper{karkaletsis1999named}{Named-entity recognition from Greek and English texts}
    describe a prototype NER (Named Entity Recognition) system for Greek texts 
    developed based on a NER system for English. 
    Both systems are evaluated on corpora of the same domain and of similar size. 
    \mycomment{
        Named-entity recognition (NER) involves the identification and classification of named
        entities in text. This is an important subtask in most language engineering applications, in particular
        information extraction, where different types of named entity are associated with specific roles in
        events. In this paper, we present a prototype NER system for Greek texts that we developed based on
        a NER system for English. Both systems are evaluated on corpora of the same domain and of similar
        size. The time-consuming process for the construction and update of domain-specific resources in
        both systems led us to examine a machine learning method for the automatic construction of such
        resources for a particular application in a specific language.
    }

\paper{demiros2000named}{Named Entity Recognition in Greek Texts}
describe a named entity recognizer for Greek
based on pattern matching  and non-recursive regular expressions.
\mycomment{
    The system aims at
    information extraction applications where large scale text processing is needed. Speed of analysis, system robustness, and results accuracy have been the basic guidelines for the system’s design. Our system is an automated pipeline of linguistic components for Greek text processing based on pattern matching techniques. Non-recursive regular expressions have been implemented on top of it in order to capture different types of named entities. For development and testing purposes, we collected a corpus of financial texts from
    several web sources and manually annotated part of it. Overall precision and recall are 86\% and 81\% respectively. 
}

\paper{Papageorgiou2004Multimodal}{CIMWOS: A Multimedia Retrieval System Based on Combined Text, Speech and Image Processing} present a multimedia, multimodal and multilingual retrieval system.  In terms of text processing, for the text produced by the speech processing subsystem   named entity detection, term recognition, story segmentation, and topic classification is performed. For the NER task  a finite state parser on the basis of a pattern grammar is employed.

%


\paper{lucarelli2007named}{Named entity recognition in greek texts with an ensemble of svms and active learning}
describe a freely available named-entity recognizer\myfootnote{\label{aueb}\url{http://nlp.cs.aueb.gr/software.html}} for Greek texts that identifies temporal expressions, person, and organization names. An ensemble of \textit{support vector machines} and \textit{active learning} was used for person and organization names while for the recognition of temporal expressions semi-automatically produced patterns.
\mycomment{
    We present a freely available named-entity recognizer for Greek texts that identifies temporal expressions, person, and organization names. For temporal expressions, it relies on semi-automatically produced patterns. For person and organization names, it employs an ensemble of Support Vector Machines that scan the input text in two passes. The ensemble is trained using active learning, whereby the system itself proposes candidate
    training instances to be annotated by a human during training. The recognizer was evaluated on both a general collection of newspaper articles and a more focussed, in terms of topics, collection of financial articles.
}

\paper{Vonitsanou2011keyword}{Keyword Identification within {G}reek URLs} 
introduce a keyword extraction technique focusing on Greek URLs by combining a transliteration engine and a tokenizer.

\paper{angelidis2018named}{Named Entity Recognition, Linking and Generation for Greek Legislation} 
describe a method for named entity recognition in Greek legislation using \textit{deep neural network} architectures. 
The recognized entities are used to enrich the Greek legislation knowledge graph with more detailed information about persons, organizations, geopolitical entities, legislation references, geographical
landmarks and public document references.

\paper{Bartziokas2020}{Datasets and Performance Metrics for Greek Named Entity Recognition} 
introduce a dataset focused on named entity recognition and further describe the experimentation with various state-of-the-art models that are trained on this dataset.

\paper{Papantoniou2021nel}{{EL-NEL:} Entity Linking for Greek News Articles} 
describe a pipeline for entity linking for Greek news articles. The pipeline consists of a BERT-based named entity recognition step, a candidate generation step from a multitude of wiki-based KBs (Wikidata, DBpedia, Wikipedia,
YAGO) and finally a disambiguation step based on context and no context dependent  embeddings.

\mycomment{
    We also interlink the textual references of the recognized entities to the corresponding entities represented in other open public datasets and, in this way, we enable new sophisticated ways of querying
    
    Greek legislation. Relying on the results of the aforementioned methods we generate
    and publish a new dataset of geographical landmarks mentioned in Greek legislation.
    We make available publicly all datasets and other resources used in our
    study. Our work is the first of its kind for the Greek language in such an extended
    form and one of the few that examines legal text in a full spectrum, for both entity
    recognition and linking.
==}



\begin{table}[htb!]
\caption{An overview of Greek NER work. Entity type refer to the category of the identified entity e.g., Person, Organization, Location etc. Double value at the same cell mean that two annotation schemes have been examined.}
\label{tbl:GreeNERtools}
{\begin{minipage}{25pc}
\begin{tabular}{@{\extracolsep{\fill}}p{3.5cm}p{5cm}p{2cm}p{2cm}}
\hline
Work                              &Domain                 &Num. of entity types & Section\\ \hline
\citep{karkaletsis1999named}& news articles from the Greek company “Advertising Week” & 2  & [~\ref{sec:Semantics} ] \\

\citep{demiros2000named} & financial & 7  & [~\ref{sec:Semantics} ] \\

\citep{Papageorgiou2004Multimodal} & open domain transripts & 3&
[~\ref{sec:Semantics} ] \\ 

\citep{lucarelli2007named} & general newspaper
articles \& financial & 3  & [~\ref{sec:Semantics} ] \\ 
\citep{angelidis2018named} & legal & 6 & [~\ref{sec:Semantics} ]\\ 
\citep{partalidou2019design} & Wikipedia \& news articles & 3 &[~\ref{sec:Semantics} ]\\

\citep{Koutsikakis2020bert} &  news articles (4,189 sentences) & 3 & [~\ref{sec:Embeddings} ] \\ 
\citep{Bartziokas2020} &  news articles & 4 \& 18  & [~\ref{sec:Semantics} ] \\
\citep{Palladino2020} & Herodotus’ Histories & 2 & [~\ref{sec:Semantics} ] \\ 
\citep{Papantoniou2021nel} &  news articles & 4 \& 18 &[~\ref{sec:Semantics} ]\\ 
%

 \hline
    \end{tabular}
  \end{minipage}}
\label{overviewNERwork}
\end{table}

%


As regards semantic relationships,
\paper{Christou2021Katharevousa}{Extracting Semantic Relationships in Greek Literary Texts} propose a distantly supervised transformer-based~\citep{NIPS2017_3f5ee243} relation extraction model 
for semantic relationships extraction in literary texts of the 19th century Greek Literature.


\subsection{Pragmatics (Dialogue systems)}
\label{sec:Pragmatics}
Pragmatic analysis attempts to put each sentence into its general situational context, 
taking into account the contexts in which it is said. Pragmatics encompasses speech act theory, conversational implicature talk and other approaches. Unlike semantics that examines meaning  as coded in a given language, pragmatics studies how the transmission of meaning depends not only on structural and linguistic knowledge of the speaker and listener but also on the context of the utterance~\citep{Levinson1983,Mao1995PragmaticsAI}.

This process includes various (and challenging) tasks 
including:
resolution of  references (e.g., pronouns),
resolution of semantic ambiguities,
inferring the  missing objects and actions,
drawing inferences based on world knowledge and common scenarios and scripts,
linking sentences,
identifying speech acts,
understanding discussions and dialogues.
For example, 
from the perspective of Speech Acts Theory,
the phrase of Figure \ref{fig:Example} 
would be classified as Assertive.

\subsubsection{Ancient Greek}
\label{sec:pragmaticsAncient}
\paper{rydberg2011social}{Social Networks and the Language of Greek Tragedy}
uses linguistic dependency treebanks and digitized texts created by the Perseus Digital Library, 
to create social networks for a collection of Greek tragedies that allow users to visualize 
the interactions between characters in the plays.
\mycomment{
    Using the linguistic dependency treebanks and digitized texts created by the Perseus Digital Library,
    we are creating social networks for a collection of Greek tragedies that allow users to visualize the
    interactions between characters in the plays. Because the number of characters who appear on stage
    in Greek tragedy is limited, most of these social network diagrams fall into a few basic types. The
    most interesting aspect of these networks are, therefore, the edges that connect the nodes within the
    graphs. The linguistic data used to label or even create these edges becomes the jumping off point
    for visualizing and exploring the language of Greek tragedy
}

\subsubsection{Modern Greek}
\label{sec:pragmaticsModern}
\paper{Chondrogianni2011Pragmatics}{Identifying the User's Intentions: Basic Illocutions in Modern Greek} presents
a classification of propositional basic illocutions (e.g., requests, wondering, curses etc.) in Modern Greek extracted  following  the linguistic  choices  speakers  make  when  they  formulate  an utterance. 
The emphasis has been given on function of the illocutions rather than form.

The thesis of \paper{karamitsos2019chatbots}{Chatbots for Greek/EU Public Services} 
 describes a conversational chatbot system based on public services for a 
 greek web portal called {\tt{diadikasies.gr}} to help citizens find easily the desired service. 
 
``Milisteros'' is a chatbot, described in the  thesis of  \paper{tataridis2019chatbot}{Chatbot development using Python/ChatterBot technology}, that employs a library in Python \myfootnote{\url{https://github.com/gunthercox/chatterbot}} 
to provide essential information about the Aristotle University of Thessaloniki to freshmen.

 \paper{Ventoura2021TheanoAG}{Theano: A Greek-speaking conversational agent for COVID-19}
  present  Theano, a converational agent for COVID-19 information dissemination and symptom self-checking in
Greek.  Theano is implemented using the Rasa
dialogue framework~\cite{Bocklisch2017Rasa} and the code is available as open source at \myfootnote{\url{https://gitlab.com/ilsp-spmd-all/public/covid-va-chatbot}
 }.
 
\paper{Antoniadis2021PassBot}{PassBot: A Chatbot for Providing Information on Getting a Greek Passport}
  present PassBot a chatbot with the aim to provide personalized information to  citizens interested in  getting a Greek passport. Rasa was also used for this chatbot and the code is available at \myfootnote{\url{https://github.com/PantelisAntoniadis/PassBot_project}
 }.

\subsection{Sentiment analysis}
\label{sec:SA}
Sentiment analysis is the process of computationally identifying opinions, emotions and attitudes expressed in a piece of text towards a particular topic, product, etc. For example, a tool for sentiment analysis  would characterize  the phrase of 
Figure     \ref{fig:Example} as \textit{neutral}.
\mycomment{
I think the previous was focused more on opinions
}

\paper{markopoulos2015sentiment}{Sentiment Analysis of Hotel Reviews in Greek: A Comparison of Unigram Features} 
implement a sentiment  classifier applying {\em{SVM}} on hotel reviews written in Modern Greek. The corpus of hotel reviews was collected from the Greek version of TripAdvisor and consists of 1,800 reviews (900 positive and 900 negative). A manually annotated sentiment lexicon also created in the context of this work that comprises a total of 27,388 types of positive words and 41,410 types of negative words.
\mycomment{
    minor work can be excluded
}

\paper{athanasopoulou2016schizophrenia}{'Schizophrenia'on Twitter: Content Analysis of Greek Language Tweets}
analyzes the sentiment of tweets related to schizophrenia in Greek language, in comparison with other illness (diabetes).
 \mycomment{
    Twitter is an online space whose users can create and share ideas and information instantly. The term schizophrenia is frequently used in a stigmatizing way in Greek language. In Greece, Twitter is the tenth most popular website. Tweets related to schizophrenia in Greek language, have not been investigated. We aimed to examine schizophrenia Tweets in comparison with other illness (diabetes). Deductive content analysis was applied. Schizophrenia Tweets (n=239), tended to be more negative, medically inappropriate, sarcastic, and used non-medically than diabetes Tweets (n=205). Our findings confirm the frequent, non-medical misuse of the term 'schizophrenia' in online sources written in Greek language. These results show that mental health  interventions are needed to raise awareness among the general population, in order to eliminate stigmatizing behaviors. Future anti-stigma actions, could also raise awareness among internet users about the importance of, avoiding using medical terms in negative or sarcastic ways, and eliminate any potential stigmatizing content.
}

\paper{pavlopoulos2017abusiveingreek}{Deep Learning for User Comment Moderation} 
describe  methodologies for both fully automatic and semi-automatic moderation 
in user comments. They propose the use of a gating mechanism (GRU) in \textit{recurrent neural networks} (RNN) on word embeddings. The results has been evaluated both on English and Greek datasets.

\paper{spatiotis2017examining}{Examining the Impact of Feature Selection on Sentiment Analysis for the Greek Language} 
present an approach to recognize opinions in Greek language and examines the impact of feature selection on the analysis of opinions and the performance of the classifiers
(text-based and POS-based features from textual data are extracted).
\mycomment{
    Sentiment analysis identifies the attitude that a person has towards a service, a topic or an event and it is very useful for companies which receive many written opinions. Research studies have shown that the determination of sentiment in written text can be accurately determined through text and part of speech features. In this paper, we present an approach to recognize opinions in Greek language and we examine the impact of feature selection on the analysis of opinions and the performance of the classifiers. We analyze a large number of feedback and comments from teachers towards e-learning, life-long courses that have attended with the aim to specify their opinions. A number of text-based and part of speech based features from textual data are extracted and a generic approach to analyze text and determine opinion is presented. Evaluation results indicate that the approach illustrated is accurate in specifying opinions in Greek text and also sheds light on the effect that various features have on the classification performance. 
}

\paper{athanasiou2017novel}{A gradient boosting framework for sentiment analysis in languages where NLP resources are not plentiful; 
a case study for modern Greek} focus on sentiment analysis  applied in Greek texts,
 specifically it investigates a process where 
 \textit{gradient boosting}, a technique for dealing with high-dimensional data,
 is applied on Greek texts.
 
\mycomment{
    Sentiment analysis has played a primary role in text classification. It is an undoubted fact    that some years ago, textual information was spreading in manageable rates; however, nowadays,    such information has overcome even the most ambiguous expectations and constantly grows within    seconds. It is therefore quite complex to cope with the vast amount of textual data particularly if we    also take the incremental production speed into account. Social media, e-commerce, news articles,    comments and opinions are broadcasted on a daily basis. A rational solution, in order to handle    the abundance of data, would be to build automated information processing systems, for analyzing    and extracting meaningful patterns from text. The present paper focuses on sentiment analysis    applied in Greek texts. Thus far, there is no wide availability of natural language processing tools for Modern Greek. Hence, a thorough analysis of Greek, from the lexical to the syntactical level, is difficult to perform. This paper attempts a different approach, based on the proven capabilities
    of gradient boosting, a well-known technique for dealing with high-dimensional data. The main    rationale is that since English has dominated the area of preprocessing tools and there are also quite    reliable translation services, we could exploit them to transform Greek tokens into English, thus    assuring the precision of the translation, since the translation of large texts is not always reliable    and meaningful. The new feature set of English tokens is augmented with the original set of Greek,    consequently producing a high dimensional dataset that poses certain difficulties for any traditional    classifier. Accordingly, we apply gradient boosting machines, an ensemble algorithm that can learn    with different loss functions providing the ability to work efficiently with high dimensional data.    Moreover, for the task at hand, we deal with a class imbalance issues since the distribution of    sentiments in real-world applications often displays issues of inequality. For example, in political
    forums or electronic discussions about immigration or religion, negative comments overwhelm the    positive ones. The class imbalance problem was confronted using a hybrid technique that performs a    variation of under-sampling the majority class and over-sampling the minority class, respectively.    Experimental results, considering different settings, such as translation of tokens against translation    of sentences, consideration of limited Greek text preprocessing and omission of the translation    phase, demonstrated that the proposed gradient boosting framework can effectively cope with both    high-dimensional and imbalanced datasets and performs significantly better than a plethora of
    traditional machine learning classification approaches in terms of precision and recall measures.
}

\paper{ferra2017tale}{Polarization and mobilization for the Greek referendum on Twitter}
examine the hashtag \#greferendum, focusing on both social and semantic networks 
(\#Grexit, \#oxi campaign), 
and it analyses Twitter data, which were collected using NodeXL, on three significant days: 
the announcement of the referendum, the day of the bailout expiration and the actual date of the referendum. 
\mycomment{
    This chapter examines the hashtag #greferendum, focusing on both social and semantic networks (#Grexit, #oxi campaign), and it analyses Twitter data, which collected using NodeXL, on three significant days: the announcement of the referendum, the day of the bailout expiration and the actual date of the referendum. Data analysis draws on cyberconflict theory (Karatzogianni, 2006, 2015), to situate the Greek Referendum in the wider sociopolitical context of anti-austerity mobilizations in Greece and discusses the contribution of Twitter in the formation, polarization and mobilization within the context of a transnational networked public sphere.
}

\paper{tsakalidis2018nowcasting}{Nowcasting the stance of social media users in a sudden vote: The case of the Greek Referendum}
focus on  the 2015 Greek bailout referendum, aiming to nowcast  on a daily basis the voting intention of 2,197 Twitter users. 
It proposes a semi-supervised multiple convolution kernel learning   approach, leveraging temporally sensitive text and network information.
\mycomment{
    Modelling user voting intention in social media is an important     research area, with applications in analysing electorate behaviour,     online political campaigning and advertising. Previous approaches
    mainly focus on predicting national general elections, which are     regularly scheduled and where data of past results and opinion     polls are available. However, there is no evidence of how such models would perform during a sudden vote under time-constrained circumstances. That poses a more challenging task compared to     traditional elections, due to its spontaneous nature. In this paper,     we focus on the 2015 Greek bailout referendum, aiming to nowcast    on a daily basis the voting intention of 2,197 Twitter users. We    propose a semi-supervised multiple convolution kernel learning
    approach, leveraging temporally sensitive text and network infor    mation. Our evaluation under a real-time simulation framework    demonstrates the effectiveness and robustness of our approach    against competitive baselines, achieving a significant 20\% increase
    in F-score compared to solely text-based models
}

\paper{boukala2018absurdity}{Absurdity and the “blame game” within the Schengen area -  Analyzing Greek (social) media discourses on the refugee crisis} 
utilize Discourse-Historical Approach and content analysis for the analysis of Greek social media  discourses on the refugee crisis.

\mycomment{
    o explore whether and how the Greek media and social media discourses on the refugeecrisis and the political decisions regarding it contribute tothe discursive reconstruction of the Greek nation-state and its imaginaries. We assume that the refugee crisis led to a polarized climate that dominates the Greek and Europea political scene and questions European solidarity. Utilizing the Discourse-Historical Approach, we analyze the social media discourses of the Greek prime minister and the president  of  the  main  opposition  party  regarding  the European debate on the Schengen agreement’s suspension.Moreover, we employ content analysis to focus on the media coverage of the Schengen debate and the refugee crisis.Finally, based on Camus’s allegoric novel ThePlague,we emphasize  the  parallelism  between  the  discourses  on the Schengen agreement and the allegory of the plague.“Fortress Europe”and the emphasis on national border swithin the European Union prompted  us to utilize the allegory of Camus’sThe Plague to highlight the resurgence of the nation-state and the discursive deconstruction of European solidarity in times of crisis
}


In \paper{lekea2018hate}{Detecting Hate Speech within the Terrorist Argument: A Greek Case} by employing a dataset consisting of all the documents that the members of the terrorist organization ``Revolutionary Organization 17 November'' (17N) have published, a rule-based categorization task is performed to discriminate between three categories: ``no  hate  speech'', ``moderate hate speech'', and ``apparent hate speech''. Different text analyzing techniques such as \textit{critical discourse} and \textit{content analysis} were preformed to decide the most effective parameters and their thresholds.

\mycomment{how hate speech was used as a means of justification. }

\paper{tsakalidis2018sentiment}{Building and evaluating resources for sentiment analysis in the Greek language}
present a survey for sentiment analysis tools and resources in Greek, and makes publicly available
a rich set of such resources\myfootnote{\label{sentimentlexicon}\url{https://github.com/MKLab-ITI/greek-sentiment-lexicon}}, 
ranging from a manually annotated lexicon, 
to semi-supervised word embedding vectors and annotated datasets for different tasks.

\mycomment{
    Sentiment lexicons and word embeddings constitute well-established sources of information for sentiment analysis in online social media. Although their effectiveness has been demonstrated in state-of-the-art sentiment analysis and related tasks in the English language, such publicly available resources are much less developed and evaluated for the Greek language. In this paper, we tackle the problems arising when analyzing text in such an under-resourced language. We present and make publicly available a rich set of such resources, ranging from a manually annotated lexicon, to semi-supervised word embedding vectors and annotated datasets for different tasks. Our experiments using different algorithms and parameters on our resources show promising results over standard baselines; on average, we achieve a 24.9\% relative improvement in F-score on the cross-domain sentiment analysis task when training the same algorithms with our resources, compared to training them on more traditional feature sources, such as n-grams. Importantly, while our resources were built with the primary focus on the cross-domain sentiment analysis task, they also show promising results in related tasks, 
    such as emotion analysis and sarcasm detection.

}


\paper{Pontiki2018Xenophobia}{Exploring the Predominant Targets of Xenophobia-motivated behavior: A longitudinal study for Greece} present a longitudinal linguistic approach on the violence aspect of xenophobia for Greece. They  performed  event analysis e.g., physical attacks and the involved social actors, using news data from 1995 to 2016 and a study on verbal aggressiveness using Twitter data, locating xenophobic stances as expressed by Greeks
in social media for the time period 2013-2016. A follow-up work~\cite{pontiki-etal-2020-verbal} re-examine verbal aggression as an
indicator of xenophobic attitudes in Greek Twitter three years later to  trace possible changes.

\paper{spatiotis2019examining}{Discretization for Sentiment Analysis}
present an approach to  analyze Greek text and extract indicative information towards users' opinions and attitudes. 
A supervised approach has been adopted to analyze and classify comments and reviews into the appropriate polarity category. Discretization techniques are also applied to improve the performance and the accuracy of classification procedures.
\mycomment{
    Nowadays, information, communication and interaction between people worldwide have been facilitated by the rapid development of technology and they are mainly achieved through the internet. Internet users are now new creators of information data and express their ideas, their opinions, their feelings and their attitudes about products and services rather than passive information recipients. Given the evolution of modern technological advances, such as the proliferation of mobile devices social networks and services is extending. User-generated content in social media constitutes a very meaningful information source and consists of opinions towards various events and services. In this paper, we present a methodology that aims to analyze Greek text and extract indicative info towards users' opinions and attitudes. Specifically, we describe a supervised approach adopted that analyzes and classifies comments and reviews into the appropriate polarity category. Discretization techniques are also applied to improve the performance and the accuracy of classification procedures. Finally, we present an experimental evaluation that was designed and conducted and which revealed quite interesting findings.
}

\paper{kefalidou2019apo}{A discourse topicalization construction within Greek Twitter}
describe the creation of a corpus of over 1,300 tweets and 
it analyzes these exchanges based on the analytical
tools of the Lexical Constructional Model \citep{LCM} (a framework that integrates the pragmatics and discourse dimensions of language use).

\mycomment{
    The paper accounts for the Greek discourse topicalization construction APO X, Y and the sarcastic and humorous effects that arise in the context of Twitter exchanges. Our analysis is based on the analytical tools of the Lexical Constructional Model (henceforth LCM) as formulated in Ruiz de Mendoza and Mairal (2008), Ruiz de Mendoza (2013), and Ruiz de Mendoza and Galera (2014). For this purpose we have created a corpus of over 1300 real use tweets. The LCM enables us to treat the patricular uses of the APO X, Y construction. It is shown to be very useful in capturing emergent uses of an already established construction.
}

\paper{Beleveslis2019}{A Hybrid Method for Sentiment Analysis of Election Related Tweets} present a  sentiment  analysis  model for Greek tweets related to politics. A hybrid method that combines Greek lexicons and classification methods was followed.

\paper{pitenis2020offensive}{Offensive Language Identification}
 present the first Greek annotated dataset for offensive language identification: 
 the Offensive Greek Tweet Dataset (OGTD)\myfootnote{\label{ogtd}\url{https://sites.google.com/site/offensevalsharedtask/home}}. 
 OGTD is a manually annotated dataset containing 4,779 posts from Twitter annotated as offensive and not offensive. 
 The authors have also trained a number of systems on this dataset and report that the best results have been 
 obtained from a system using LSTMs and GRU with attention.
 \mycomment{
    An einai prosvasimo mhpws na to anaferoume kai st  resources?
    leei einai available soon 
    \url{https://zpitenis.com/resources/ogtd/}}

In \paper{Alexandridis2021SentimentBERT}{Emotion Detection on Greek Social Media Using Bidirectional Encoder Representations from Transformers} two available BERT type models for Greek (GreekBERT and 
PaloBERT,  see more details in Table~\ref{embedding}) were fine-tuned for the task of emotion detection in social media textual content written in the Greek language. The evaluation showed that fine-tuned models outperform other approaches, especially those based on emotion lexicons.

\subsection{Question Answering (QA)}
\label{sec:QA}
Automatic QA systems are related to the long standing goal of retrieving precise answers to questions expressed in natural language. Such systems have been introduced for the English language already in 60s like the BASEBALL system \citep{Green1961Baseball}.


\paper{Maragoudakis2001NaturalLI}{Natural Language in Dialogue Systems . A Case Study on a Medical Application} 
describe a natural language interaction system, that deals with questions for the treatment of pneumonia from a medical database. The input question is transformed into a logical  representation  by employing a hybrid model of \textit{pattern matching} and \textit{shallow syntactic parsing}.

\paper{marakakis2017apantisis}{APANTISIS -  A Greek Question-Answering System for Knowledge-Base Exploration}
present APANTISIS,
a modular QA system implemented for the Greek language for plugging it to databases or knowledge bases.
\mycomment{
    ready to be attached to any external
    database/knowledge-base. An ingestion module enables the semi/automatic construction of the data dictionary that is
    used for question answering whereas the Greek Language Dictionary, the Syntactic and the Semantic Rules are also
    stored in an internal, extensible knowledge base. After the ingestion phase, the system is accepting questions in natural
    language, and automatically constructs the corresponding relational algebra query to be further evaluated by the external
    database. The results are then formulated as free text and returned to the user. We highlight the unique features of our
    system with respect to the Greek language and we present its implementation and a preliminary evaluation. Finally, we
    argue that our solution is flexible and modular and can be used for improving the usability of traditional database
    systems.
}

\paper{mountantonakis2022QA}{A Comparative Evaluation for Question Answering over Greek Texts by using Machine Translation and BERT} evaluate five Machine Translation models, 
such as Bing and Helsinki, for translating contexts, questions and answers for the task of QA.
According to their experimental evaluation, the Bing model outperformed the others for the Greek language, since it offered a more human readable translation. In addition, they propose an approach for QA  for the Greek language by combining Machine Translation
and BERT QA approaches. Sentence similarity embeddings were used  for deciding whether a predicted answer was correct, partially correct or wrong.

\mycomment{
}

\subsection{Natural Language Generation (NLG)}
\label{sec:NLG}

Natural Language Generation refers to a process that takes as input 
data and text, and generates text in some human language.

\mycomment{ ==probably too detailed
    Natural Language Generation task seems to lack a clear definition. The more simplistic definition is that the aim of NLG systems is to generate text in some human language. In this way, the output of this task is a given but the input remains undefined. Some definitions regard as input both text and data but other restrict the input to non-linguistic representation of information \cite{Gatt2018NLG}. In this survey we follow the more wide definition and we consider as input both data and text.   
}


\paper{dimitromanolaki2001large}{A large-scale systemic functional grammar of Greek}
describe a computational grammar of Greek that is based on Systemic Functional Linguistics \citep{Matthiessen2009SystemicFunctionalLinguistics} for generating automatically descriptions of museum exhibits 
from a database.
\mycomment{
    This paper presents a large-scale computational grammar of Greek, couched in the framework of Systemic Functional Linguistics. The grammar is being developed in the context of M-PIRO, a multilingual natural language generation project, where personalized descriptions of museum
    exhibits are generated automatically from a single database source. Although the grammar is still under development, it already provides a wide coverage of the Greek syntax and morphology. Our long-term goal is to produce a wide-coverage computational grammar of Greek suited to generation applications
}

\paper{Karberis2002}{Transforming Spontaneous Telegraphic Language to Well-Formed Greek Sentences for Alternative and Augmentative Communication}
present an approach for the transformation of spontaneous telegraphic input to well-formed Greek sentences by adopting a feature-based surface realization for NLG.

\paper{2005Mamakissummary}{An algorithm for automatic content summarization in Modern Greek language} 
present an algorithm based on grammatical rules and semantic information dedicated for Greek language for the extraction of document content summary. The proposed method has been tested on the news articles domain.
 \mycomment{
 relative old the only I found dedicated to the greeklanguage. Later publications are mostly multilingual pure statistical approaches.
 }


\paper{Androutsopoulos2013NLG}{Generating Natural Language Descriptions from OWL Ontologies: the NaturalOWL System}
 present the NaturalOWL an open-source natural language generation system that produces texts describing individuals or classes of OWL ontologies (in both English and Greek).
 \mycomment{
        We present Naturalowl, a natural language generation system that produces texts
    describing individuals or classes of owl ontologies. Unlike simpler owl verbalizers, which
    typically express a single axiom at a time in controlled, often not entirely fluent natural
    language primarily for the benefit of domain experts, we aim to generate fluent and coherent multi-sentence texts for end-users. With a system like Naturalowl, one can publish
    information in owl on the Web, along with automatically produced corresponding texts
    in multiple languages, making the information accessible not only to computer programs
    and domain experts, but also end-users. We discuss the processing stages of Naturalowl,
    the optional domain-dependent linguistic resources that the system can use at each stage,
    and why they are useful. We also present trials showing that when the domain-dependent
    linguistic resources are available, Naturalowl produces significantly better texts compared
    to a simpler verbalizer, and that the resources can be created with relatively light effort.

    --
    Our system, called Naturalowl, is open-source and supports both English and Greek.
    Hence, Greek texts can also be generated from the same owl statements, as in the following
    product description, provided that appropriate Greek linguistic resou
}

\paper{lampridis2020lyrics}{Greek Lyrics Generation} 
document the efforts for the automatic lyrics generation in the Greek language for the music genre of ``Éntekhno'' based on LSTM neural networks\myfootnote{\url{https://github.com/orestislampridis/Greek-Lyrics-Generation}}.

\mycomment{
his paper documents the efforts in implementing lyric generation machine learning models in the Greek language for the genre of Éntekhno music.To accomplish this, we used three different Long Short-Term Memory Recurrent Neural Network approaches. The first method utilizes word-level bi-directional network models, the second method expands on the first by learning the wordembeddings on the initial layer of the network, while the last method is based ona char-level network model. Our experimental procedure, which utilized a highsample of human judges, shows that texts of lyrics generated by our models areof high quality and are not that easily distinguishable from actual lyrics.}


\subsection{Machine Translation (MT)}
\label{sec:MT}
MT aims at translating a text from a source language to its counterpart in a target language.


\paper{Dologlouetal2003METIS}{Using monolingual corpora for statistical machine translation: the {METIS} system} present METIS a low-cost
resourced system that relies on monolingual corpora (in the target language) and  bilingual lexica for the translation. In METIS the translation is determined through
pattern recognition-based algorithms.

\paper{Avramidis2008machinetranslation}{Enriching morphologically poor languages
for Statistical Machine Translation} 
address  the  problem  of  translating  from
morphologically poor (English is this case) to morphologically rich languages (Greek and Czech) by adding per-word  linguistic information to the source language. \mycomment{\new{\st{The syntax  of  the  source  sentence  is used to extract
information for noun cases and verb persons
and then the corresponding words are annotated accordingly.}}}
\mycomment{if we lack space we can delete the last sentence}

\paper{Kotsonis2008IE}{Greek-English Cross Language Retrieval of Medical Information} apply machine translation for the cross-language (Greek-English) retrieval
task in the medical domain. The challenging aspect of this work is the  removal of  disambiguation introduced by stemming. The best performance was obtained with the phrase-based retrieval approach.

\paper{anastasiou2009greek}{Greek Idioms Processing in the Machine
Translation System CAT2}
present improvements to the Machine
Translation System CAT2~\citep{sharp1988cat2} for the Greek
to German language pair in relation to idiomatic processing.

\paper{tsoumari2011coreference}{Coreference Annotator A new annotation tool for aligned bilingual corpora}
present an annotation tool that allows the manual annotation of certain linguistic items in the source text and
their translation equivalent in the target
text, by entering useful information about
these items based on their context.
\mycomment{
    This paper presents the main features of an
    annotation tool, the Coreference Annotator, which manages bilingual corpora consisting of aligned texts that can be grouped
    in collections and subcollections according to their topics and discourse. The
    tool allows the manual annotation of certain linguistic items in the source text and
    their translation equivalent in the target
    text, by entering useful information about
    these items based on their context.
}

\paper{Panteli2011greeklish}{A Random Forests Text Transliteration System for Greek Digraphia} 
focus on the transliteration of \textit{Greeklish} to Greek which could be considered as a preprocessing task of MT. They adopt a supervised approach based on \textit{random forests}~\citep{Breiman2001}. 
As a parenthetical note, Greeklish is Greek written using the latin alphabet, 
a writing method that was popular in the early stages of adoption of web in Greece 
and as such  it has attracted also research interest 
\citep{Chalamandaris2006Greeklish,Tsourakis2007Greeklish,Stamou2008Greeklish,Lyras2010Greeklish}.   
It is a fairly unrestricted language without following specific rules. 
The work in~\citep{Mouresioti2021kalimera} depicts the usage of Greeklish today.

\paper{Nikolaenkova2019machinetranslation}{Applying CLP to machine translation: a Greek case study}
applies \textit{constraint logic programming} techniques to deal with polysemy issue in machine translation. 
\mycomment{
    The translation is based on the results of two parsers; the English and the Greek one. 
    Parsing uses a set of constraints (both semantic and grammatical) and the main 
    objective of this research is to evaluate backtracking as one of CLP algorithms for machine translation. 
    For that reason they used Prolog as it is the only programming language that supports backtracking. 
}

\paper{papadopoulos-etal-2021-penelopie}{PENELOPIE: Enabling Open Information Extraction for the {G}reek Language through Machine Translation} apply machine translation for the Open Information Extraction\footnote{It is the task of generating a structured, machine-readable representation of the information in text, usually in the form of triples or n-ary propositions.} (OIE) task for Greek. The authors first built  neural machine translation  models for English-to-Greek and Greek-to-English based on the Transformer architecture and then they leverage the  models to produce English translations of Greek text as input for their pipeline. A series of pre-processing and triple extraction tasks were applied. The code and the related resources are available at~\myfootnote{\url{https://github.com/lighteternal/PENELOPIE}}



\subsection{Miscelanea}
\label{sec:Misc}

\subsubsection{Dialect detection}
\label{sec:Dialect Detection}

Dialect detection  refers to the capability
to identify which particular dialect of a language is being spoken during a recording.

The thesis \paper{sababa2018classifier}{A classifier to distinguish between 
Cypriot Greek and  standard Modern Greek}
describes the collection of a bidialectal corpus of Greek\myfootnote{\url{https://github.com/hb20007/greek-dialect-classifier}} (Cypriot Greek and Modern Greek) and the construction of a classifier based on character and word n-gram features to distinguish between the dialects.

\mycomment{
    This thesis describes the collection of a bidialectal corpus of Greek and the construction of a  classifier to distinguish between the dialects.  The corpus of Cypriot Greek (CG) and Standard Modern Greek (SMG) was compiled from  social media websites such as Facebook, Twitter and online forums. N-gram features were  extracted and three classification algorithms were applied and tested on labeled sentences: multinomial naive Bayes (NB), linear support vector machine (SVM) and logistic regression. All  algorithms classified the test data with an accuracy of over 90\%, with the multinomial NB  classifier performing best, yielding a mean accuracy of 95\%.  This study adds to the existing body of work on the problem of discriminating similar  languages and is the first to examine CG and SMG. The results demonstrate the feasibility of an  accurate Greek dialect classifier for academic or applied purposes.
}

\subsubsection{Lip reading}
\label{sec:LipReading}

Lip reading refers to understanding speech by visually interpreting the movements of the lips, 
face and tongue when normal sound is not available.

\paper{kastaniotis2019lip}{Lip Reading in Greek words at unconstrained driving scenario}
focus on 
the task of lip reading with Greek words in an unconstrained driving scenario by using only visual information. A recognition pipeline  that consists of a Convolutional Neural Network (CNN) followed by a LSTM with a plain attention mechanism is presented. In addition, a dataset with image sequences from Greek  
(10 persons spoke 50 words  while they were either driving or simply sitting in the
    passenger’s seat of a car) is provided.
\mycomment{
    This work focuses on the problem of Lip Reading with Greek words in an unconstrained driving scenario. The goal of Lip Reading (LR) is to understand the spoken work using only visual information, a process also known as Visual Speech
    Recognition (VSR). This method has several advantages over
    Speech Recognition, as it can work from a distance and is not
    affected by other sounds like noise in the environment. In this
    manner, LR can be considered as an alternative method for
    speech decoding which can be combined with state-of-the-art
    speech recognition technologies. The contribution of this work
    is two-fold. Firstly, a novel dataset with image sequences from
    Greek words is presented. In total, 10 persons spoke 50 words
    while they were either driving or simply sitting in the
    passenger’s seat of a car. The image sequences were recorded
    with a mobile phone mounted on the windshield of the car.
    Secondly, the recognition pipeline consists of a Convolutional
    Neural Network followed by a Long-Short Term Memory
    Network with a plain attention mechanism. This architecture
    maps the image sequences to words following an end-to-end
    learning scheme. Experimental results with various protocols
    indicate that speaker independent Lip Reading is an extremely
    challenging problem
}

\subsubsection{Greek Sign Language}
\label{sec:SignLanguage}

The Greek Sign Language (ISO 639-3 code: gss) is a  natural visual language that originates from the French Sign language family and it is officially  used by the Greek deaf community.

 %

 
 \paper{simos2016greek}{Greek sign language alphabet recognition using the leap motion device} 
    present a method for sign language recognition using the Leap Motion controller \myfootnote{\url{https://www.ultraleap.com/tracking/}} and apply it to the recognition of the Greek Sign Language (GSL) alphabet. 
    The method utilizes 3D positional data provided by the device along with \textit{SVM classiﬁcation}.
    \mycomment{
        the goal of this paper is to present a method for sign lan-guage recognition using the Leap Motion Controller and ap-ply it to the recognition of the Greek sign language alpha-bet. The method utilizes 3D positional data provided by thedevice along with SVM classiﬁcation and achieves over 99\%classiﬁcation accuracy in an experiment involving 6 subjects and leave-one-person-out cross validation
    }
    
    \paper{gkigkelos2017greek}{Greek sign language vocabulary recognition using Kinect}
    describe a system that can recognize GSL vocabulary 
    in translation mode using Kinect technology
    (the sensor captures 3D hands movement trajectory and 
    then a set of features in the form of body joints are fed 
    to a classifier to recognize the input sign).
    \mycomment{
        Sign language recognition is a challenging problem both when
        tracking continuous signs (communication mode) or single words
        (translation mode) 1
        . We have developed a system that can
        recognize Greek sign language vocabulary in translation mode
        using Kinect technology. The sensor captures 3D hands
        movement trajectory and then a set of features in the form of body
        joints are fed to a classifier to recognize the input sign.
        Normalization is used to align test and stored trajectories using the
        dynamic time warping algorithm before matching is done using
        the Nearest-Neighbor approach. The low computational
        complexity of the involved algorithms allows for building a
        system with real-time response times. The system was evaluated
        with a sample of 5 individuals and is capable of recognizing 15
        signs of the Greek sign language. Different configurations were
        tested and the best accuracy achieved was 99.33
    }
    
    \paper{kouremenos2018novel}{Rule based machine translation scheme from Greek to Greek sign language (Production of different types of large corpora and language models evaluation)}
    propose a Rule Based Machine Translation (RBMT) system 
    for the creation of large and quality written GSL glossed corpora from Greek text. 
    The proposed RBMT system assists the professional GSL translator in speeding up the production of different kinds of GSL glossed corpora.
    \mycomment{
        One of the aims of assistive technologies is to help people with disabilities to communicate with others and to provide means of access to information. As an aid to Deaf people, in this work we present a novel prototype Rule Based Machine Translation (RBMT) system for the creation of large and quality written Greek Sign Language (GSL) glossed corpora from Greek text. In particular, the proposed RBMT system assists the professional GSL translator in speeding up the production of different kinds of GSL glossed corpora. Then each glossed corpus is used for the production/creation of Language Model (LM) n-grams. With the GSL glossed corpus from Greek text, we can build, test and evaluate different kinds of Language Models for different kinds of glossed GSL corpora. Here, it should be noted that it does not require grammar knowledge of GSL but only very basic GSL phenomena covered by manual RBMT rules as it assists the professional human translator. Furthermore, it should also be stressed that Language Models for written GSL gloss are missing from the scientific literature, thus this work is pioneer in this field. Evaluation of the proposed scheme is carried out for the weather reports domain, where 20,284 tokens and 1000 sentences have been produced. By using the BiLingual Evaluation Understudy (BLEU) metric score, our prototype RBMT system achieves a relative score of 0.84 (84\%) for 4-grams and 0.9 (90\%) for 1-grams.
    }


\subsubsection{Keyword spotting}
\label{sec:KeywordSpotting}

Keyword Spotting refers to the identification of keywords in utterances. It mostly concerns the context of speech and document image processing.


\paper{kesidis2020providing}{Providing Access to {\em old Greek documents} Using Keyword Spotting Techniques}
focus on keyword spotting, a methodology for document indexing based on spotting words directly 
on images without the use of a character recognition system, and 
they describe all steps of this process
(preprocessing for image binarization, enhancement and segmentation, feature representation, matching and word retrieval).
\mycomment{
    Keyword spotting is an alternative methodology for document indexing based on spotting words directly on images without the use of a character recognition system. In this paper, an overview of recent techniques and available databases for keyword spotting is presented focusing on the specific characteristics of old Greek machine-printed and handwritten documents. These documents are import treasures of cultural heritage and a valuable source of information for scholars. Indexing of such content is a very challenging task considering the additional problem of having many character classes as well as a variety of different diacritic mark combinations that may appear above or below Greek characters. All steps of a keyword spotting system are highlighted, namely, preprocessing for image binarization, enhancement and segmentation, feature representation, matching and word retrieval, in order to report the efficiency of current keyword spotting approaches when applied to old Greek documents.
}


\subsubsection{Argument mining}
\label{sec:ArgumentMining}

Argument mining refers to the automatic extraction and 
identification of opinionated structures (e.g., premises, conclusions) from natural language text.

\paper{Goudas2014argument}{Argument extraction from news, blogs, and social Media}
    present a two-step classification approach for argument extraction from social media texts. 
    During the first step,  argumentative sentences are detected while in the second step, the premises in the argumentative sentences are identified.
    An analogous procedure is presented in \cite{sardianos2015argument}  to identify boundaries 
    for claims and premises in argumentative text segments with the addition of word embeddings.

\subsubsection{Computational journalism}
\label{sec:ComputationalJournalism}

Computational journalism refers to the application of computation to the activities of 
journalism such as information gathering, organization, sensemaking, communication and dissemination of news 
information, while upholding values of journalism such as accuracy and verifiability \citep{diakopoulos2011functional}.

\paper{papanikolaou2016journalism}{Just the Facts" with {PALOMAR:} Detecting protest events in Media Outlets and Twitter}
    present a platform for automated data processing in the context of computational journalism 
    along with a general methodology for event extraction from different data sources (news articles and tweets). 
    Another platform that can support journalists and advertisers is the Social Web Observatory 
    \paper{tsekouras2020Observatory}{Social Web Observatory: A Platform and Method for Gathering Knowledge on Entities from Different Textual Sources} 
    that through a pipeline offers text collection/crawling, entity recognition, clustering of texts into events related to entities, entity-centric sentiment analysis, and also text analytics and visualization functionalities.  
\mycomment{We can delete the last two sentences from Papanikolaou paper}

\subsubsection{Health domain}\label{sec:HealthDomain}


Some work has been also done in applying NLP approaches to the health domain customized for the Greek language.


\paper{Vagelatos2011biomedical}{Developing tools and resources for the biomedical domain of the Greek language}
present a suite of linguistic resources and tools for use in high level NLP applications 
in the domain of biomedicine. In this context a Greek morphological lexicon of about 100,000 words, a lemmatiser and a morphosyntactic tagger, 
a specialized corpus of biomedical texts and an ontology of medical terminology has been developed.

\paper{Varlokosta2016aphasia}{A Greek Corpus of Aphasic Discourse: Collection, Transcription, and Annotation Specifications} 
describe GREECAD corpus that is annotated Greek corpus of aphasic spoken discourse.
 \mycomment{
 with 72 patients and 28 controls.
 
 It was followed by an annotated Greek Corpus of Aphasic Discourse17 with 72 patients and 28 controls aged 39–71 years under a large multidisciplinary project ‘THALES-Levels of impairment in Greek aphasia: relationship with processing deficits, brain region, and therapeutic implications’ This project aims to collect, annotate, document, and analyze the spoken discourse of Greek speakers with aphasia.}
 
Another corpus of aphasic speech is presented in \paper{Kasselimis2020Aphasia}{Word Error Analysis in Aphasia: Introducing the Greek Aphasia Error Corpus (GRAEC)}. This is a publicly available corpus and  data
comprising  picture  descriptions  and  free  narrations that  have  been  
transcribed  and  annotated  with  language errors.



\paper{toulakis2019language}{Language Processing for predicting Suicidal Tendencies: A Case Study in Greek Poetry}
use a  set of language dependent and language independent linguistic features  
to represent the poems of 13 Greek poets of the 20th century,
and applies multiple machine-learning algorithms
to predict a writer’s likelihood of committing suicide.
\mycomment{
    Natural language processing has previously been used with fairly high success to predict a writer’s likelihood of committing suicide, using a wide variety of text types, including suicide notes, micro-blog posts, lyrics and even poems. In this study, we extend work done in previous research to a language that has not been tackled before in this setting, namely Greek. A set of language-dependent (but easily portable across languages) and language-independent linguistic features is proposed to represent the poems of 13 Greek poets of the 20th century. Prediction experiments resulted in an overall classification rate of 84.5\% with the C4.5 algorithm, after having tested multiple machine-learning algorithms. These results differ significantly from previous research, as some features investigated did not play as significant a role as was expected. This kind of task presents multiple difficulties, especially for a language where no previous research has been conducted. Therefore, a significant part of the annotation process was performed manually, which likely explains the somewhat higher classification rates compared to previous efforts. 
}

\paper{Rigas2020Stroke}{Analysis of Spontaneous Speech Using Natural Language Processing Techniques to Identify Stroke Symptoms} analyze the spontaneous speech of 58 stroke survivors in relation with their symptoms. From their transcribed responses after prepossessing steps the tf and latent Dirichlet allocation~\cite{Blei2003lda} (LDA) processing techniques were applied. \textit{Speech} and \textit{arm} were the concepts referred most in agreement with the related bibliography.

\subsubsection{NLP and crowdsourcing}
\label{sec:NLPCrowdsourcing}

Crowdsourcing is a sourcing model in which individuals or organizations 
obtain goods and services
from a large, relatively open and often rapidly evolving group of participants, typically through the internet.

\paper{Giouvanakis2013game}{A game with a purpose for annotating Greek folk music in a web content management system} 
present 
``Erasitechnis'', a game for collecting music annotations.

\mycomment{
GWAP, has been designed and deployed on Facebook as an enjoyable way encouraging users to enroll and provide music annotations. Erasitechnis GWAP is a combined approach of existing games with a purpose, which focuses on tagging Greek folk music promoting less frequent tags. By doing so, descriptive tags are collected enabling the creation of a fully annotated dataset, which is exploited to train recommendation or autotagging systems.}

\paper{takoulidou2016social}{Social media and NLP tasks: Challenges in crowdsourcing linguistic information}
describe data collection tasks for parallel translation implemented using a crowdsourcing platform.

\mycomment{
    In the framework of the TraMOOC1(Translation for Massive Open Online Courses) research and innovation project, data collection tasks for parallel translation are implemented using a crowdsourcing platform. The educational genre (videolectures subtitles, forums discussions, course assignments), the type of text (segmentation, misspellings, syntax errors, specialized terminology, scientific formulas, limited knowledge on context) of the source data, and the multilingual approach of the involved activities (the focus is on a total of 12 European and BRIC languages) provides a challenging setting for the success of the project. Experimental trials reveal significant findings for the purposes of Language Technology research as well as limitations in crowdsourcing linguistic data collections for multilingual tasks.
}

\subsubsection{Metaphor detection}
\label{sec:Metaphor}

Metaphor detection refers to the identification
of metaphors, where a metaphor 
is a figure of speech that describes an object or action in a way that is not literally true, but helps explain an idea or make a comparison.
Metaphors are pervasive in natural language, and
detecting them requires  contextual reasoning about whether specific situations can actually happen. 

\paper{florou2018neural}{Neural embeddings for metaphor detection in a corpus of Greek texts}
focus on the automatic differentiation between literal and metaphorical meaning in authentic non-annotated phrases from the Corpus of Greek Texts\myfootnote{\label{goutsos}\url{http://www.sek.edu.gr}} \citep{Goutsos2010corpora} by means of computational methods of machine learning.
\mycomment{
    One of the major challenges that NLP faces is metaphor detection, especially by automatic means, a task that becomes even more difficult for languages lacking in linguistic resources and tools. Our purpose is the automatic differentiation between literal and metaphorical meaning in authentic non-annotated phrases from the Corpus of Greek Texts by means of computational methods of machine learning. For this purpose the theoretical background of distributional semantics is discussed and employed. Distributional Semantics Theory develops concepts and methods for the quantification and classification of semantic similarities displayed by linguistic elements in large amounts of linguistic data according to their distributional properties. In accordance with this model, the approach followed in the current work takes into account the linguistic context for the computation of the distributional representation of phrases in geometrical space, as well as for their comparison with the distributional representations of other phrases, whose function in speech is already “known” with the objective to reach conclusions about their literal or metaphorical function in the specific linguistic context. This procedure aims at dealing with the lack of linguistic resources for the Greek language, as the almost impossible up to now semantic comparison between “phrases”, takes the form of an arithmetical comparison of their distributional representations in geometrical space.
}

\subsubsection{Irony detection}

Irony detection  refers to the identification of irony in texts,
where a text utterance is perceived to be ironic 
if its intended meaning is opposite to what it literally expresses.

\paper{charalampakis2016comparison}{A comparison between semi-supervised and supervised text mining techniques on detecting irony in greek political tweets}
describe a classification schema for irony detection in Greek political tweets
relying  on limited labelled training data, thus a semi-supervised approach is followed, where \textit{collective learning} algorithms take both labelled and unlabelled data into consideration. 
\mycomment{
    The present work describes a classification schema for irony detection in Greek political tweets. Our hypothesis states that humorous political tweets could predict actual election results. The irony detection concept is based on subjective perceptions, so only relying on human-annotator driven labor might not be the best route. The proposed approach relies on limited labeled training data, thus a semi-supervised approach is followed, where collective-learning algorithms take both labeled and unlabeled data into consideration. We compare the semi-supervised results with the supervised ones from a previous research of ours. The hypothesis is evaluated via a correlation study between the irony that a party receives on Twitter, its respective actual election results during the Greek parliamentary elections of May 2012, and the difference between these results and the ones of the preceding elections of 2009.
}

\mycomment{
    \subsubsection{Irony detection}
    
    \paper{charalampakis2016tweets}{A comparison between semi-supervised and supervised text mining techniques on detecting irony in greek political tweets"} proposes a semi-supervised approach for irony detection in Greek political tweets and the comparison with a previous presented supervised approach.
}

\subsubsection{Genre identification}

Genre Identification refers to the task of assigning a particular piece of a text to a literary category (e.g., poem, prose etc).

\paper{Stamatatos2000TextCategorization}{Automatic Text Categorization In Terms Of Genre and Author} describe a method that utilizes 22 stylometric features for the tasks of \textit{genre detection}, \textit{author identification}, and \textit{author verification}. Multiple regression and discriminant analysis have been applied over a Greek corpus containing texts from web. This work among the features already used in previous works introduces  analysis-level style markers (e.g., chunks' morphological descriptions/total detected chunks) that represent the way in
which the input text has been analyzed by a NLP tool.



\paper{tambouratzis2000automatic}{Automatic Style Categorisation of Corpora in the Greek Language}
    present a method for  {\em automatic style categorisation}
    of text corpora in the Greek language,
    where each text is represented
    by a vector of both structural and morphological characteristics,
    and categorisation is achieved by comparing this vector to 
    given archetypes using a statistical-based method.

\mycomment{
 a system is proposed for the automatic style categorisation of text corpora in the Greek language. This categorisation is
based to a large extent on the type of language used in the text, for example whether the language used is representative 
of formal Greek or not. To arrive to this categorisation, the highly inflectional nature of the Greek language is exploited. 
For each text, a vector of both structural and morphological characteristics is assembled. 
Categorisation is achieved by comparing this vector to given archetypes using a statistical-based method. 
Experimental results reported in this article indicate an accuracy exceeding 98\% in the
categorisation of a corpus of texts spanning different registers.
}

\paper{Simaki2012}{Empirical Text Mining for Genre Detection} 
present  an  \textit{exploratory  study}  for identifying  the  features  that  characterize the  genre  of different text types i.e., short stories, new articles, recipes and dictionary.

In \paper{gianitsos2019Stylometric}{Stylometric Classification of Ancient {G}reek Literary Texts by Genre} almost all classical Greek literature was identified as prose or verse with supervised learning based on a stylometric feature set for Ancient Greek language.

\paper{Kontges2020genre}{Measuring Philosophy in the First Thousand Years of Greek Literature} employs LDA topic modelling to automatically retrieve passages in an Ancient Greek corpus by calculating three scores, namely \textit{good and virtue}, \textit{scientific enquiry} and \textit{philosophicalness}.


\subsubsection{Fake news detection}


Fake news detection refers to the process
of spotting the deliberate presentation of (typically) false or misleading claims that are typically presented as news. Note that the term fake news has been gradually been replaced by  term information disorder which is more inclusive~\citep{ERGA2021fake}. 

\paper{mavridis2018fake}{Fake news and social media -  How Greek users identify and curb misinformation online} in his thesis examines the methods and tools Greek users implement in order to spot fake news on social media and counter its spread.   
 The data were collected from the members of  the  Ellinika  Hoaxes Facebook group. 
 
\paper{karidi2019}{Automatic Ground Truth Dataset Creation for Fake News Detection in Social Media} 
propose a system for the automatic creation of fake news datasets 
that is based on the fake news stories 
(e.g., fabrication, propaganda and parody) provided by curated fact-checking websites. 
The proposed method was used for the creation of a Greek dataset with fake news tweets. Unfortunately, the
created dataset is not available.
 

\subsubsection{Legal domain}
\label{sec:DocsToData}

The automatic processing of law text 
could 
support both practitioners and general public to 
have better access to justice, rights and entitlements. 
However, 
the specialized characteristics of legal language requires a different treatment from general text.
We have identified four such works for the Greek language.

\mycomment{
    The automatic processing of law text 
    seem important 
    to support both practitioners and general public to have better access to justice, rights and entitlements. The specialized characteristics of legal language requires a different treatment from general text,
    and to this direction four works for the Greek language has been  inspected.
}


\paper{Markopoulos2012Law}{Stylometric profiling of the Greek Legal Corpus}
study standard stylometric features  in a corpus of Greek legal texts. Further, statistical analysis is applied for the comparison with a corpus of the general language. The formulaic language, the preference
to nouns and impersonal constructions, the use of technical vocabulary, and the length and complexity
of sentences are among of the findings for legal texts.

\paper{Tsimpouris2014LegalAcronym}{Acronym identification in Greek legal texts} 
present a method based on regular expressions for the automated extraction of acronyms in legal Greek texts.

\paper{Chalkidis2017LegislationSW}{Modeling and querying Greek legislation using Semantic Web Technologies} focus on Greek legislation documents and 
propose  an approach for  modeling and querying them based on Semantic Web technologies, as well as
for their integration with other Linked Oped Data.

\paper{garofalakis2018project}{A project for the transformation of Greek legal documents into legal open data} 
present an ongoing project  for  the  automated  analysis  and  processing  
of  Greek legal documents towards their transformation into Legal Open Data.

\paper{papaloukas2021multigranular}{Multi-granular Legal Topic Classification on Greek Legislation} study the task of classifying Greek legal texts. A dataset based on Greek legislation is introduced and then several machine learning approaches are tested ranging from traditional machine learning, RNN-based and Transformer-based methods (both monolingual and multilingual). Monolingual transformer-based models achieved the best performance.

\mycomment{
    In  modern  states,  the  operation  of  the  three  branches  of government  (executive,  legislative  and  judicial)  results  in  the generation of a huge volume of data (e.g., legislative documents, decisions, reports, statistics etc.). Publication of government data in the form of Open  Data  is expected, among other benefits, to drive economic development and promote transparency. This is also  true  for  legal  data,  since  new  services  for  citizens, companies, legal professionals and governments could emerge as a  result  of  the  availability  of  Legal  Open  Data.  Since  such information  is  usually  published  in  unstructured  formats, automated approaches could highly facilitate the transformation of unstructured data into structured Open Data, according to the 5-star Open Data scheme. In this paper, we present an ongoing project  about  the  automated  analysis  and  processing  of  Greek legal documents for their transformation into Legal Open Data. We  briefly  review  the  current  state  of  Legal  Open  Data  in Greece, we present the project’s research questions and analyze our  initial  thoughts  for  the  implementation  methodology;  we discuss the challenges of such an effort and finally we elaborate the expected contributions.
}

\subsubsection{Authorship}
\label{sec:AuthorshipResearch}

Authorship identification (detection or attribution) refers to the process of determining the writer of a document based on a collection of known documents. Author verification, plagiarism detection, author  profiling  or  characterization and the detection  of  stylistic  inconsistencies are some other tasks of authorship analysis.

In \paper{Tambouratzis2004style}{Discriminating the Registers and Styles in the Modern Greek Language-Part 2: Extending the Feature Vector to Optimize Author Discrimination} a method for discriminating among a close set of authors is presented. Each text is represented as a vector of features covering several linguistic aspects and the discrimination  analysis is based on statistical techniques.

\paper{mikros2007topic}{Investigating Topic Influence in Authorship Attribution} explore topic influence in
authorship attribution. They used a balanced corpus of two authors, whose newswire articles are
equally divided in two distinctive topics, culture and politics. Many commonly used stylometric variables were calculated and for each one they performed a two-way ANOVA test~\citep{girden1992anova}. The results  reveal that many stylometric variables  are actually discriminating topic rather than authors.

\paper{Komianos2012BigFive}{Predicting Personality Traits from Spontaneous Modern Greek Text: Overcoming the Barriers} present a machine learning approach to automatically identify Big Five personality traits i.e., openness to experience, conscientiousness, extraversion, agreeableness, neuroticism  from the  linguistic analysis of author's text. The analysis was based on texts produced by   382 participants that each one wrote an  essay  and complete a questionnaire.

In \paper{stamatatos2017Rhesus}{Devising Rhesus: A strange ‘collaboration’ between Aeschylus and Euripides} a long lasting dispute about the authorship of Rhesus play, traditionally attributed to Euripides, is investigated. Frequency of words and character-grams analysis lead the authors to conclude that the  author of Rhesus was probably an actor  whose repertoire included Euripides and Aeschylus (but not Sophocles)
who was a great admirer of Aeschylean grandeur and style. 

\mycomment{OLD:
an actor in whose repertoire Euripides and Aeschylus, but not Sophocles, were included and a great admirer of Aeschylean grandeur and style.
}
In another work for Ancient Greek \citep{gorman2019}  morpho-syntactic data without regard to specific vocabulary items are examined for the task of author identification in short texts by exploiting supervised machine learning approaches. For Modern Greek, \paper{Mikros2015tweetsauthorship}{Authorship attribution in greek tweets using author’s multilevel n-gram profiles} employ a supervised learning approach for authorship attribution and use as features a mixture 
of n-grams both in terms of size (bigrams and trigrams) and type (word and character n-grams).
\paper{Mikros2018Comparison}{Authorship attribution challenges in Greek and in English} present a comparative assessment of the difficulty of authorship attribution in Greek and in English. Based on a corpus introduced in this work different methods of author attribution were applied and the results show that, overall, performance on English is higher than performance on Greek.

\paper{Yamshchikov-etal-2022-plutarch}{BERT in Plutarch’s Shadowsn}  introduces a  BERT model for Ancient Greek. It
demonstrates that Modern Greek BERT~\citep{Koutsikakis2020bert} after transfer learning via Masked Learning Modelling (MLM) on several corpora of Ancient Greek texts could
be further fine-tuned for the  authorship attribution
task. The model is then used   to assist an author and style attribution of a couple of recently discovered texts.

\subsubsection{Education - analytics}
\label{sec:Education}

NLP techniques has been also applied to facilitate teaching and learning processes in Greek.

\paper{Tzimokasetal2015Readability}{Readability scores: Application and reliability issues} 
investigate
some of the most widely known {\em readability} measures namely Flech-Kincaid \citep{Kincaid1975DerivationON}, SMOG \citep{McLaughlin1969} and FOG \citep{Gunning1952Fog}, 
for the Greek language and analyze the degree of their agreement.

\paper{Neofytou2018diakeimenou}{A Tool for Assessing Text Suitability for Greek Language Teaching} present the tool ``diaKeimenou" that assesses the  suitability of a text corpus for use in Greek language teaching by employing various NLP tools (e.g., POS tagger, dependency parser etc). The assessment is based on four criteria: 
(i) text readability, (ii) content, (iii) genre, and (iv) grammatical information.

\paper{Vagelatos2021Logopaignio}{Utilizing NLP Tools for the Creation of School Educational Games} describe the project ``Lexipaignio''  that intends to develop a NLP environment for the creation of digital educational games for students of primary and secondary education in Greece. The games aim to cover a variety of school subjects such as Geography and Biology apart from Linguistics.

\paper{Chatzipanagiotidis2021Readability}{Broad Linguistic Complexity Analysis for {G}reek Readability Classification} address the task of classifying Greek texts according to their readability score. 
For this purpose, textbook Greek corpora are trained by using a total of 215 features.

\paper{korre2021elerrant}{ELERRANT: Automatic Grammatical Error Type Classification for Greek}
introduce the Greek version of the automatic grammatical error annotation tool ERRANT \citep{bryant-etal-2017-automatic} along with two evaluation datasets with errors from native Greek learners and Wikipedia Talk pages edits. Such a tool can have multiple applications to computer aided learning (CAL), for example to help learners of Greek as a Second Language.


\subsubsection{Controlled natural languages}
\label{sec:Control}

Controlled natural languages (CNLs) are subsets of natural languages that are obtained by restricting the grammar and vocabulary. The purpose of such languages is to reduce or eliminate ambiguity and complexity of natural language in order to facilitate tasks such as translation and  retrieval. CNLs can have  human or  machine orientation.

\paper{petasis2001greek}{A Greek Morphological Lexicon and its Exploitation by a Greek Controlled Language Checker}
describe a large-scale Greek morphological lexicon
(along its architecture, construction and evolution process),
that was used to develop a lemmatiser 
and a morphological  analyser. These tools were integrated in a controlled
language checker for Greek.
    
    \mycomment{
    
    This paper presents a large-scale Greek morphological lexicon, developed by
    the Software & Knowledge Engineering Laboratory (SKEL) of NCSR
    "Demokritos". The paper describes the lexicon architecture, the procedure
    followed to develop it, as well as the provided functionalities to update it. The
    morphological lexicon was used to develop a lemmatiser and a morphological
    analyser that were included in a controlled language checker for Greek. The
    paper discusses the current coverage of the lexicon, as well as remaining
    issues and how we plan to address them. Our long-term goal is to produce a
    wide-coverage morphological lexicon of Greek that can be easily exploited in
    several natural language processing applications
}

\paper{vassiliouetal2003evaluating}{Evaluating specifications for controlled Greek} 
describe  a set of controlled language specifications defined for
Modern Greek,
and examine
the effectiveness and suitability of these
specifications by assessing the performance of a commercial machine translation
system over controlled texts.

\mycomment{ 
    {\bf Greek Mythology vocabulary} building \cite{levinson2019greek}:
    ABSTRACT:
    The guiding question addressed in this Capstone is: How will the use of task-based vocabulary activities to support literacy development affect the vocabulary acquisition of elementary Korean English language learners? It documents one teacher’s curriculum development of two units that offer a new perspective to ELL vocabulary acquisition by building vocabulary using morphological training then access and build on background knowledge through practical applications which lead to higher comprehension. This project explores a) the role that task-based curriculum plays on literacy development focused on the development activities that supports the vocabulary development of elementary Korean English language learners using Greek root words and affixes beyond vocabulary word lists. b) Uses grapheme, phoneme, morpheme awareness, integrated in a way that helps students understand how words are built and takes the form of a unit plan adapted from Understanding by Design Backward (Tomlinson & McTighe, 2006). c) Activates student knowledge in the application phase through personalized communication practice.
}

\mycomment{ == Den einai sxetiko. Afora grammatikh mono.
        Greek prepositions in a systemic functional linguistic framework \cite{porter2017greek}:
        ABSTRACT:
        Greek prepositions belong to a class of words that are
        usually called particles. These function words are morphologically
        invariable and enable their function by indicating some kind of
        relationship between larger units. This means that prepositions are
        part of a larger category of words that include not only prepositions
        but conjunctions, adverbs, and possibly other lexemes. Systemic
        Functional Linguistics does not have an explicit theory of the
        preposition. However, prepositions are important within both
        syntagmatic and paradigmatic structure, and function at various ranks
        and as components of various structures at those ranks. In this paper,
        I discuss five topics regarding prepositions: word groups and phrases,
        types of prepositions, prepositions and other relators, the meaning of
        prepositions, and the function of prepositional groups within SFL
        architecture. (Article)
==}




\section{Resources and Evaluation Collections}
\label{sec:EvaluationCollections}

This section is organized as follows:
Section \ref{sec:Portals}
lists a few  {\em portals} containing various NLP resources for the Greek language,
Section \ref{sec:DictionariesLexiconsThesauri}
lists {\em Greek dictionaries, lexicons and thesauri},
Section \ref{sec:CorporaDatasets}
lists  {\em corpora} and {\em datasets}
suitable for comparatively evaluating various NLP tasks (over Greek resources),
and finally
Section \ref{sec:Tools} presents some of the available tools for processing Greek text.

\mycomment{=======
    OLD"
    Here we list a few  {\em portals} containing various NLP resources for the Greek language.
    Then we list {\em Greek dictionaries, lexicons and thesauri}, then some {\em corpora} and in Table~\ref{datasets} some {\em datasets}. Annotated corpora and datasets are suitable for evaluating various NLP tasks (over Greek resources) 
    since they are very useful for obtaining comparative results 
    and thus advancing the current techniques.
    Although roughly every paper reports some evaluation results,  
    and thus it contains or uses an evaluation collection, 
    here we list only some of them,
    therefore 
    we recommend the reader to refer to the  particular papers of interest. Lastly, we present some of the available tools for processing Greek text. 
=================}

Overall, we concentrated on documenting  linguistic resources with permissive licences and whenever it was feasible to provide the link to the downloadable resource. 
Besides of the open-source criterion, 
the selection of the resources was based on other criteria too 
such as the community evaluation 
(when the resource is shared via repositories like GitHub\footnote{\url{https://github.com/}}, e.g., stars), 
whether the resource is related to a peer-reviewed published work and our subjective view about its usefulness and  easiness of integration in an application  etc. Although, the availability of a resource was a determinant factor for the inclusion of the resource in the following lists, we strongly advise the reader that want to reuse a resource to carefully check the license and/or contact the creator(s) for  further clarifications.


\subsection{Portals}
\label{sec:Portals}

We identified the following 12 portals that provide access to resources and informative content. We do not include in this list more general purpose repositories like GitHub, Zenodo\footnote{\url{https://zenodo.org}}  and OSF\footnote{\url{https://osf.io}}.
\begin{itemize}[label={-},leftmargin=0.2cm]
\setlength\itemsep{0.7em}
\item \textbf{{\tt metashare.ilsp.gr}}\myfootnote{\url{http://metashare.ilsp.gr:8080/repository/search/?q=greek}}\\
This portal provides access to  99 resources related to the Greek language (22 with unrestricted use availability).
\item \textbf{{\tt qt21.metashare.ilsp.gr}}\myfootnote{\url{http://qt21.metashare.ilsp.gr/repository/search/?q=greek}}\\
This portal contains 179 resources and tools related to machine translation.  
\item \textbf{{\tt clarin:el}}\myfootnote{\url{https://inventory.clarin.gr/} and 
\url{https://inventory.clarin.gr/resources/search/?q=greek&page=6}}\\ 
The Common Language Resources and Technology Infrastructure (CLARIN) provides access to 756 resources related to the Greek language. The way of accessing these resources is  varying due to restrictions  posed by licences.
\item \textbf{{\tt ELRC-SHARE}} \myfootnote{\url{https://elrc-share.eu/}}\\ It provides access to 256 resources related to the Greek language.  They have been collected through the European Language Resource Coordination and considered useful for feeding the CEF Automated Translation (CEF.AT) platform.
\item {\tt{NLP explorer}} \myfootnote{\url{http://lingo.iitgn.ac.in:5001/}}\\
It provides search and visualization services over data provided at ACL Anthology website \citep{Parmar2019nlpexplorer}.
\item {\tt{glottolog.org}}\myfootnote{\url{https://github.com/glottolog/glottolog}, \url{https://glottolog.org}}\\
Glottolog is a freely available bibliographic database with  
information about the different languages, dialects, and language families of the world.

\item {\tt{phoible.org}}\myfootnote{\url{https://phoible.org/languages/mode1248}}\\
Phoible is a repository of cross-linguistic phonological inventory data that covers Modern Greek. \citep{Moran2019Phoible}

\item {\tt{WALS}}\myfootnote{\url{https://wals.info}}\\
The World Atlas of Language Structures is a 
database of structural (phonological, grammatical, lexical) properties of languages gathered from descriptive materials (such as reference grammars) for many languages among them Greek, Cypriot Greek and Greek Sign Language \citep{Dryer2018Wals}.

\item \textbf{{\tt greek-language.gr}}\myfootnote{\url{http://www.greek-language.gr/greekLang}}\\
A portal with informative content for the greek language both Modern and Ancient.
\item  {\tt{lexilogos}} \myfootnote{\url{https://www.lexilogos.com/english}}\\
It offers a list of resources and an interface to bilingual dictionaries for 100+ languages among them Modern Greek and Ancient Greek.

\item  {\tt{Hugging Face}} \myfootnote{\url{https://huggingface.co/models?search=greek} and \url{https://huggingface.co/languages}}\\
It is a popular hub for state-of-the-art  mostly transformer-based models and datasets.

\item  {\tt{Endangered languages project}} \myfootnote{\url{https://www.endangeredlanguages.com/lang/country/Greece} and \url{https://www.endangeredlanguages.com/userquery/download/}}\\
It is an online resource for samples and research on endangered languages.

\end{itemize}
\mycomment{may be the last bullet to be removed because to save some space (it does not focus on CL and NLP approaches)}



\subsection{Dictionaries, Lexicons and Thesauri}
\label{sec:DictionariesLexiconsThesauri}

There are dictionaries, lexicons and thesauri
for both  Ancient and Modern Greek.

\subsubsection*{Ancient Greek}
\begin{itemize}[label={-},leftmargin=0.2cm]
\setlength\itemsep{0.7em}
\item {\tt{Ancient Greek Wordnet}}\myfootnote{\url{https://dspace-clarin-it.ilc.cnr.it/repository/xmlui/handle/20.500.11752/ILC-56}}\\
It is a lexico-semantic database for Ancient Greek mapped on Princeton WordNet 3.0. Initially, by employing three Ancient Greek -English dictionaries, each Greek word in the dictionaries was linked to the corresponding synsets in the Princeton WordNet. Then, a validation step was performed to a sample. This resource is available through a RESTful API and in the form of triples \citep{bizzoni2014makingancientgreekwordnet,ancientgreekwordnet}. 
\item {\tt{Liddell \& Scott (LSJ)}}\\
It is a
Greek-English lexicon of the Ancient Greek language first published on 1843. A freely available version in electronic form is available by the Perseus Project. For this lexicon various web search interfaces are offered such as the LSJ Wiki\myfootnote{\url{https://lsj.gr/wiki/Main_Page}}.



\item{\tt{Ancient Greek valency lexicon (AGVaLex)}} \myfootnote{\url{https://figshare.com/ndownloader/files/27951132}}\\
A corpus-driven valency (subcategorization) lexicon with 71,868 records automatically extracted from the Ancient Greek Treebank (Perseus)~\citep{McGillivray2021AGVaLex}.

\end{itemize}
\subsubsection*{Modern Greek}
\begin{itemize}[label={-},leftmargin=0.2cm]
\setlength\itemsep{0.7em}
\item  {\tt{General}}\\
\begin{itemize}[label={-},leftmargin=0.6cm]
\item {\tt{Greek Wordnet}}\myfootnote{\url{https://okfn.gr/983/}}\\
It is an RDF
conversion of the Greek Wordnet Database. The database contains 18,461 set of synonyms (synsets) that resulted in a total of 172,066 triples involving 106,432 properties and 18,457 \textit{sameAs} links.
\item {\tt{Cypriot Greek Dictionary}}\myfootnote{\url{https://github.com/themistocleous/CG_Dictionary}}\\
It is a reverse dictionary of Cypriot Greek~\citep{Themistokleous2012Cypriot}.
\end{itemize}

\clearpage
\item  {\tt{Offensive language}}\\
\begin{itemize}[label={-},leftmargin=0.6cm]
\item {\tt{Greek offensive lexicon}}\myfootnote{\url{https://osf.io/t5jey/}}\\
It is a cleaned and expanded version of the Greek part of the  HURTLEX~\citep{Bassignana2018HurtLex} lexicon. The details of this cleaned version are presented in~\citep{Stamou2022GOL}.
\end{itemize}

\item  {\tt{Psycholinguistic}}\\
\begin{itemize}[label={-},leftmargin=0.6cm]
\setlength\itemsep{0.2em}
\item {\tt{GreekLex2}}\myfootnote{\url{https://github.com/CypressA/GreekLex-2}
    and \url{https://psychology.nottingham.ac.uk/greeklex/}}\\
It is a freely available lexical database with part-of-speech, syllabic, phonological, and stress information~\citep{Kyparissiadis2017Lexicon}.

\item {\tt{IPLR}}\myfootnote{\url{http://speech.ilsp.gr/iplr/downloads.htm}}\\
It is a freely accessible \textit{online} psycholinguistic resource for Greek based on analysis of the HNC corpus (see 
 Section \ref{sec:CorporaDatasets}). This resource provides analyses of user-submitted letter strings (words and nonwords) as well as frequency tables for important units and conditions such as syllables, bigrams, and neighbours.~\citep{IPLR2012}.

\item {\tt{HelexKids}}\myfootnote{\url{https://www.helexkids.org/home}}\\
It is an online  word frequency database  derived from a corpus of 116 textbooks used in primary education in Greece and Cyprus. It provides values for Zipf, frequency per million, dispersion, estimated word frequency per million, standard word frequency, contextual diversity, orthographic Levenshtein distance, and lemma frequency~\citep{Terzopoulos2016}.

\item {\tt{Subtlex-GR}}\myfootnote{\url{https://www.bcbl.eu/es/subtlex-gr}}\\
It is an open Modern Greek word frequency database listing more than 23 million Modern Greek words taken from 6,000 subtitle files.~\citep{Dimitropoulou2010subtitles}.



\end{itemize}

\item  {\tt{Sentiment}}\\
\begin{itemize}[label={-},leftmargin=0.6cm]
\setlength\itemsep{0.2em}
\item {\tt{Affective lexicons}} \citep{Palogiannidi2016affective}: It provides two affective Greek lexicons one manually and one automatically annotated. The first one consists of 1,034 words and it is based on the English affective lexicon ANEW \citep{anew}. First, ANEW entries were translated to Greek and then human annotators provided ratings for the three continuous affective dimensions of valence, arousal and dominance. The second lexicon was created by automatically expanded the above lexicon through selecting a small number of the manually annotated words to bootstrap the process of estimating affective ratings of unknown words, resulting in a lexicon with 65,536 words. Both lexicons are freely available (\myfootnote{\url{https://www.researchgate.net/publication/312472262_greek_affective_lexicon_manually_created}}, \url{https://www.researchgate.net/publication/312472196_greek_affective_lexicon_automatically_created}).
\setlength\itemsep{0.2em}

\item  {\tt{Sentiment lexicons}}\myfootnote{\url{https://github.com/MKLab-ITI/greek-sentiment-lexicon}} \citep{tsakalidis2018sentiment}: It provides three sentiments lexicons, one manually annotated and two Twitter-based automatically annotated lexicons. The manual annotated is based on the online version of Triantafyllides' Lexicon \myfootnote{\url{http://www.greek-language.gr/greekLang/modern_greek/tools/lexica/triantafyllides/index.html}} that has then manual annotated by humans and further expanded with all the inflected forms. The creation of the two automatically annotated lexicons, one keyword-based and one emoticon-based, used two annotated Twitter datasets that are also provided.
\end{itemize}

\item  {\tt{Multi-word expressions (MWEs)}}\\
\begin{itemize}[label={-},leftmargin=0.6cm]
\setlength\itemsep{0.2em}
\item {\tt{IDION}} \myfootnote{\url{http://idion.ilsp.gr/data}}\\
A web resource of richly documented \textit{multiword expressions (MWEs)} of  Modern  Greek. In its  current version 
contains about  2,000     verb     MWEs (VMWEs) \citep{markantonatou-etal-2019-idion}. In the context of MWE discovery\footnote{``MWE discovery'' refers to detecting new MWEs in a corpus for lexicographic purposes~\citep{Constant2017Survey}.}, \citep{stamou2020vmwe1,Stamou2020vmwe}  employ \textit{mwetoolkit3}~\citep{ramisch2010multiword} for the detection of verb multiword expressions on fiction  and Twitter data  respectively. They compare the outcome via various  measures and plan the future enrichment of IDION with the detected expressions after human inspection.



\setlength\itemsep{0.2em}
\item {\tt{POLYTROPON}} \\ 
A conceptual lexicon that contains both single and multi-word entries \citep{Fotopoulou2017FromET}.  The lexicon has been based on existing models for semantic representation, i.e., \textit{frame semantics}.

\end{itemize}

\item  {\tt{Slang}}\\
\begin{itemize}[label={-},leftmargin=0.6cm]
\setlength\itemsep{0.2em}
\item {\tt{slang.gr}}\\
A collaborative generated online resource that in its current state contains 24,189 lemmas with definition and examples.  It covers informal or
marginal vocabulary such as professional jargon,
dialectal vocabulary, idiomatic expressions, youth vocabulary, swear words, nonce
words etc. A detailed description for this resource can be found in~\citep{Xydopoulosetal2011slang}.

\item {\tt{lexislang.neurolingo} \url{https://lexislang.neurolingo.gr/}}\\
An online repository for Greek slang words and expressions. The platform is open for the insertion of terms and comments after registration. A group of experts check the appropriateness and the quality of each inserted term.

\item {\tt{slang.fandom}}
\url{https://slang.fandom.com/el/wiki}\\
A wiki-based resource for Greek slang language. Lemmas can be added from registered users. It, in its current state, contains more than 3,500 lemmas.
\item {\tt{cyslang}}
\url{https://www.cyslang.com/}\\
It is an open online dictionary of Cypriot Greek slang. It includes vernacular,
youthful and slang words and expressions~\citep{cyslang2019}.

\end{itemize}

\end{itemize}

\subsubsection*{Greek Sign Language}
\begin{itemize}[label={-},leftmargin=0.6cm]
\item {\tt{NOIMA}}\myfootnote{\url{http://sign.ilsp.gr/signilsp-site/index.php/el/noima/}} \\ 
A bilingual online dictionary of about 10,000 lemmas of general language for languages of Modern Greek and Greek Sign Language.\\
\end{itemize}



\subsection{Corpora and Datasets}
\label{sec:CorporaDatasets}

Annotated corpora and datasets are useful for comparatively evaluating various NLP tasks (over Greek resources) 
    and thus advancing the current techniques.
    Although roughly every paper reports some evaluation results,  
    and thus it contains or uses an evaluation collection, 
    here we list only some of them,
    therefore 
    we recommend the reader to refer to the  particular papers of interest. Table~\ref{datasets} lists annotated datasets that are domain specific and Table~\ref{embedding} lists available  embeddings trained on Greek corpora.

\subsubsection*{Ancient Greek}
\begin{itemize}[label={-},leftmargin=0.2cm]
\setlength\itemsep{0.7em}
\item  {\tt{Treebanks and corpora with morphosyntactic annotation}}\\
\begin{itemize}[label={-},leftmargin=0.6cm]
\item {\tt{Perseus Digital Library}}\myfootnote{\url{https://github.com/PerseusDL/treebank_data}} \citep{Crane1995perseus}\\
It consists of Ancient Greek texts (about 13,507,448 words). Both raw text (https://github.com/PerseusDL/canonical-greekLit) and part-of-speech information is provided.


\item {\tt{Ancient Greek Dependency Treebank 2.0}} \citep{bamman2009ownership} \myfootnote{\url{https://github.com/PerseusDL/treebank_data}} \\
It consists of 550K tokens that encompass Archaic poetry, Classical poetry and prose.
    \item {\tt{Gorman treebanks}}  \citep{gorman2020dependency} \myfootnote{\url{https://openhumanitiesdata.metajnl.com/articles/10.5334/johd.13/}, \url{https://perseids-publications.github.io/gorman-trees/}}\\
   This is a collection of dependency syntax trees (550,000 tokens) of representative texts from ancient Greek prose authors (Aeschines, Antiphon, Appian, Aristotle, Athenaeus, Demosthenes, Dionysius of Halicarnassus, Diodorus Siculus, Herodotus, Josephus, Lysias, Pseudo-Xenophon, Plutarch, Polybius, Thucydides, and Xenophon). The collection is manually annotated by one person (Professor Vanessa B. Gorman) and represents primarily Attic and Atticizing Greek.
  \item {\tt{PROIEL Treebank}} \citep{haug2008Proiel} \myfootnote{\url{https://github.com/proiel/proiel-treebank/}}\\
  A dependency treebank (248K tokens) with morphosyntactic and information-structure annotation.
 \item {\tt{Diorisis Ancient Greek Corpus}} \citep{vatri_mcgillivray_2018} \myfootnote{\url{https://figshare.com/articles/The_Diorisis_Ancient_Greek_Corpus/6187256}} \\
 This is an annotated corpus of literary Ancient Greek corpus consists of 820 texts from Homer to the early fifth century AD, and it counts 10,206,421 words. The texts have been automatically enriched with morphological information for each word. 
\item {\tt{OpenText}}\\
OpenText.org is a web-based initiative dedicated on building  a  representative  corpus  of  Hellenistic  Greek. In its current state is focused on the New Testament manuscript. It follows a bottom-up annotation procedure and it is available at \myfootnote{\url{https://github.com/OpenText-org}}.


\item {\tt{GLAUx}} \citep{keersmaekers2021glaux} \myfootnote{\url{https://perseids-publications.github.io/glaux-trees}}\\ 
An ongoing effort  to develop a large (currently more than 27 million tokens) long-term
diachronic corpus of Greek, covering
sixteen centuries of literary and non-literary material automatically annotated with NLP
methods.

 \item {\tt{Greek Learner Texts Project}} \myfootnote{ \url{https://greek-learner-texts.org/}}\\
The Greek Learner Texts Project is a collaborative effort to produce openly-licensed, annotated Ancient Greek texts 
to help learners of the language. It contains texts from the following categories: 
(i) 19th/20th Century Easy Greek Readers, (ii)  Plato, 
(iii) Other Post-Beginner Greek Prose, (iv)  Biblical and Early Christian Texts, and 
(v)  New Testament Epitomes.
\end{itemize}

\begin{itemize}[label={-},leftmargin=0.2cm]
\setlength\itemsep{0.7em}
\item  {\tt{Raw corpora}}\\ 
\begin{itemize}[label={-},leftmargin=0.6cm]
\item {\tt{First1KGreek}}\myfootnote{\url{https://opengreekandlatin.github.io/First1KGreek/}}\\
This resource collected Greek works composed between Homer and 250CE with a focus on texts that do not already exist in the Perseus Digital Library. The repository contains 23,366,087 words.

 \item {\tt{Patrologia Graeca}} \myfootnote{ \url{https://github.com/OGL-PatrologiaGraecaDev}}\\
An effort for the machine readable representation of the  volumes of the Patrologia Graeca\footnote{Patrologia Graeca is an edited collection of writings by the Christian Church Fathers and various secular writers, in the Greek language. It consists of 161 volumes produced in 1857–1866 by J. P. Migne's Imprimerie Catholique, Paris \url{https://patristica.net/graeca/}.}
\end{itemize}

\begin{itemize}[label={-},leftmargin=0.2cm]
\setlength\itemsep{0.7em}
\item  {\tt{Corpora for NER}}\\
\begin{itemize}[label={-},leftmargin=0.6cm]
\item {\tt{Herodotos}} \citep{Erdmann2019Herodotus} \myfootnote{\url{https://github.com/Herodotos-Project/Herodotos-Project-Latin-NER-Tagger-Annotation}}\\ A manually annotated corpus for people and group names. The original texts were drawn from the Perseus corpus. The corpus contains about 522 annotations.

\end{itemize}
\end{itemize}

\end{itemize}

\end{itemize}

\subsubsection*{Modern Greek}
\begin{itemize}[label={-},leftmargin=0.2cm]
\setlength\itemsep{0.7em}
\item  {\tt{Treebanks and corpora with morphosyntactic annotation}}\\
\begin{itemize}[label={-},leftmargin=0.6cm]
\item {\tt{Greek UD treebank (UD$\_$Greek$\_$GDT)}}\\
The Greek UD treebank (UD$\_$Greek$\_$GDT) is a subset of the Greek Dependency Treebank\myfootnote{\url{http://gdt.ilsp.gr}}. 
The Greek Dependency Treebank is an ongoing project
of the Institute for Language and Speech Processing in order to manually annotate Modern Greek texts for morphology, syntax and semantics. The UD$\_$Greek$\_$GDT resource consists of texts that are in the public domain, including wikinews articles and normalized transcripts of European parliamentary sessions. The current version of the UD$\_$Greek$\_$GDT consists of 2,521 sentences (61,673 tokens) and it is available at\myfootnote{\url{https://github.com/UniversalDependencies/UD_Greek-GDT}}.
\item {\tt{Universal Morphology (UniMorph)}} \citep{sylak-glassman-etal-2015-language}
\myfootnote{\url{https://unimorph.github.io/}}.
\\
The UniMorph project is  a collaborative and open-source project at cataloguing inflectional
and derivational morphology for many languages including Modern Greek and Ancient Greek. Greek has been added in the third version of the project~\citep{mccarthy-etal-2020-unimorph}.
\end{itemize}

\item  {\tt{Raw corpora}}\\
\begin{itemize}[label={-},leftmargin=0.6cm]
\item {\tt{Corpus of Greek Texts (CGT)}} \citep{Goutsos2010corpora}\\
It constitutes the first electronic corpus of Greek with about 30 million words. It contains texts, i.e., transcribed utterances of verbal speech and written text, from two decades (1990 to 2010). The motivation (from a linguistic and pedagogical point of view) and its design is described in \cite{goutsos1994towards}.
\mycomment{
            Our aim in this paper is twofold: to provide a state-of-the-art description of corpora of Modern Greek and, on the basis of this, to argue for the development of a spoken corpus of Modern Greek. To this end, we discuss the main attempts at electronic text collection in the context of empirical research in Greek linguistics, present a survey of existing corpora we have conducted, and provide an assessment and an identification of current needs at the beginning of the nineties. in the second part, we put forward a linguistic and pedagogical rationale for a corpus of Modern Greek spoken texts and delineate the basic features of its design.
         }
\item {\tt{Hellenic National Corpus (HNC)}} \citep{hatzigeorgiu2000design}\\
It is a general language corpus of Modern Greek with currently 97 million words. It  consists of written texts published, in their majority, from 1990 and beyond. A web interface is available for searching the corpus and it is free for research purposes\myfootnote{\url{http://corpus.ilsp.gr}}. A processed subset of this dataset with about 100,000 words entitled {\tt{Golden Part of Speech Tagged Corpus}} is freely available. The processing includes among others cleaning of boilerplate, manual error correction, POS tagging and lemmatization.\myfootnote{\url{https://inventory.clarin.gr/corpus/870}}.

\item {\tt{Corpus of Modern Greek}} \citep{Eloeva2019Corpora}\\
It consists of approx. 35.7 million tokens
Corpus  of  Modern  Greek and it  is  not  restricted  to  any  specific  type  of  texts (e.g., fiction, poetic, official,  scientific,  and  religious  texts) but the majority of them originates from contemporary newspapers \myfootnote{\url{http://web-corpora.net/GreekCorpus/search/?interface_language=en}}.

\item {\tt{CC-100 - Greek dataset}}\myfootnote{\url{http://data.statmt.org/cc-100/}} \citep{Wenzek2020CCNet}\\
This is the greek part of a large corpus with monolingual data for 100+ languages as an attempt to recreate the dataset used for training XLM-R \citep{Conneau2020Unsupervised}. The data resulted in a 50 GB corpus originated from Common Crawl snapshots at the period January-December 2018 following the procedure described in the CC-Net repository \myfootnote{https://github.com/facebookresearch/cc\_net}. This corpus is free available without any restrictions.

\item {\tt{elTenTen: Corpus of the Greek Web}}\\
The Greek Web Corpus (elTenTen) is a language corpus made up of crawled web content. The corpus belongs to the TenTen corpus family which is a set of the web corpora built using the same method with a target size 10+ billion words. The corpus is available through various tools (e.g., n-grams, connotations etc.) and for some user roles fees are applied for its use\myfootnote{\url{https://www.sketchengine.eu/eltenten-greek-corpus/}}.

\item {\tt{OSCAR}}\\
The Greek part of Open Super-large Crawled Aggregated coRpus obtaining from Common Crawl corpus \myfootnote{\url{https://huggingface.co/datasets/oscar-corpus/OSCAR-2201}}.
\item {\tt{Greek Learner Corpus (GLC)}} \citep{Tantos2016GLC}\\
A collection of written productions based on the language proficiency assessment tests `Let's speak Greek I, II, III', taken by students of Greek in `Reception Classes' of Greek schools. GLC is accessible through UAM Corpus Tool \citep{Donnell2008UAMTool}. A part of this  corpus, for the time being, has been annotated with part of speech tags and with learner's errors \myfootnote{\url{http://hdl.handle.net/11500/AUTH-0000-0000-4B62-E}}. 
\end{itemize}

\item  {\tt{Corpora for NER}}\\
\begin{itemize}[label={-},leftmargin=0.6cm]
\item {\tt{elNER}} \citep{Bartziokas2020}\\
A manually annotated corpus of Greek newswire articles, coined as elNER. Two versions of elNER are provided, namely elNER-4 and elNER-18\myfootnote{\url{https://github.com/nmpartzio/elNER}}. The elNER-4 concerns four categories, i.e., Location, Organization, Person and Misc, via merging similar categories of that used in elNER-18. \\

\end{itemize}

\item  {\tt{Corpora for QA}}\\
\begin{itemize}[label={-},leftmargin=0.6cm]
\item {\tt{GreekTexts}} \citep{mountantonakis2022QA}\\
An evaluation collection for QA systems over greek texts. Several answers of that collection were manually annotated. The manually labelled answers
can be also exploited for other tasks, such as for sentence similarity.

\end{itemize}

\item  {\tt{Corpora for coreference resolution}}\\
\begin{itemize}[label={-},leftmargin=0.6cm]
\item {\tt{Coreference Corpus (GCC)}} \citep{GCC2015} \myfootnote{\url{https://inventory.clarin.gr/corpus/752}}\\
Greek corpus annotated with coreference, near identity and bridging relations.
\end{itemize}

\item  {\tt{Corpora for textual entailment}}\\
\begin{itemize}[label={-},leftmargin=0.6cm]
\item {\tt{Greek Textual Entailment Corpus (GTEC)}} \citep{Marzelou2008EntailmentCorpus} \myfootnote{\url{https://inventory.clarin.gr/corpus/689}}\\
The GTEC consists of 600 T-H pairs manually annotated for entailment (i.e., whether T entails H or not).

\item {\tt{Dedropped XNLI}} \citep{Amanakietal2022NLI} \myfootnote{\url{https://github.com/GU-CLASP/LREC_2022/blob/main/datasets/en_el_drop_dev.txt}}\\
This is a de-dropped version of the Greek XNLI dataset~\citep{Conneau2018xlni}, where the pronouns that are missing due to the prodrop nature of the language are inserted.

\item {\tt{Extended Greek FraCaS}} \citep{Amanakietal2022NLI} \myfootnote{\url{https://github.com/GU-CLASP/LREC_2022/tree/main/datasets}}\\
This is an extension of the FraCaS~\citep{Pulman1996} dataset for Greek and  comprises 774 examples of inference.  The original FraCaS test suite for
Greek (346 examples) has been validated by experts and nonexperts while  428 further examples of inference have been added.

\end{itemize}

\item  {\tt{Spoken data}}\\
\begin{itemize}[label={-},leftmargin=0.6cm]
\setlength\itemsep{0.2em}
\item {\tt{Corpus of spoken Greek}} \new{\citep{Pavlidou2012CSG}}\myfootnote{ \url{http://ins.web.auth.gr/index.php?lang=en&Itemid=251}}\\
It consists of three parts (i) audiovisual material of naturally-occurring communication, 
(ii) transcriptions for a 
            subset of the audiovisual content ($\sim$2 million words), and
(iii) subset of the transcriptions that are freely available online (upon request).\\
\item {\tt{Greek audio dataset}} \citep{GAD2014} \myfootnote{\url{https://hal.inria.fr/hal-01391043}}\\
A freely available collection of audio features and metadata, e.g., lyrics, for a thousand popular Greek tracks.\\
\item {\tt{SpeDialDatasets}} \citep{lopes-etal-2016-spedial} \myfootnote{\url{https://github.com/zedavid/SpeDialDatasets}}\\
A spoken dialogue dataset that consists of 200 dialogues in Greek  collected  through  a  call  center  service  for  retrieving information about movies, e.g., show times information and ticket  booking.\\
\item {\tt{Greek Single Speaker Speech Dataset}} \citep{park2019css10}
\myfootnote{\url{https://github.com/kyubyong/css10}}\\
The Greek part of a collection of single speaker speech datasets for 10 languages. Each of them consists of audio files recorded by a single volunteer and their aligned text sourced from LibriVox\footnote{\url{https://librivox.org}}.\\
\item {\tt{Greek Heritage Language Corpus (GHLC)}} \citep{Gavriilidou2019Corpus} \myfootnote{\url{https://synmorphose.gr/index.php/el/projects-gr/ghlc-gr-menu-gr/prosvasi-sti-ghlc}}
It is a freely available spoken
corpus of Greek as a heritage language, including data from the 1st, 2nd and 3rd generation
Greek heritage speakers living in Chicago, Moscow and Saint Petersburg. It contains 144,987
tokens and approximately 90 hours of recordings.
\item {\tt{Multi-CAST cypriot}} \citep{hadjidas2015multicast} \myfootnote{\url{https://github.com/StergiosCha/Greek_dialect_corpus}}\\
The Cypriot Greek part of the Multi-CAST corpus (Multilingual Corpus of Annotated Spoken Texts) contains three annotated narratives (1,071 clauses units) originally recorded in the 1960s. For each text a translation, as
well as morphological glossing and syntactic annotations are provided, along
with background information and additional sources.
\end{itemize}

\item  {\tt{Image data}}\\
\begin{itemize}[label={-},leftmargin=0.6cm]
\setlength\itemsep{0.2em}
\item {\tt{GRPOLY-DB \& Polyton-DB}}~\citep{Gatos2015GRPOLYDB,simistira2015recognition} \myfootnote{\url{https://users.iit.demokritos.gr/~nstam/GRPOLY-DB/}}
and \myfootnote{\url{http://media.ilsp.gr/PolytonDB/}}
\\ These collections contain polytonic Greek documents from different periods. The Polyton-DB extends of GRPOLY-DB with the latter to  consist  of scanned pages only.

\item {\tt{Pogretra}} \citep{Robertson2021Pogretra} \myfootnote{\url{https://zenodo.org/record/4774201}}\\ A collection of training data and pre-trained models for a wide variety of polytonic Greek texts.

\end{itemize}

\end{itemize}

\subsubsection*{Endangered}
\begin{itemize}[label={-},leftmargin=0.2cm]
\setlength\itemsep{0.7em}
\item {\tt{Griko -Italian parallel corpus}} \citep{boito2018grikocorpus}\\
This is a small parallel speech corpus between the endangered language Griko and Italian. It consists of 330 sentences, with speech, machine extracted pseudo-phones, transcriptions, translations and sentence alignment information and it is available at \myfootnote{\url{https://github.com/antonisa/griko-italian-parallel-corpus}}.

\end{itemize}

\subsubsection*{Greek Sign Language}

\begin{itemize}[label={-},leftmargin=0.2cm]
\setlength\itemsep{0.7em}
\item { Greek Sign Language Corpus (GSLC)}~\citep{Efthimiou2007GSC}\\
The first annotated  corpus for the Greek Sign Language (GSL).

\item {Polytropon Parallel Corpus}
\citep{Efthimiou2018Polytroponcorpus}
\myfootnote{\url{https://inventory.clarin.gr/corpus/835}}\\
An annotated corpus for the language pair Greek Sign Language – Greek. The corpus volume incorporates 3,600 clauses performed by a single signer and its download requires registration.

\item {\tt{Greek  Sign Language Sign (GSL) dataset}} \citep{Adaloglou2020comprehensive}
\myfootnote{\url{https://vcl.iti.gr/dataset/gsl/}}\\
It is a large-scale dataset (10,290 sentence instances)  suitable for Sign Language Recognition  and Sign Language Translation. It contains temporal gloss annotations, sentence  annotations,  and  translated annotations to the Modern Greek language.
\end{itemize}

\begin{small}
\begin{longtable}{p{3cm}p{2cm}p{70mm}}
Dataset & Category & Description \\ 
MultiLing 2011 Pilot
\citep{Giannakopoulosetal2011multiling}&summarization&A summarization dataset in 7 languages. The data were prepared by sentence-by-sentence translation of summaries written in English. The translation was performed by humans, mostly native speakers of each language. $|$\myfootnote{\url{
http://multiling.iit.demokritos.gr/file/download/353}}\\ \midrule

Sentiment ratings for Greek tweets \citep{kalamatianos2015sentiment}& sentiment & 683 manually rated tweets for their sentiment intensity. $|$ \myfootnote{\url{https://www.researchgate.net/publication/301888705_Sentiment_Ratings_for_Greek_Tweets_Dataset/link/572b181908aef7c7e2c68639/download}}\\ \midrule
GREECAD \citep{Varlokosta2016aphasia} & aphasia & Corpus of aphasic discourse (no data has been made  available yet).\\ \midrule
Greek political tweets \citep{charalampakis2016comparison} & POS
& It contains 61,427 Greek tweets collected on the week before and the week after the May 2012 parliamentary elections in Greece. This dataset is available for research purposes. $|$\myfootnote{\url{https://hilab.di.ionio.gr/wp-content/uploads/2020/02/HILab-Greek_POS_Tagged_Tweets.csv}} \\ \midrule
 Gazzetta dataset \citep{pavlopoulos2017abusiveingreek}& abusive & 1.6M moderated user comments from a Greek sports news portal. $|$\myfootnote{\url{https://archive.org/details/gazzetta-comments-dataset.tar}} \\ \midrule
Legislation dataset \citep{angelidis2018named}& NER \& NEL & A manually annotated dataset from the legislation domain (254 daily issues  of the Greek Government Gazette over the period 2000-2017) for use in named entity recognition and named entity linking tasks. $|$~\myfootnote{\url{http://legislation.di.uoa.gr/publications/ner_dataset}} \\ \midrule

Hate in Greek tweets \citep{Charitidis2019HateDataset}&  hate speech & This  dataset contains 61,481 Greek tweet ids and corresponding annotations for hate speech and personal attack. The gathering of tweets and the annotation process followed an active learning procedure analytically described at~\citep{CHARITIDIS2020Publication}. $|$~\myfootnote{\url{https://zenodo.org/record/3520157}} \\ \midrule

GRAEC~\citep{Kasselimis2020Aphasia} & aphasia & Greek Aphasia Error Corpus is a publicly available corpus (upon request) with data from Greek patients with aphasia. Data comprise  of picture description and free narration that have been transcribed and annotated with language errors. The corpus currently contains 17,507 words with 2,397 annotated errors. $|$ \myfootnote{\url{http://aphasia.phil.uoa.gr/}}\\ \midrule
OGTD \citep{pitenis2020offensive} & offensive & Offensive Greek Tweet Dataset (OGTD) is a manually annotated dataset containing 4,779 posts from Twitter. The posts has been annotated as offensive and not offensive. $|$ \myfootnote{\url{https://zpitenis.com/resources/ogtd/}} \\  \midrule

GLC \citep{papaloukas2021multigranular} & topic classification & Greek Legal Code (GLC) is a publicly available dataset consisting of 47k Greek legislation resources categorized by topic. $|$\myfootnote{\url{https://github.com/christospi/glc-nllp-21}} \\  \midrule
Grico OCR~\citep{rijhwani2020ocr} & folk tales in Grico & It is a benchmark dataset containing transcribed images for OCR and OCR post-correction in Grico and other endangered languages. For the case of Grico a book with folk tales is used resulting in 807 annotated sentences. $|$\myfootnote{\url{https://shrutirij.github.io/ocr-el/}}\\  \midrule
Hate in Greek tweets \citep{Perifanos2021Hate} & hate speech  & It is a manually annotated dataset of 4,004 Greek tweets concerning refugees and migrants (1,040 toxic and 2,964 non-toxic).  $|$\myfootnote{\url{https://github.com/kperi/MultimodalHateSpeechDetection}}\\  \midrule
GNC \& GWE \citep{korre2021elerrant}  & edit classification \& grammatical error correction  & Greek Native Corpus (GNC) contains with errors from native Greek learners and  Greek WikiEdits Corpus (GWE) contains Wikipedia Talk Pages edits. $|$\myfootnote{\url{https://github.com/katkorre/elerrant}} \\  \midrule
elAprilFoolCorpus~\citep{Papantoniou2021ElApril}  & deception  & This dataset consists of diachronic April Fools' Day (AFD) and normal articles in the Greek language. The current version of the dataset consists of 254 AFD articles and 254 normal articles collected from 112 different newspapers and websites, spanning over the period of 1979 - 2021.  $|$\myfootnote{\url{https://gitlab.isl.ics.forth.gr/papanton/elaprilfoolcorpus}} \\  \midrule

GMW \citep{soundexGR2021}  & error correction, approximate matching etc.   & Greek Misspelled Words (GMW) dataset contains 574,883 distinct Greek words and for each one of them it contains various misspellings that they have produced algorithmically. $|$\myfootnote{\url{http://islcatalog.ics.forth.gr/dataset/gmw}}, \myfootnote{\url{https://github.com/YannisTzitzikas/SoundexGR}} \\ \midrule

{\tt{Parliamentary questions corpus}\citep{fitsilis_fotios_2021_4747452}} & parliament data & It contains 2,000 written parliamentary questions from all Greek political parties, submitted during the period 2010 - 2019. $|$ \myfootnote{\url{https://zenodo.org/record/4747452}}\\ 

\midrule

 {\tt{Parliament Proceedings }\citep{Dritsa2022Parliament}} & parliament data & It includes 1,280,927 speeches of Greek parliament members with a total volume of 2.15 GB, that where exported from 5,355 parliamentary sitting record files and extend chronologically from 1989 up to 2020. $|$ \myfootnote{\url{https://zenodo.org/record/7005201}}\\ 
\midrule

 {\tt{Wikipedia current event pages }\citep{papantoniouetal2022nernel}} &  NER \& NEL &   An automated benchmark dataset for NER and NEL tools, based on Greek Wikipedia current events pages. The current version contains 18,617 events annotated with 40,798 entity mentions and 36,189 links to elWikipedia (and wikidata ids).  $|$ \myfootnote{\url{https://zenodo.org/record/7429037}}\\

\midrule
\caption{Domain specific annotated datasets.\label{datasets}} 
\end{longtable}
\end{small}

\begin{table}[htbp]
\caption{Embeddings for the Modern Greek language.}
\begin{tabular}{@{\extracolsep{\fill}}p{2cm}p{2cm}p{7.5cm}}
\hline
Algorithm/Model                               &Corpus                 &Description \\ \hline
Word2Vec Continuous Skipgram            &tweets                 &Word embeddings of length 300 produced from a 15M tweets corpus~\citep{tsakalidis2018sentiment}. \\  
Embeddings from Language Models (ELMo)  &Greek CoNLL17 &ELMo word embeddings \citep{Che2018elmo}\myfootnote{\url{http://vectors.nlpl.eu/repository/20/143.zip}}.\\ 
Word2Vec Continuous Skipgram &Greek CoNLL17 &  Word embedding of 100 dimension with window  size 10 and without lemmatization\myfootnote{\url{http://nlp.polytechnique.fr/resources-greek}}. \\ 

Word2Vec Continuous Skipgram&crawled web content & Word embeddings of dimension 300 produced from the processing of  20 million URIs resulting in a corpus of size 50GB.\myfootnote{\url{http://nlp.polytechnique.fr/resources-greek}} \citep{outsios2018word}.\\

Word2Vec
Skipgram

&legislation documents  & 100 dimensional word embeddings for a vocabulary of 428,963 words, based on 615 millions of tokens. \myfootnote{\url{http://legislation.di.uoa.gr/publications/ner_dataset}} \citep{angelidis2018named}.\\



BERT &web content and parliament  &Uncased, 12-layer, 768-hidden, 12-heads, 110M parameters on a corpus of 29 GB with a 35k subword BPE (Byte Pair Encoding) vocabulary \citep{Koutsikakis2020bert}\myfootnote{\url{https://github.com/nlpaueb/greek-bert}}. \\ 


ELECTRA small& crawled web content &Uncased, 12-layer, 256-hidden, 14M parameters on a corpus of size 80GB. 
\myfootnote{\url{http://nlp.polytechnique.fr/uploads/electra_small_el.zip}}. \\  


PaloBERT&458,293 texts from social media &Greek language model based on RoBERTa \citep{roberta}, uncased \myfootnote{\url{https://huggingface.co/gealexandri/palobert-base-greek-uncased-v1}} \citep{Alexandridis2021PaloBERT}.\\

gpt2-greek&5GB Greek texts&It is a GPT-2 type model~\citep{Radford2019language} trained on data mostly extracted from Greek Wikipedia\myfootnote{\url{https://huggingface.co/nikokons/gpt2-greek}}. This model is the ``small'' version of GPT-2 (12-layer, 768-hidden, 12-heads).\\ 

GreekBART&77GB Greek texts&Various versions are provided at\myfootnote{\url{http://nlp.polytechnique.fr/resources-greek}}~\citep{evdaimon2023greekbart}.\\ 

%


\hline
\end{tabular}
   
 \label{embedding}
\end{table}

\begin{table}[htbp]
\caption{Embeddings for the Ancient Greek language.}
\begin{tabular}{@{\extracolsep{\fill}}p{2cm}p{2cm}p{7.5cm}}
\hline
Algorithm/Model                               &Corpus                 &Description \\ \hline
Embeddings from Language Models (ELMo)  &Ancient Greek CoNLL17 &ELMo word embeddings ~\citep{Che2018elmo}
\myfootnote{\url{http://vectors.nlpl.eu/repository/20/153.zip}}.\\ 
Word2Vec Continuous Skipgram &Ancient Greek CoNLL17 & Word embedding of 100 dimension with window  size 10 and without lemmatization\myfootnote{\url{http://vectors.nlpl.eu/repository/20/30.zip}}.\\ 




Word2Vec
Skipgram

&homeric texts  & 300 dimensional word embeddings trained on the homeric texts. \myfootnote{\url{https://github.com/amasotti/homer-skipgram}}.\\




Ancient Greek BERT& collection of ancient Greek texts &Uncased, unaccent, 12-layer, 768-hidden, data from First1KGreek Project, Perseus Digital Library, PROIEL 
\myfootnote{\url{https://github.com/pranaydeeps/Ancient-Greek-BERT}} \citep{Singh2021ancient-greek-bert}.\\

GRC-ALIGNMENT& 12 million tokens&It is a XLM-RoBERTa-based model, fine-tuned for automatic multilingual text alignment at the word level. The model is trained on 12 million monolingual Ancient Greek tokens. Further, the model is fine-tuned on 45k parallel sentences, mainly in Ancient Greek-English, Greek-Latin, and Greek-Georgian. \myfootnote{\url{https://huggingface.co/UGARIT/grc-alignment}}~\citep{yousef2022Alignment}\\
%




\hline
\end{tabular}
   
 \label{ancientembedding}
\end{table}

\subsection{Tools and Toolkits}
\label{sec:Tools}

Tables~\ref{toolkits} and~\ref{ancienttoolkits}
list  suites of tools and the functionality (NLP tasks) supported  by each one of them for Modern and Ancient Greek respectively.


\begin{table}[htbp!]
\caption{Toolkits for Modern Greek NLP.}
   \begin{tabular}{lp{1.1cm}p{1cm}p{1cm}p{1cm}p{1cm}p{1.2cm}p{1cm}p{0.8cm}}\hline
                   	&	  ILSP           	&	  Neural ILSP          	&	  SpaCy   	&	Polyglot 	&	  Stanza          	&	  
                   gr-nlp-toolkit
                   	       & Trankit & Spark NLP\\ \hline
                      	&	   \citep{Prokopidis2011}            	&	  \citep{Prokopidis2020}    	&	       	&	 \cite{Rami2006polyglot} 	&	   \citep{Qi2020Stanza}         	&	\citep{Dikonimaki2021aueb1,Smyrnioudis2021aueb2}	&	 \citep{Nguyen2021trankit} &\citep{Kocaman2021sparknlp} \\ \hline
Tokenizer              	&	  $\bullet$   	&	  $\bullet$  	&	   $\bullet$   	&	 $\bullet$  	&	  $\bullet$  	&	& $\bullet$	&	 \\ 
Lemmatizer             	&	  $\bullet$   	&	  $\bullet$  	&	   $\bullet$   	&	  	&	  $\bullet$  	&		& $\bullet$	& $\bullet$ \\ 
Prosody Scanning       	&	                 	&	                	&	       	&	  	&	             	&		&	&  \\ 
Stopword filtering     	&	                 	&	                	&	        	&	  	&	            	&		& & $\bullet$ \\ 
Sentence splitter      	&	  $\bullet$   	&	  $\bullet$  	&	   $\bullet$   	&	 $\bullet$  	&	  $\bullet$  	&		& $\bullet$	& \\ 
POS                    	&	  $\bullet$   	&	  $\bullet$  	&	   $\bullet$   	&	  	&	  $\bullet$   	&	  $\bullet$  	& $\bullet$	&	  $\bullet$   \\ 
NER                    	&	  $\bullet$   	&	  $\bullet$  	&	   $\bullet$   	&	  $\bullet$  	&	         	&	  $\bullet$  &	&         \\ 
Chunker                	&	  $\bullet$   	&	  $\bullet$ (noun)  	&	   $\bullet$ (noun)   	&	  	   \\   
Dependency parser      	&	  $\bullet$   	&	  $\bullet$  	&	$\bullet$      	&	  	&	  $\bullet$          	&	  $\bullet$  	& $\bullet$	&	    \\ 
Transliterator         	&	  $\bullet$   	&	                	&	           	&	 $\bullet$  	&	         	&		&	&  \\ 
Sentiment analyzer     	&	                 	&	            	&	       $\bullet$  	&	 $\bullet$  	&	    	&		&	     &         \\ 
Topic Classification   	&	                 	&	  $\bullet$  	&	   $\bullet$  	&	  	&	   	&		&	    &      \\
Translation   	&	                 	&	 	&	   	&	  	&	   	&		&	    &   $\bullet$           \\
Word2Vec               	&	                 	&	                	&	   $\bullet$  	&	 $\bullet$  	&	   	&		& & $\bullet$ \\ 
BERT               	&	                 	&	                	&	   $\bullet$  	&	 $\bullet$  	&	   	&		& & $\bullet$   \\
\hline

 \end{tabular}
  \label{toolkits}
\end{table}

\begin{table}[htbp]
 {\begin{minipage}{25pc}
\caption{Toolkits for Ancient Greek NLP.}
\centering
\begin{tabular}{lp{1.2cm}p{1.2cm}p{1.2cm}p{1.2cm}p{1.2cm}}\hline
 & Spark NLP [Ancient]	&	 CLTK [Ancient] &	 Trankit [Ancient] & greCy & Stanza [Ancient]\\ \hline
                      	&  & \citep{CLTK20019} & & \citep{greCy2022} \\ \hline
Tokenizer           	&     &  $\bullet$  &  $\bullet$  &  $\bullet$  &  $\bullet$  \\ 
Lemmatizer             	&	  $\bullet$   	&	  $\bullet$     &  $\bullet$  &  $\bullet$  &  $\bullet$ \\ 
Prosody Scanning       	&     &  $\bullet$  \\ 
Sentence  splitter     	&     & &  $\bullet$ &  $\bullet$ &  $\bullet$  \\ 
Stopword filtering     	&     &  $\bullet$ &&& \\ 
POS                	&	  $\bullet$   	&	  $\bullet$ &	  $\bullet$ &	  $\bullet$ &	  $\bullet$    \\ 
NER               	&     &  $\bullet$ &&& \\ 
Transliterator       	&     &  $\bullet$ &&& \\ 
Word2Vec            	&     & $\bullet$  &&&\\  
\hline 
 \end{tabular}
  \label{ancienttoolkits}
  \end{minipage}}
\end{table}

Table~\ref{tools} 
provides a categorized list of 25 tools.
It is worth noting  that the code in most of these tools is available, 
something that facilitates their reuse and improvement.


\begin{longtable}{p{22mm}p{25mm}p{6cm}}
\caption{Tools. The [multi] notation means that the specific tool supports more languages but a specialized solution (e.g., rules) has been introduced for the support of the Greek language.} \label{tools} \\
\textbf{Category} & \textbf{Tool} & \textbf{Description $|$ References} \\ \hline
\endfirsthead




Phonemes & espeak-ng [multi] & An open-source text-to-speech synthesizer that supports more than 100 languages and accents. It can be used for the translation of text to phonemes. $|$ \myfootnote{\url{https://github.com/espeak-ng}}\\ \midrule
Syllabication     & pyphen [multi]         & \myfootnote{\url{https://github.com/Kozea/Pyphen}}  \\\midrule

Sentence Segmentation     & sentence-splitter [multi] & Text to sentence splitter.  $|$ \myfootnote{\url{https://github.com/mediacloud/sentence-splitter}}  \\\midrule
Lemmatizers             & cstlemma  [multi]        & CST’s lemmatiser analyses not just the endings of words for suffixes
that undergo change under lemmatisation, but also
prefixes and infixes, if necessary.  $|$ ~\citep{Jongejan2009cstlemma}\myfootnote{\url{https://github.com/kuhumcst/cstlemma}} \\
                        & GLEM [Ancient]    & This lemmatizer employs POS information to disambiguate and  also assigns output to unseen words. $|$~\citep{bary2017memory}    \\
                                  & AGILe [Ancient]   &  It is a lemmatizer for Ancient Greek inscriptions. $|$  ~\citep{degraaf2022LREC} \myfootnote{\url{https://github.com/agile-gronlp/agile}}    \\
                        
                        \midrule

Inflection & Flexy  &  A tool in Python for the inflection of Greek words. $|$ \myfootnote{\url{https://github.com/stevestavropoulos/flexy}}  \\\midrule

Stemmers                & Mitos             & A stemmer designed for the Greek Language.   $|$~\citep{papadakos2008anatomy}\myfootnote{\url{https://github.com/YannisTzitzikas/GreekMitosStemmer}} \\ 
                        & Skroutz           & This is stemmer for the Greek language offered as plugin for ElasticSearch\footnote{\url{https://www.elastic.co}} $|$ \myfootnote{\url{https://github.com/skroutz/elasticsearch-skroutz-greekstemmer}}\\
                        &  NCSR             & A stemming system for the Greek language written in Python. $|$ \myfootnote{\url{https://github.com/kpech21/Greek-Stemmer}} \\\midrule   
POS taggers             & TreeTagger [multi]      &  It is a tool for annotating text with POS and lemma information.  $|$ ~\citep{Schmid1994TreeTagger} \myfootnote{\url{https://www.cis.uni-muenchen.de/~schmid/tools/TreeTagger}}  \\



                        & AUEB1             & DL type POS tagger and dependency parser  for Greek.  $|$ ~\citep{Kyriakakis2018aueb1} \myfootnote{\url{https://github.com/makyr90/DL_Syntax_Models}}\\
                        & AUEB2             & POS tagger. $|$ \myfootnote{\url{http://nlp.cs.aueb.gr/software.html}} \\ 
                   &Wiktionary& A POS tagger and a lemmatizer based on greek Wiktionary. $|$ \myfootnote{\url{https://github.com/polyvios/el-wiktionary-parser}}  \\
                    & Morpheus [Ancient]& Morpheus is a morphological parsing tool. $|$ \myfootnote{\url{https://github.com/perseids-tools/morpheus}} \myfootnote{\url{https://sites.tufts.edu/perseusupdates/2012/11/01/morphology-service-beta/}} \\ 
                        & Kanónes [Ancient]& A framework for building corpus-specific parsers for Ancient Greek. $|$ \myfootnote{\url{https://neelsmith.github.io/kanones/}} \\ 
                        & AnGEL [Ancient]& A neural network based morphology tagger. $|$ \myfootnote{\url{https://github.com/chrisdrymon/angel}}  \\ 
                        \hline   
NER                     & AUEB  & \citep{lucarelli2007named} \myfootnote{\url{http://nlp.cs.aueb.gr/software.html}}  \\
                        & Palladino [Ancient] &\citep{Palladino2020} \myfootnote{\url{https://github.com/farikarimi/NERonAncientGreek}} \\ \midrule
Summarization           & JInsect [multi] &  Summarization, summary evaluation etc. $|$\citep{Giannakopoulos2009summary}  \myfootnote{\url{https://github.com/ggianna/JInsect}}        \\  
\midrule
Similarity & corpus\_similarity [multi] & A tool that measures the similarity between two corpora. It supports texts in 74 languages. $|$ \myfootnote{\url{https://github.com/jonathandunn/corpus_similarity}}   \\
\midrule
OCR &Gamera Addon [Polytonic]&\myfootnote{\url{https://gamera.informatik.hsnr.de/addons/index.html}}         \\
& Ancient Greek OCR [Ancient]
& A tool for the  conversion of scanned printed Ancient Greek into unicode text and PDF files. $|$ ~\citep{white2012training}\myfootnote{\url{https://ancientgreekocr.org/}}  \\
\midrule
Linguistic development environment & NooJ [multi]& Greek linguistic resources for the NooJ platform\footnote{\url{https://nooj.univ-fcomte.fr}}.   $|$ ~\citep{Papadopoulou2021Nooj}  \myfootnote{\url{http://www.nooj-association.org/resources/el.zip}}

 \\
 

\midrule
Transliteration& GrElotConverter & An online tool for the Latin transliteration of Greek. $|$ \myfootnote{\url{https://www.passport.gov.gr/passports/GrElotConverter/GrElotConverterEn.html}}  
\\ 

\midrule
MedicalNLP & Iatrolexi &  A suite of tools and resources for the biomedical domain of the Greek language.  $|$ ~\citep{Vagelatos2011biomedical} \myfootnote{\url{ http://www.iatrolexi.gr/iatrolexi/ergaleia.html}} \\ 

\end{longtable}


Table \ref{interfaces}  provides information about \revision{19} search,  visualization and application tools offered as web interfaces.\\

\begin{table}[htbp!]
 {\begin{minipage}{25pc}
\caption{Search, visualization and application tools offered in web interfaces.}
\label{interfaces}
\centering
\begin{tabular}{p{3cm}p{10cm}}\hline
Tool  & References $|$ Description    \\\hline
Abridged TLG [Ancient] & \myfootnote{\url{http://stephanus.tlg.uci.edu/abridged.php}}  $|$ It provides free searching and browsing to a subcorpus of Thesaurus Linguae Graecae~\citep{mariacpantelia2022}.\\
Alpheios [Ancient] & \myfootnote{\url{http://alpheios.net/}}  $|$ Free open-source studying facilities for Classical literature.\\

Antigrapheus [Ancient]& \myfootnote{\url{https://dcthree.github.io/antigrapheus/about.html}} $|$ It is an in-browser OCR of Ancient Greek and Latin. The code is also available at \myfootnote{\url{https://github.com/dcthree/antigrapheus}}. The Ancient Greek OCR training file for Tesseract is used for Ancient Greek language (see section~\ref{sec:OCRancient}). \\
Centre for the Greek Language lexicons&\myfootnote{\url{http://georgakas.lit.auth.gr/dictionaries}}~\citep{CGL2022} $|$ A lemma-based web search interface for various lexicons (for the time being five) that concern Ancient, Modern and dialects of Greek Language. \\     

 DBBE [Ancient] &\myfootnote{\url{https://www.dbbe.ugent.be}}  $|$  The Database of Byzantine Book Epigrams (DBBE) makes available both textual and contextual data of book epigrams from medieval Greek manuscripts.\\
DGE [Ancient] &
                       \myfootnote{\url{http://dge.cchs.csic.es/xdge/}}  $|$  This is an online version of the  Greek-Spanish Dictionary (DGE). This resource is at the time being incomplete. 
                       
                       \\
   
 Dialects [Ancient] &\myfootnote{\url{https://ancdialects.greek-language.gr/}}  $|$  Material for Ancient Greek dialects such as documented inscriptions and a lexicon of proper names.\\

 Diogenes [Ancient] &\myfootnote{\url{https://d.iogen.es}} $|$ It is a search interface for the Perseus corpora and the LSJ dictionary. Diogenes is available as a standalone application and via a web interface.      \\ 
      Eulexis [Ancient] & \myfootnote{\url{https://outils.biblissima.fr/en/eulexis}} \citep{Eulexis2019dictionaries} $|$ It is a free and open source software, available both as standalone application and via a web interface for the lemmatization and the lookup in Ancient Greek dictionaries.   \\
      Greek Word Embeddings& \myfootnote{\url{http://nlp.polytechnique.fr/resources-greek}} \citep{outsios2018word,lioudakis2019ensemble} $|$ Word embeddings visualizer and various retrieval tools.
\\

 GREgORI [Ancient] &
                       \myfootnote{\url{https://www.gregoriproject.com/greek/}}~\cite{Gregori2020}  $|$  It offers open-access via an online interface to lemmatized corpora in Classical and Byzantine Greek.   
                      \\

Idioms& \myfootnote{\url{https://idioms.iliauni.edu.ge/?q=en}}~\citep{idioms} $|$ It is an online dictionary of  idiomatic expressions in Georgian and Modern Greek. It allows the search for idioms  by keywords or phrases as well as alphabetically.\\

LMPG [Ancient] &
                       \myfootnote{\url{http://dge.cchs.csic.es/lmpg/}}  $|$  This is an online version of the Greek-Spanish lexicon for magic and religion in the Greek magical papyri (Léxico de magia y religión en los papiros mágicos griegos) originally published in 2001.\\  
           Logeion [Ancient] &
                       \myfootnote{\url{https://logeion.uchicago.edu/about}}  $|$  Open-access database of Latin and Ancient Greek dictionaries.\\
         Metadrasi&\myfootnote{\url{https://www.metadrasi.org/lexiko/select_language.htm}} $|$ A web interface that provides essential educational content and vocabulary for novice learners of Greek language. The languages that are covered are Albanian, Arabic, Georgian, Russian, Urdu and Punjabi.\\     
                

 Neurolingo&\myfootnote{\url{https://www.neurolingo.gr/}} $|$It provides  various tools among them a lemmatizer, a toponym dictionary, a speller checker and  a grammar checker.\\  
         
                      Packard [Ancient]& \myfootnote{\url{https://inscriptions.packhum.org}} $|$ Greek inscriptions.\\
                       
Readability check&\myfootnote{\url{https://www.greek-language.gr/certification/readability}} \citep{Mikros2021Tool} $|$ A web interface that offers the calculation of the readability index score for a given piece of text. For this calculation a readability formula has been introduced.\\             
 Scaife [Ancient]&\myfootnote{\url{https://scaife.perseus.org/}}  $|$ It is a reading environment for pre-modern text collections in both their original languages and in translation.\\

\hline
    \end{tabular}
    \end{minipage}}
\end{table}




\clearpage

\section{Concluding Remarks}
\label{sec:CR}

To aid those who are interested in  developing or advancing the techniques for processing Greek language,  
in this paper we have presented  a list of related works and  resources,
organized according to some general thematic categories. In total we  have found, categorized and described in brief, around  166 papers,
{38 corpora, 18 datasets, 10 toolkits, 25 tools and 19 web interfaces. 
We chose not to put emphasis to language agnostic approaches, without however question their significance, but instead to focus on works that take into account language and cultural characteristics of the Greek language, following  in this respect a recent publication thread  ~\citep{Kucera2022Culture,Hershcovich2022culture}.

%

We have observed a rapid progress the last 7 years in all manifestations of Greek language (Ancient, Modern, dialects) that spreads out to  versatile tasks of NLP,  
and an increase of open resources.
Around 19\% of the works are about Ancient Greek. Most of the studies about Modern Greek originate from Greek universities and institutes, while most of the works about Ancient Greek originate from organizations outside Greece. 
We have found a relatively high number of papers on sentiment analysis,
and a rekindle interest for dialogue systems. On the antipode, we have observed a relatively low number of works in tasks like question answering, summarization  and coreference resolution. Lastly, the need of more datasets of a certain quality is a diachronic request
for the progress of the area.
The taxonomy in~\citep{joshi-etal-2020-state} that is  based on a quantitative investigation on labelled and unlabelled datasets for 2,485 languages, places Greek in the middle of the  6-scale classification in the ``Rising Stars'' category. 
Clearly, there is room for improvement, 
at least for the creation of lablelled datasets. 
Certain  requirements like  
distribution in interoperable and standard formats, 
continuous maintenance, and  application of  ethical collection processes,
must also be taken into account.







    We believe our survey provides a timely guidance on the literature and resources about Greek NLP
    for researchers, engineers and  educators. 
    

\bibliographystyle{unsrt}
\bibliography{template}

\end{document}